%% file: main.tex
\documentclass[10pt,twocolumn,letterpaper]{article}

\usepackage[pagenumbers]{cvpr} % To force page numbers, e.g. for an arXiv version

\input{preamble}

\definecolor{cvprblue}{rgb}{0.21,0.49,0.74}
\usepackage[pagebackref,breaklinks,colorlinks,allcolors=cvprblue]{hyperref}

\input{math_commands.tex}

\def\paperID{3106} % *** Enter the Paper ID here
\def\confName{CVPR}
\def\confYear{2026}

\title{Unleashing the Intrinsic Visual Representation Capability of\\Multimodal Large Language Models}

\author{Hengzhuang Li$^{\text{\ding{168}},*}$\quad Xinsong Zhang$^{\text{\ding{169},\text{\Letter}}}$\quad Qiming Peng$^{\text{\ding{169}}}$\quad Bin Luo$^{\text{\ding{169}}}$\\
\vspace{3pt}Han Hu$^{\text{\ding{169}}}$\quad Dengyang Jiang$^{\text{\ding{170}}}$\quad Han-Jia Ye$^{\text{\ding{171}}}$\quad Teng Zhang$^{\text{\ding{168},\text{\Letter}}}$\quad Hai Jin$^{\text{\ding{168}}}$\\
\vspace{-4pt}\normalsize$^{\text{\ding{168}}}$HUST\quad $^{\text{\ding{169}}}$Tencent Hunyuan Research\quad $^{\text{\ding{170}}}$HKUST\quad $^{\text{\ding{171}}}$NJU \\
\small $^*$Work done at Tencent Hunyuan Research\quad $^{\text{\Letter}}$Corresponding authors
}

\begin{document}
\maketitle
% \renewcommand{\thefootnote}{\fnsymbol{footnote}}
% \footnotetext[1]{Work done during internship at Tencent Hunyuan Research.}
% \renewcommand{\thefootnote}{\arabic{footnote}}

\input{sec/0_abstract}    
\input{sec/1_intro}

\input{sec/2_related}

\input{sec/3_preliminary}

\input{sec/4_method}

\input{sec/5_expr}
\input{sec/6_conclude}
% \input{sec/3_finalcopy}
{
    \small
    \bibliographystyle{ieeenat_fullname}
    \bibliography{main}
}

% WARNING: do not forget to delete the supplementary pages from your submission 
\input{sec/X_suppl}

\end{document}

%% file: preamble.tex
\usepackage{mathtools}
\usepackage{amsmath}
\usepackage{algorithm}
\usepackage{algpseudocode}
\usepackage{multirow}
\usepackage{float}
\usepackage{booktabs}
\usepackage{colortbl}
\usepackage{xcolor}
\usepackage{makecell}
\usepackage{wrapfig}
\usepackage{upgreek}
\usepackage{pifont}
\usepackage{nicefrac}
\usepackage{marvosym}

\definecolor{BoldDelta}{HTML}{287EB8}
\definecolor{CiteBlue}{HTML}{001473}
\definecolor{Highlight}{HTML}{C74167}
\definecolor{LightGray}{HTML}{F0F1F2} % F0F1F2 ebeced
\definecolor{superlighgray}{HTML}{F2F2F2}
\definecolor{Grey}{HTML}{D9D9D9}
\definecolor{GainGreen}{HTML}{009667}
\definecolor{LossRed}{HTML}{de2d2d}

\newcommand{\gain}[1]{\textcolor{GainGreen}{\small$\uparrow$#1}}
\newcommand{\loss}[1]{\textcolor{LossRed}{\small$\downarrow$#1}}

%% file: math_commands.tex
\usepackage{amsmath,amsfonts,bm}

\def\eqref#1{equation~\ref{#1}}
\def\1{\bm{1}}

\DeclareMathAlphabet{\mathsfit}{\encodingdefault}{\sfdefault}{m}{sl}
\SetMathAlphabet{\mathsfit}{bold}{\encodingdefault}{\sfdefault}{bx}{n}

%% file: sec/0_abstract.tex
\begin{abstract}
Multimodal Large Language Models (MLLMs) have demonstrated remarkable proficiency in multimodal tasks.
Despite their impressive performance, MLLMs suffer from the modality imbalance issue, where visual information is often underutilized compared to textual representations in deeper layers, leading to degraded visual performance or hallucinations.
This issue stems from the predominant reliance on next-text-token-prediction during training, which fails to provide direct visual supervisory signals, resulting in progressive homogenization of visual representations throughout the layers.
To this end, we propose \textit{\textbf{La}tent \textbf{V}isual R\textbf{e}const\textbf{r}uction} (\textbf{LaVer}), a novel training framework that facilitates MLLMs in learning more discriminative visual representations via masked image modeling in the joint latent semantic space of LLM.
Our method offers direct visual activation to MLLMs, which exhibit increased visual attention allocation, indicating enhanced utilization of visual information.
Extensive experiments across diverse benchmarks prove the superiority of our approach in various scenarios, especially those requiring dense visual capabilities.
Code of LaVer is available at \textsf{\url{https://github.com/Fir-lat/LaVer}}.
\end{abstract}
\vspace{-10pt}

% Our method encourages the MLLMs to improve the intrinsic visual representation modeling more effectively throughout the training process.
% Our method encourages MLLMs to jointly learn on the language task and the vision pretext task within the same high-level semantic space, thus producing more unified representations for multimodal tasks.

%% file: sec/1_intro.tex
\section{Introduction}
\label{sec:intro}

% introduction
Multimodal Large Language Models (MLLMs)~\cite{10.5555/3600270.3601993,10.5555/3666122.3668264,liu2023llava,bai2023qwenvlversatilevisionlanguagemodel,liu2023improvedllava,liu2024llavanext,li2025llavaonevision,10.5555/3737916.3740687,an2025llavaonevision15fullyopenframework,bai2025qwen25vltechnicalreport} have emerged as a transformative paradigm for real-world multimodal tasks, fusing the linguistic reasoning power of Large Language Models (LLMs)~\cite{Radford2018ImprovingLU,radford2019language,10.5555/3495724.3495883,bai2023qwentechnicalreport,touvron2023llamaopenefficientfoundation,grattafiori2024llama3herdmodels,openai2024gpt4technicalreport,yang2024qwen2technicalreport,yang2025qwen3technicalreport} with the advanced cognition and perception abilities of vision models~\cite{pmlr-v139-radford21a,Zhai_2023_ICCV,tschannen2025siglip2multilingualvisionlanguage,oquab2024dinov,siméoni2025dinov3}. MLLMs have demonstrated significant potential in enhancing visual capabilities across a wide spectrum of tasks, ranging from visual question answering to complex multimodal reasoning~\cite{Antol_2015_ICCV,zhang2023multicot,fu2025mmecomprehensiveevaluationbenchmark,li2025imaginereasoningspacemultimodal}. 

Despite their impressive capabilities, MLLMs are afflicted by the \textit{modality imbalance} issue~\cite{10.1145/3746027.3755364,zheng2025mllmsdeeplyaffectedmodality,10.1016/j.patcog.2025.111670}, where the model exhibits a systematic bias towards textual information in multimodal tasks. MLLMs frequently generate confident responses even when presented with absent or incongruent visual inputs, as empirically demonstrated in~\cite{10.1145/3746027.3755364,zheng2025mllmsdeeplyaffectedmodality,10.1609/aaai.v39i19.34183,liu2025faithfulnessvisualthinkingmeasurement}. When rich visual information is available, MLLMs still tend to produce outputs predominantly grounded in textual context~\cite{jia2024symdpoboostingincontextlearning,leng2024cursemultimodalitiesevaluatinghallucinations}. Qualitative analysis reveals that MLLMs allocate substantially more attention to textual content than vision tokens~\cite{wu2025languageoverrulesrevealingtext}, as the text modality demonstrably contributing more to the predictions~\cite{10.1609/aaai.v39i19.34183}.

\begin{figure}[t]
  \centering
  % \fbox{\rule{0pt}{2.5in} \rule{0.9\linewidth}{0pt}}
   %\includegraphics[width=0.8\linewidth]{egfigure.eps}
   \includegraphics[width=0.99\linewidth]{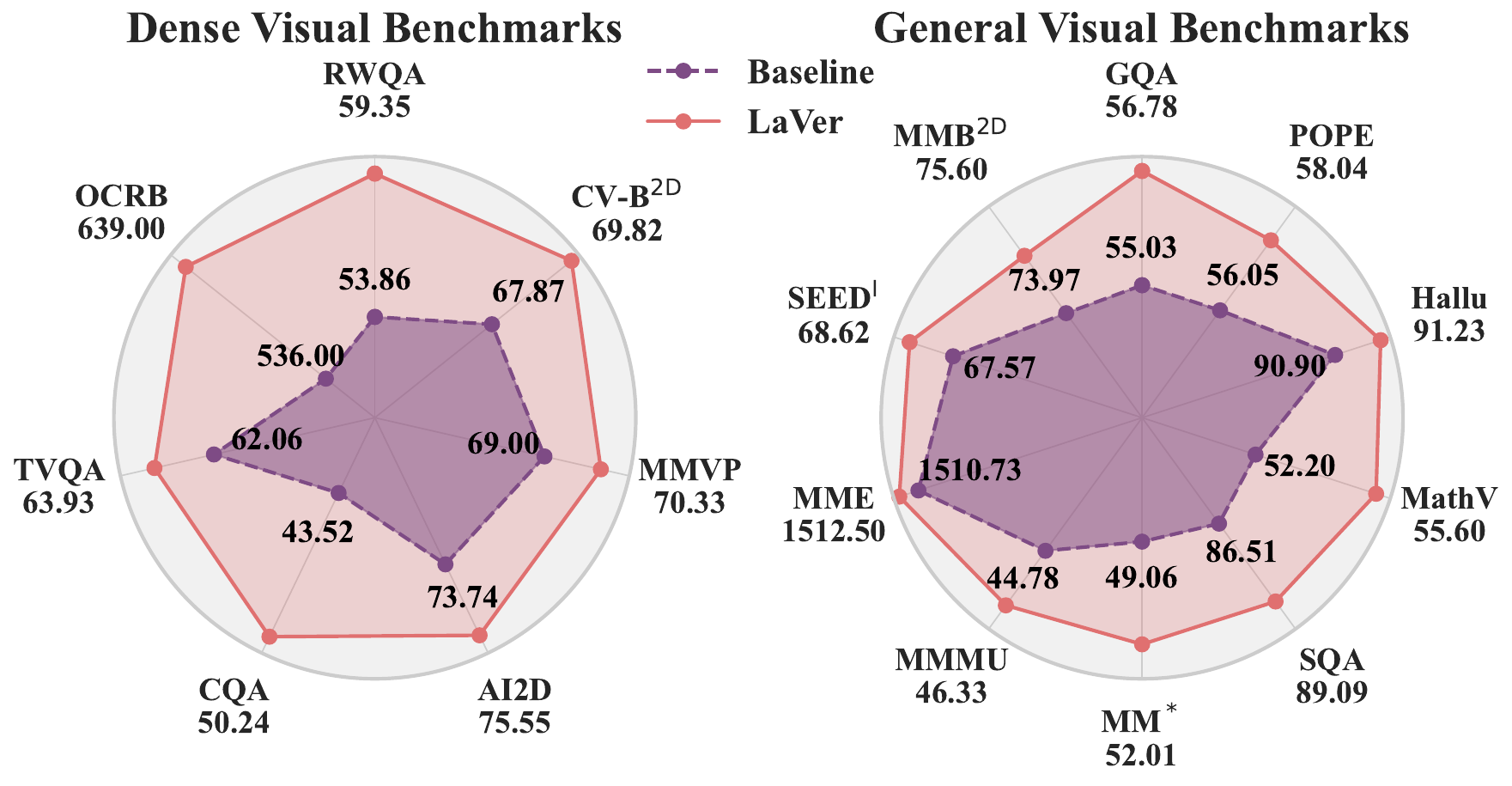}
   \vspace{-10pt}
   \caption{\textbf{Benchmark Performance.} LaVer consistently outperforms the baseline across diverse benchmarks, especially on dense visual tasks such as OCRB~\cite{Liu_2024} and CQA~\cite{masry-etal-2022-chartqa}. The results are obtained with SigLIP 2~\cite{tschannen2025siglip2multilingualvisionlanguage} and Qwen2.5-7B-Instruct~\cite{qwen2025qwen25technicalreport}.}
   \label{fig:laver_performance}
   \vspace{-10pt}
\end{figure}

\begin{figure*}
  \centering
  \begin{subfigure}[c]{0.52\linewidth}
    % \fbox{\rule{0pt}{2.25in} \rule{.9\linewidth}{0pt}}
    \includegraphics[width=0.99\linewidth]{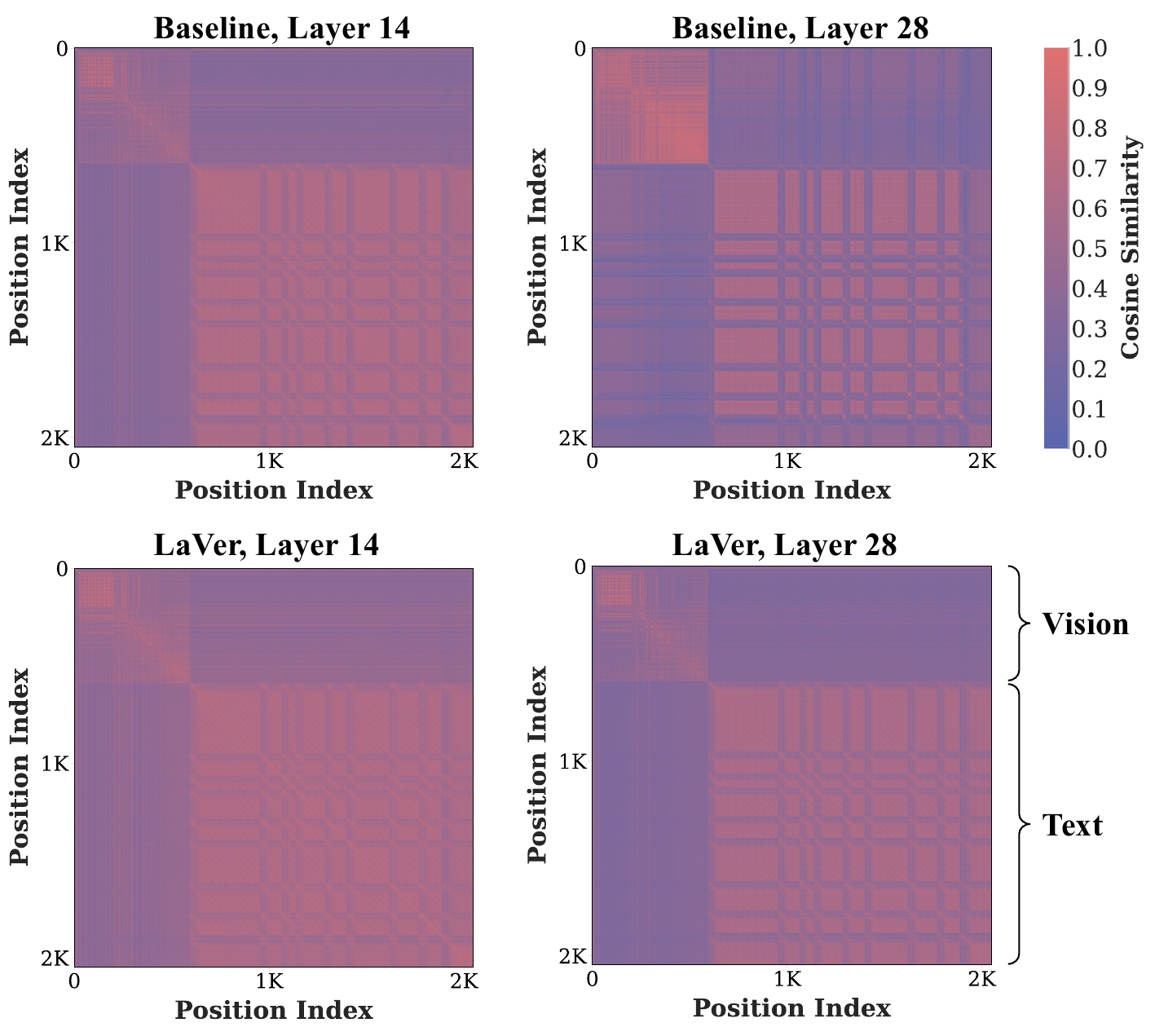}
    \caption{Cosine similarity between all tokens.}
    \label{fig:motivation-a}
  \end{subfigure}
  \hfill
  % \begin{subfigure}[c]{0.24\linewidth}
  %   % \fbox{\rule{0pt}{2.25in} \rule{.9\linewidth}{0pt}}
  %   \includegraphics[width=0.99\linewidth]{fig/2_fig_tsne.pdf}
  %   \caption{t-SNE Visualization.}
  %   \label{fig:motivation-b}
  % \end{subfigure}
  \begin{minipage}[c]{0.22\linewidth}
    \centering
    \begin{subfigure}{\linewidth}
      % \fbox{\rule{0pt}{1in} \rule{.9\linewidth}{0pt}}
      \includegraphics[width=0.99\linewidth]{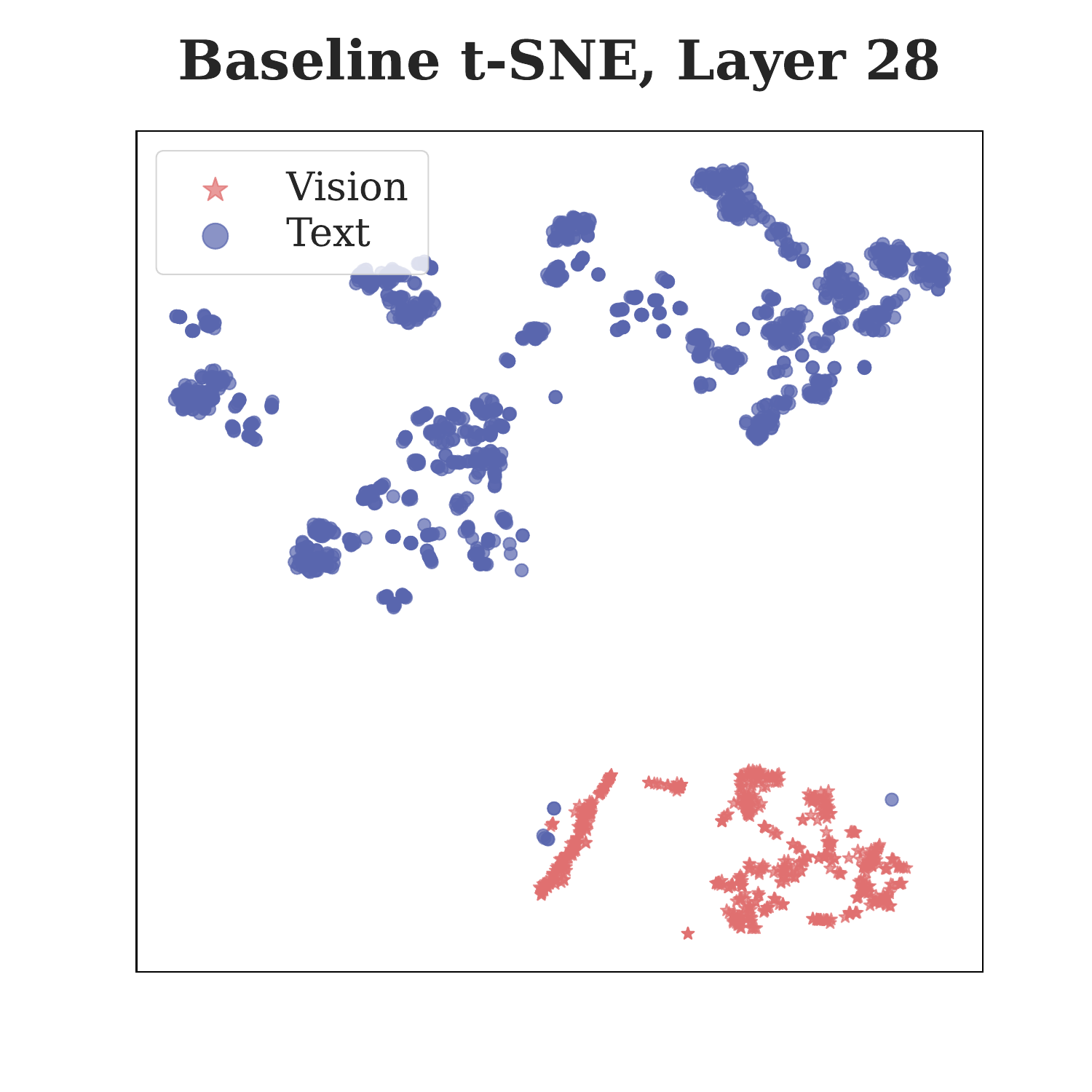}
      \caption{t-SNE of baseline.}
      \label{fig:motivation-b}
    \end{subfigure}
    \vfill
    \begin{subfigure}{\linewidth}
      % \fbox{\rule{0pt}{1in} \rule{.9\linewidth}{0pt}}
      \includegraphics[width=0.99\linewidth]{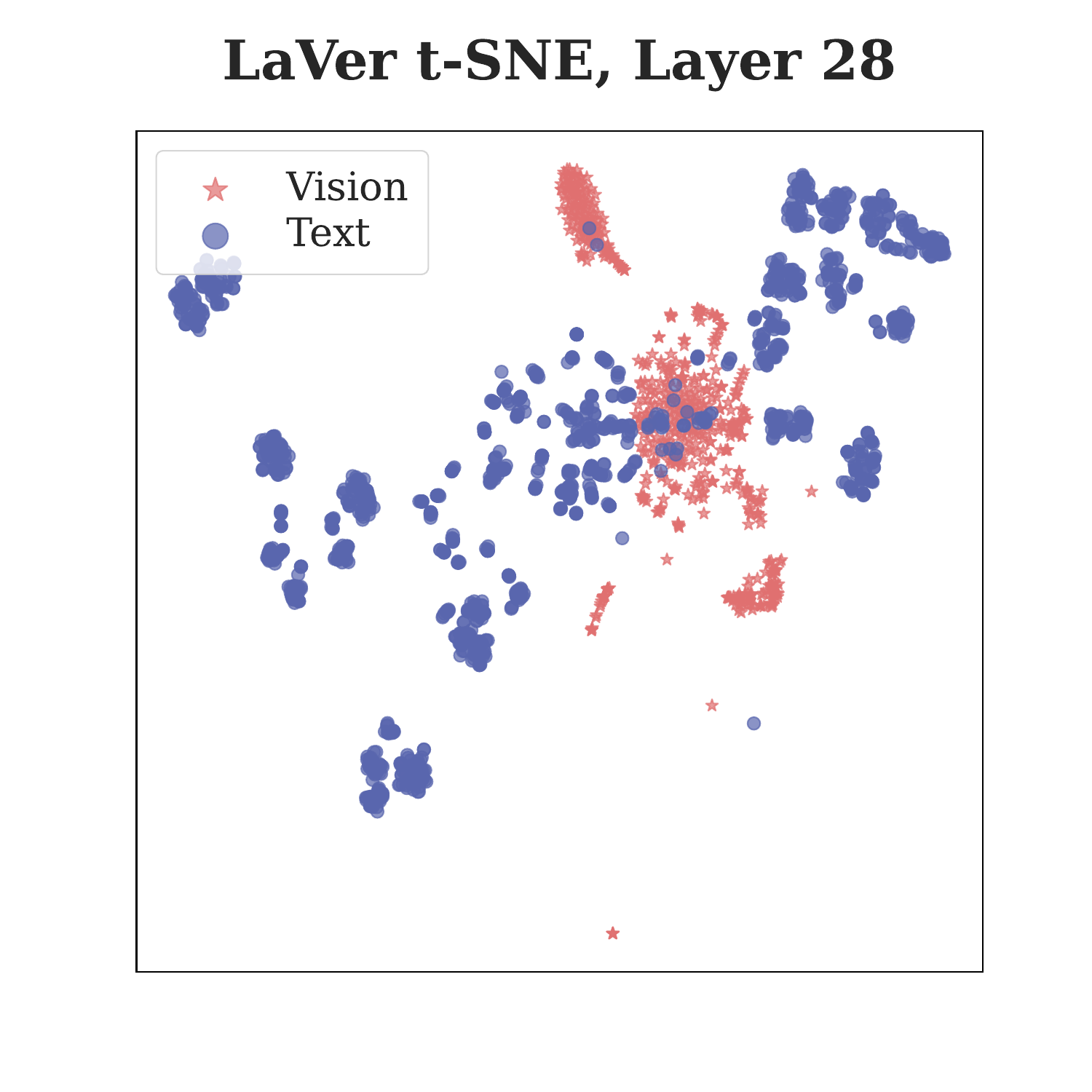}
      \caption{t-SNE of LaVer.}
      \label{fig:motivation-c}
    \end{subfigure}
  \end{minipage}
  \hfill
  \begin{minipage}[c]{0.24\linewidth}
    \centering
    \begin{subfigure}{\linewidth}
      % \fbox{\rule{0pt}{1in} \rule{.9\linewidth}{0pt}}
      \includegraphics[width=0.99\linewidth]{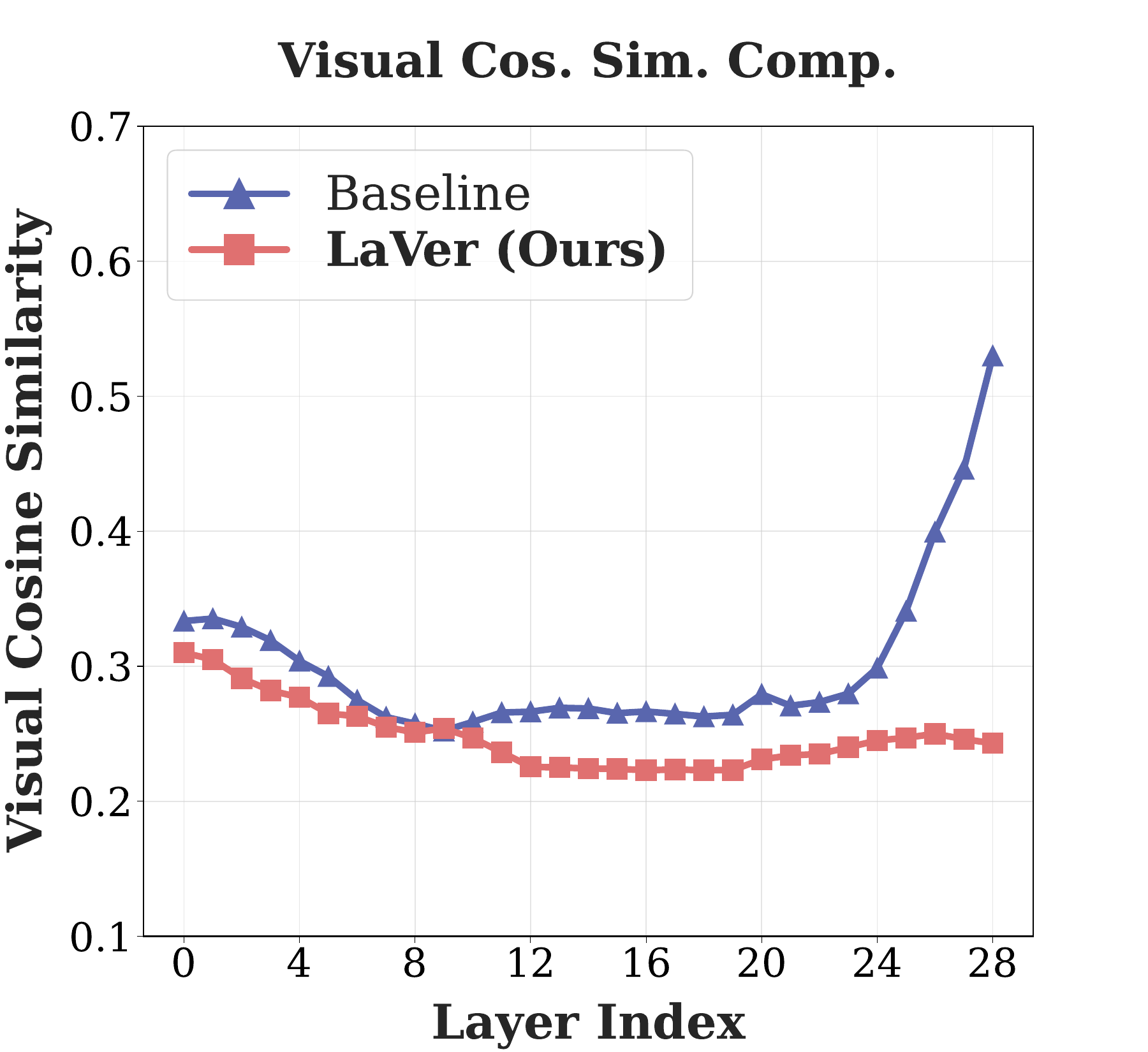}
      \caption{Visual cosine similarity.}
      \label{fig:motivation-d}
    \end{subfigure}
    \vfill
    \begin{subfigure}{\linewidth}
      % \fbox{\rule{0pt}{1in} \rule{.9\linewidth}{0pt}}
      \includegraphics[width=0.99\linewidth]{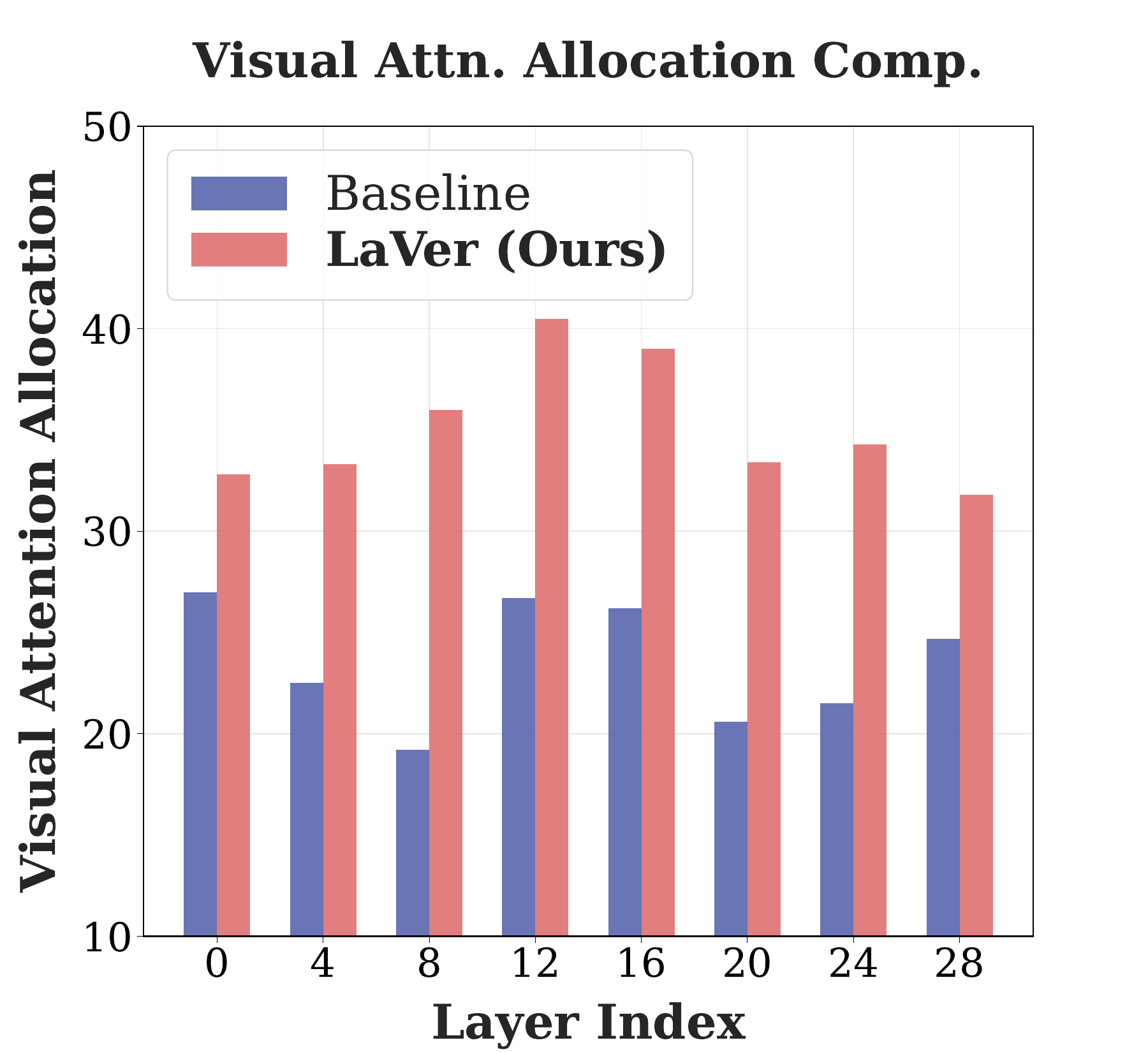}
      \caption{Visual attention allocation.}
      \label{fig:motivation-e}
    \end{subfigure}
  \end{minipage}
\vspace{-5pt}
  \caption{\textbf{Progressive visual representation homogenization.} \textbf{(a)} presents higher feature cosine similarities of the last layer than middel layer. \textbf{(b-c)} display the t-SNE visualizations of the output embeddings. \textbf{(d)} quantifies the averaged vision cosine similairity. \textbf{(e)} quantifies the allocated attention score for vision tokens. LaVer outputs discriminative visual representations with higher attention allocation. The quantitative results are obtained with with SigLIP 2~\cite{tschannen2025siglip2multilingualvisionlanguage} and Qwen2.5-7B-Instruct~\cite{qwen2025qwen25technicalreport}, averaged across the images from MMVP~\cite{10655378}.}
  \label{fig:motivation}
  \vspace{-10pt}
\end{figure*}

% consequence & significance
The consequences of this modality imbalance are multifaceted. Model outputs become systematically biased toward the dominant textual modality, fundamentally failing to leverage the full potential of underrepresented visual information~\cite{10.1109/TMM.2024.3380259,10.5555/3326943.3327085}, leading to degraded performance across various benchmarks~\cite{zheng2025mllmsdeeplyaffectedmodality,liu2025insightsightexploringvisionknowledge,qi2025semanticsrediscoveringspatialawareness} and increased visual hallucinations~\cite{rohrbach-etal-2018-object,leng2023mitigatingobjecthallucinationslarge,leng2024cursemultimodalitiesevaluatinghallucinations,guan2024hallusionbenchadvanceddiagnosticsuite,wang2024mitigatinghallucinationslargevisionlanguage}, ultimately diminishing the reliability~\cite{liu2025faithfulnessvisualthinkingmeasurement,10655378} of MLLMs. Addressing this challenge is crucial, as ideal MLLMs should seamlessly integrate and leverage the full potential of each modality to achieve robust multimodal understanding~\cite{chen2024quantifyingmitigatingunimodalbiases,10655378}.

% root cause: supervision derives exclusively from text data~\cite
% {wang2025reconstructive,wang2025ross3dreconstructivevisualinstruction,
% zhu2024languagebindextendingvideolanguagepretraining}; visual 
% comprehensnsion largely depends on vision-to-text alignment~\cite
% {10655378,10.5555/3737916.3740687,
% lyu2024unibindllmaugmentedunifiedbalanced}.
% specifically, MLLMs are trained with multimodal inputs (e.g., 
% encoding images into prefix multimodal tokens), but with only text 
% outputs, on which the language modeling loss is calculated by 
% predicting the next text token, offering indirect supervision for 
% model's intrinsic visual representation~\cite{10655378,
% wang2025reconstructive}.

The modality imbalance stems from the fundamental asymmetry of MLLM's training paradigm: supervision signals are derived exclusively by \textit{next-text-token-prediction}~\cite{wang2025reconstructive,wang2025ross3dreconstructivevisualinstruction,zhu2024languagebindextendingvideolanguagepretraining}, while visual modeling largely depend on implicit vision-to-text alignment~\cite{10655378,10.5555/3737916.3740687,lyu2024unibindllmaugmentedunifiedbalanced}, thereby providing indirect and weaker supervision signals for the model's intrinsic visual representation~\cite{10655378,wang2025reconstructive}. This asymmetric supervision naturally biases the model toward prioritizing textual information and discarding visual information that is not meaningful for text outputs, particularly given the LLM backbone's dominant language capability~\cite{zheng2025mllmsdeeplyaffectedmodality}, which is trained on large-scale corpora~\cite{qwen2025qwen25technicalreport}.

We further provide empirical validation of modality imbalance by revealing the \textit{progressive visual representation homogenization} in Fig.~\ref{fig:motivation}. We observe that visual tokens exhibit significant deterioration in deeper layers, characterized by drastically enlarged token-wise cosine similarity shown in Fig.~\ref{fig:motivation-a} and Fig.~\ref{fig:motivation-d}. Besides, t-SNE visualization in Fig.~\ref{fig:motivation-b} demonstrates that vision tokens remain significantly separated from text tokens in the output space. This homogenization pattern indicates substantial visual information loss as representations propagate through the model, which aligns with prior findings that MLLMs generate increasingly aligned yet less informative visual representations in deeper layers~\cite{venhoff2025visualrepresentationsmaplanguage}, and that vision-text information flow occurs predominantly in the early layers~\cite{zhang-etal-2025-redundancy,yin2025clearsightvisualsignalenhancement,jiang2024devils}. This evidence underscores the fundamental limitations in MLLMs' \textit{intrinsic visual representation capability}.

To this end, we propose \textit{\textbf{La}tent \textbf{V}isual R\textbf{e}const\textbf{r}uction} (\textbf{LaVer}) to enhance the intrinsic visual modeling capability of MLLMs by learning discriminative multimodal representations within the joint high-level semantic space of LLMs. Technically, we employ the \textit{Masked Image Modeling} (MIM) paradigm~\cite{pmlr-v139-touvron21a,9879206,bao2022beitbertpretrainingimage,zhou2021ibot,gidaris2024mocaselfsupervisedrepresentationlearning}, capitalizing the spatial redundancy of natural visual signals~\cite{zhou2021ibot,9879206,zhang-etal-2025-redundancy}. Different from previous techniques that apply masks to raw pixels~\cite{9879206,gidaris2024mocaselfsupervisedrepresentationlearning} or require reconstruction of fine-grained visual signals~\cite{wang2025reconstructive}, we randomly mask vision tokens in the input embedding space and train the model to recover the masked tokens in the latent space of LLM, thereby providing direct intrinsic supervision for visual representations. Our method concurrently shapes the model's latent semantic space for all modalities by learning to predict the missing vision token and the next text token. LaVer inhibits the homogenization of vision tokens in deeper layers (Fig.~\ref{fig:motivation-a} and Fig.~\ref{fig:motivation-d}), generates more unified multimodal representations (Fig.~\ref{fig:motivation-c}), and consistently increases the visual attention allocation (Fig.~\ref{fig:motivation-e}), indicating more comprehensive utilization of visual information. Furthermore, we propose \textit{Clipped Gram-Anchoring} to prevent the model from hacking MIM by outputting identical visual embeddings.

Empirical analysis demonstrates that LaVer achieves competitive performance across a wide suite of multimodal benchmarks, especially for dense visual tasks~\cite{singh2019vqamodelsread,masry-etal-2022-chartqa,Liu_2024,10655378,10.5555/3737916.3740687} that require comprehensive visual information utilization, e.g., 19.22\% improvement on OCRB~\cite{Liu_2024}, shown in Fig.\ref{fig:laver_performance}. The unleashed visual representations also facilitate sophisticated visual reasoning capabilities, such as reasoning segmentation where precise visual understanding must be tightly coupled with language instructions~\cite{lai2024lisareasoningsegmentationlarge,tang2025ufounifiedapproachfinegrained}. These results validate that our approach enables more effective multimodal learning that provides unified representations for all modalities. The primary contributions of this work are summarized as follows.

% Contribution:
% 1. first work to jointly train visual representation and language modeling in the same latent space;
% 2. competitive performance;
% 3. potential for better performance on downstream tasks, such as reasoning-based segmentation.

\begin{itemize}
  \item We investigate self-supervised paradigm for MLLMs and propose \textbf{LaVer}, a novel multimodal training framework that derives direct visual supervisory signals by reconstructing the masked vision token in the latent space. %, enhancing intrisnic multimodal representation capability.
  \item We further validate the modality imbalance problem by revealing the progressive homogenization of visual features in deeper layers of MLLMs. %, indicating the underutilization of visual information.
  \item We conduct extensive empirical analysis to validate that LaVer can consistently improve multimodal understanding, especially on dense visual tasks. % Comprehensive ablation studies further exhibits the robustness of LaVer.
  % \item First, LaVer represents the first systematic attempt to jointly optimize visual representation learning and language modeling within a unified latent semantic space~\cite{todo} for MLLMs. By establishing this joint training paradigm, we fundamentally address the modality imbalance problem at its core, enabling MLLMs to develop more balanced and harmonious multimodal representations that genuinely integrate visual and textual information~\cite{todo}. 
  % \item Second, extensive experimental evaluation demonstrates that our approach achieves competitive performance across a diverse range of multimodal benchmarks, validating the effectiveness of our method in enhancing both visual understanding and cross-modal reasoning capabilities~\cite{todo}. 
  % \item Third, beyond standard benchmark evaluations, LaVer exhibits promising potential for improved performance on sophisticated downstream applications that demand tight integration of visual perception and linguistic reasoning, such as reasoning-based segmentation tasks where models must precisely ground textual instructions in visual content~\cite{todo}. These contributions collectively advance the state-of-the-art in multimodal learning by providing a principled framework for unleashing the intrinsic visual representation capability of MLLMs~\cite{todo}.
\end{itemize}

%% file: sec/2_related.tex
\section{Related Works}
\label{sec:related}

\subsection{Multimodal Large Language Models}
\label{sec:related_mllm}

Recent advancements in LLMs~\cite{NEURIPS2020_1457c0d6,touvron2023llamaopenefficientfoundation,bai2023qwentechnicalreport,yang2024qwen2technicalreport,openai2024gpt4technicalreport,deepseekai2025deepseekv3technicalreport,yang2025qwen3technicalreport} and vision models~\cite{9709990,pmlr-v139-radford21a,Zhai_2023_ICCV,oquab2024dinov,tschannen2025siglip2multilingualvisionlanguage,siméoni2025dinov3} have catalyzed the rapid evolution of MLLMs~\cite{10.5555/3666122.3668264,10.5555/3600270.3601993,liu2023llava,li2025llavanextinterleave,li2025llavaonevision,bai2025qwen25vltechnicalreport,an2025llavaonevision15fullyopenframework,bai2023qwenvlversatilevisionlanguagemodel,liu2023improvedllava,liu2024llavanext,10.5555/3737916.3740687}. The typical paradigm for these models is a plug-in architecture~\cite{liu2023llava,an2025llavaonevision15fullyopenframework}, which incorporates a visual encoder to convert images into visual features, a connector (e.g., an MLP~\cite{liu2023llava}) to map these features into the input embedding space of LLM, and a pretrained LLM that integrates the visual information to execute multimodal tasks. MLLMs can be boosted by scaling data~\cite{bai2025qwen25vltechnicalreport,an2025llavaonevision15fullyopenframework}, powerfull LLMs~\cite{grattafiori2024llama3herdmodels,yang2025qwen3technicalreport}, and advanced ViTs\cite{dehghani2023patchnpacknavit,qwen2025qwen25technicalreport,fini2025multimodal}. However, the \textit{next-text-token-prediction} potentially hinders MLLMs from developing intrinsic visual modeling capacities.
% Preliminary studies~\cite{liu2023llava,li2025llavanextinterleave,li2025llavaonevision} established fundamental visual proficiency in LLMs using visual instruction tuning on multimodal datasets.  Recent endeavors~\cite{bai2025qwen25vltechnicalreport,an2025llavaonevision15fullyopenframework} have focused on scaling by utilizing extensive and varied multimodal corpora to markedly improve the understanding capability of MLLMs. \TODO{Native resolution Vit + Encoder-free MLLM} Nevertheless, a significant constraint endures: the training process for most MLLMs relies solely on textual supervision~\cite{wang2025reconstructive}.  This dependence can result in improper visual representations~\cite{10.5555/3737916.3739555} and significant information loss~\cite{DBLP:conf/emnlp/LinCZ024,li2025lostembeddingsinformationloss} during the inference time.  Consequently, the pursuit of enhanced and sophisticated visual supervision in MLLMs remains to be a crucial research domain.

\subsection{Vision-enhanced MLLMs}
\label{sec:related_vision}

Numerous initiatives have been undertaken to mitigate the modality imbalance~\cite{10.1145/3746027.3755364,zheng2025mllmsdeeplyaffectedmodality,10.1016/j.patcog.2025.111670} and improve the visual understanding of MLLMs for complex tasks~\cite{zhang2023multicot,10.5555/3737916.3740687,zhang2025mllmsknowlooktrainingfree}. These include curating robust, modality-balanced datasets for evaluation and training~\cite{10.1609/aaai.v39i19.34183,chen2024quantifyingmitigatingunimodalbiases,chen2024rightwayevaluatinglarge}, prompting strategies to encourage visual utilization~\cite{liu2025insightsightexploringvisionknowledge}, and visual contrastive decoding techniques to mitigate textual overreliance~\cite{niu2021counterfactualvqacauseeffectlook,gupta2022swapmixdiagnosingregularizingoverreliance}. Reinforcement learning-based approaches further enhance vision-centric optimization~\cite{liu2025faithfulnessvisualthinkingmeasurement,zhang2025debiasingmultimodallargelanguage,pi2024strengtheningmultimodallargelanguage,li2025devildetailstacklingunimodal}, while alternative strategies enrich visual inputs via additional expert modules~\cite{10655378,10.5555/3737916.3740687,lu2024deepseekvlrealworldvisionlanguageunderstanding,li2025llavaonevision,yao2024dense,chung2025unifying,yoon2025visualrepresentationalignmentmultimodal}, advanced projector routing~\cite{mckinzie2024mm1methodsanalysis,wang2025visioncentricactivationcoordinationmultimodal}, or fusion strategies~\cite{cao2024mmfusermultimodalmultilayerfeature,11094373}. Methods focusing on intrinsic visual modeling have also gained attention, including directly allocating greater attention to vision tokens~\cite{zhang2025debiasingmultimodallargelanguage,liu2025insightsightexploringvisionknowledge} and employing visual reconstruction objectives to retain the fine-grained visual information~\cite{wang2025reconstructive,wang2025ross3dreconstructivevisualinstruction,wu2025visualjigsawposttrainingimproves}, being potentially impeded by redundancy and noise in low-level pixels~\cite{nam2025extract}, thus suboptimal for tasks requiring high-level semantic reasoning~\cite{mistretta2025cross}. In contrast to these methods, LaVer predicts the masked vision tokens in the same latent space of LLM, thus producing more unified multimodal representations.

\subsection{Masked Image Modeling}
\label{sec:related_masked_signal}

Masked Image Modeling (MIM)~\cite{pmlr-v139-touvron21a,bao2022beitbertpretrainingimage,zhou2021ibot,9879206} has emerged as a cornerstone of self-supervised learning~\cite{doersch2016unsupervisedvisualrepresentationlearning,gidaris2019boostingfewshotvisuallearning}, deriving powerful supervisory signals via pretext tasks.
Early approaches explored predicting contextual relationships within images~\cite{7410524,10.1007/978-3-319-46466-4_5,9156540}.
Recent advances focus on reconstructing masked image patches: BEiT~\cite{bao2022beitbertpretrainingimage} and DAVINCI~\cite{diao2023writepaintgenerativevisionlanguage} reconstruct discrete vision tokens from VQVAE~\cite{esser2021tamingtransformershighresolutionimage}, iBOT~\cite{zhou2021ibot} employs an online teacher to provide visual targets, MAE~\cite{9879206} demonstrates the effectiveness of pixel-space reconstruction, and JEPA~\cite{assran2023self, bardes2024revisitingfeaturepredictionlearning} formalizes prediction within a learned latent space.
Extensions to multimodal scenarios~\cite{geng2022multimodalmaskedautoencoderslearn,10.1145/3539618.3591721,chen2024eveefficientvisionlanguagepretraining} foster learning unified multimodal representations. However, MLLMs haven't fully leveraged this paradigm. This study explores enhancing the intrinsic visual modeling capability of MLLMs by forcing them to reconstruct latent visual representations.

%% file: sec/3_preliminary.tex
\section{Preliminaries}

\begin{figure*}
\centering
\begin{subfigure}{0.72\linewidth}
    % \fbox{\rule{0pt}{2.25in} \rule{.9\linewidth}{0pt}}
    \includegraphics[width=0.99\linewidth]{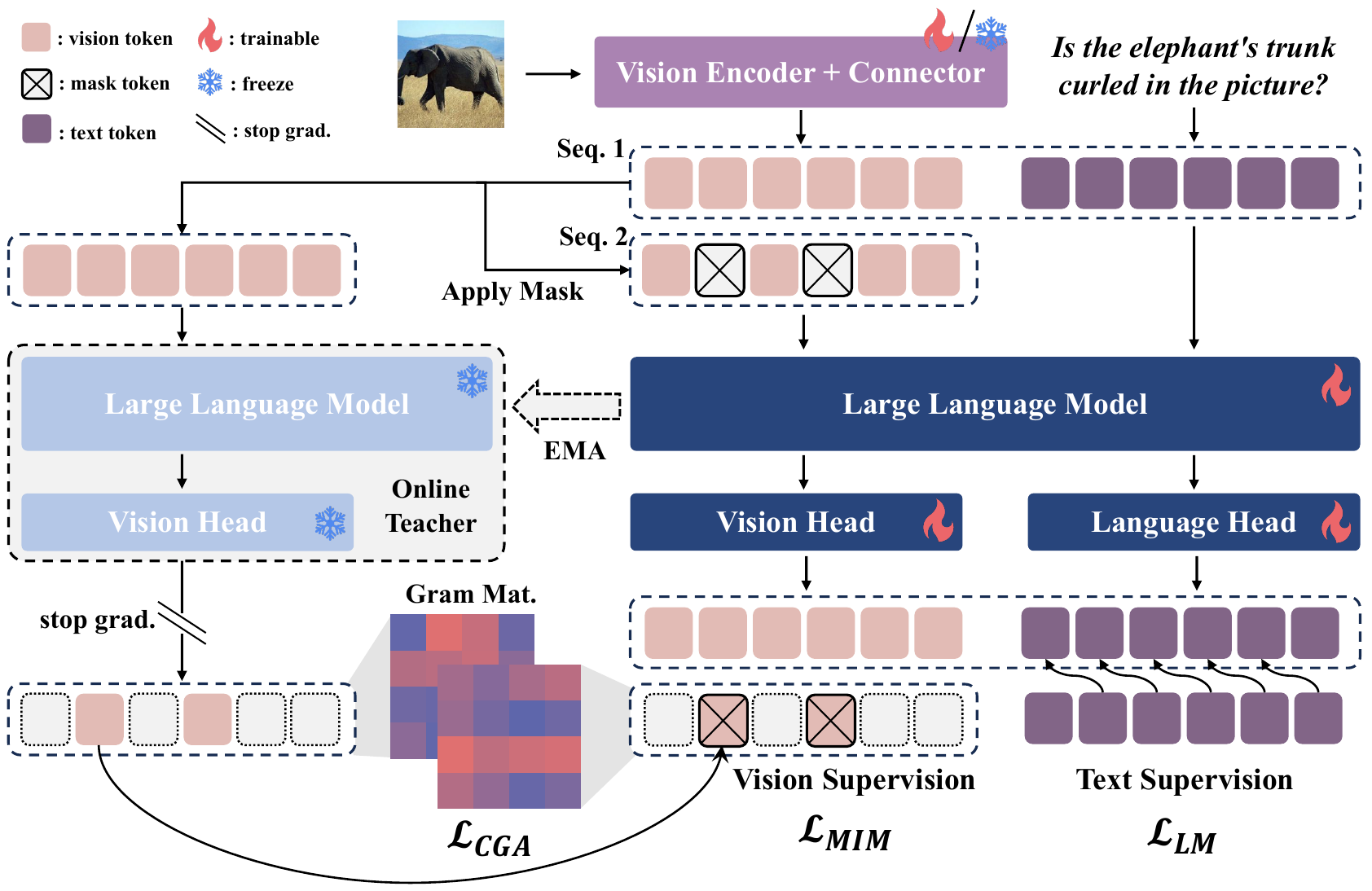}
    \caption{Illustration of model architecture of LaVer.}
    \label{fig:method-a}
\end{subfigure}
\hfill
\begin{minipage}[b]{0.24\linewidth}
    \centering
    \begin{subfigure}{\linewidth}
    % \fbox{\rule{0pt}{1in} \rule{.9\linewidth}{0pt}}
    \includegraphics[width=0.9\linewidth]{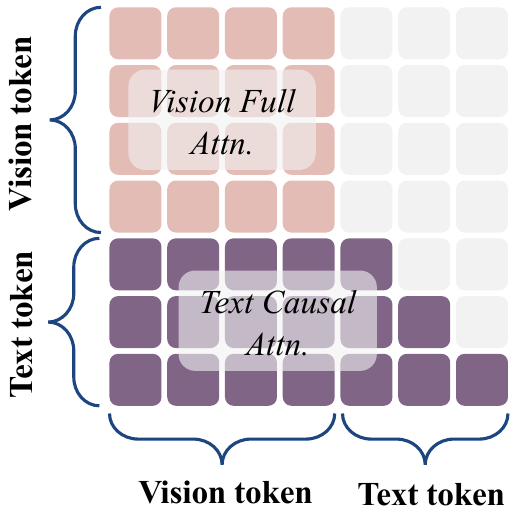}
    \caption{Illustration of mixed attention.}
    \label{fig:method-b}
    \end{subfigure}
    \vfill
    \begin{subfigure}{\linewidth}
    % \fbox{\rule{0pt}{1in} \rule{.9\linewidth}{0pt}}
    \includegraphics[width=0.99\linewidth]{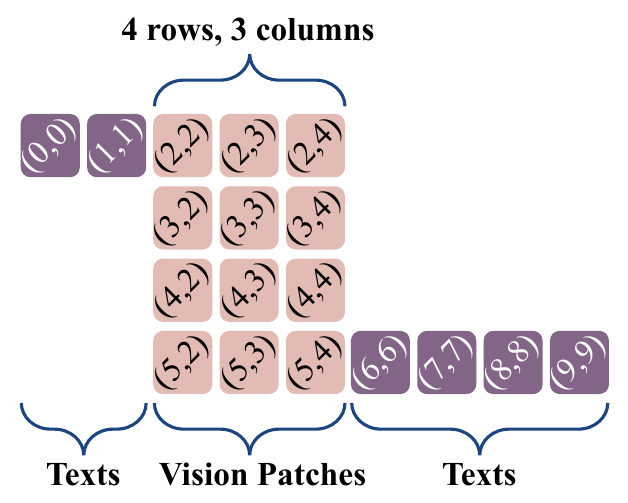}
    \caption{Illustration of 2D-ROPE.}
    \label{fig:method-c}
    \end{subfigure}
\end{minipage}
\vspace{-8pt}
\caption{\textbf{Overview of LaVer.} \textbf{(a)} depicts the student-teacher framework where the MLLM is trained to predict the teacher's visual output embeddings for the masked positions, regularized by the Clipped Gram-Anchoring to prevent feature inconsistency. \textbf{(b)} depicts the mixed attention mechanism. \textbf{(c)} depicts the 2D-ROPE mechanism. LaVer learns discriminative visual representations by self-supervised MIM.}
\label{fig:method}
\vspace{-10pt}
\end{figure*}

\textbf{Multimodal Large Language Models.} Denote the LLM backbone as $\mathcal{F}_{\theta}$, parameterized by $\theta$, modeling the canonical causal distribution $p_{\theta}(\bm{x}) = \prod_{i=1}^{T}p_{\theta}(\bm{x}_i | \bm{x}_{<i})$ with respect to each text token $\bm{x}_{i}$, where $\{\bm{x}_i\}_{i=1}^T$ denotes the sequence of text tokens and $T$ denotes the sequence length. Typical MLLMs~\cite{liu2023llava,liu2023improvedllava,an2025llavaonevision15fullyopenframework} adopt the cascade-style architecture. Specifically, the image $\bm{I}\in \mathbb{R}^{H\times W\times 3}$ ($H$ and $W$ denote the height and width respectively) is first encoded into visual features by a $\upxi$-parameterized visual encoder $\mathcal{G}_{\upxi}$ and then projected into vision tokens $\bm{V}=\{\bm{v}_1, \cdots, \bm{v}_N\}\in \mathbb{R}^{N\times D}$ by a $\phi$-parameterized connector $\mathcal{H}_{\phi}$, where $N$ denotes the length of vision token sequence and $D$ denotes the hidden dimension of the LLM. The canonical causal distribution for a multimodal sequence is formulated as:
\begin{equation}
    p_{\Theta}(\bm{x}) = \prod_{i=1}^{T}p_{\Theta}(\bm{x}_i | \bm{x}_{<i}, \bm{V}), \bm{V}=\mathcal{H}_{\phi}\circ\mathcal{G}_{\upxi}(\bm{I}),
\end{equation}
where $\Theta=\{\theta, \upxi, \phi\}$ denotes the parameters. The visual encoder could be pretrained vision models\cite{pmlr-v139-radford21a,Zhai_2023_ICCV,tschannen2025siglip2multilingualvisionlanguage,9709990,oquab2024dinov,siméoni2025dinov3,NEURIPS2023_06ea400b,fini2025multimodal,bai2025qwen25vltechnicalreport} or even a simple MLP which projects raw pixel patches into vision tokens directly~\cite{chen2024a,10.5555/3737916.3739581,lei2025sail}. 
% , which extracts visual features via a visual encoder $\mathcal{G}_{\upxi}$ (i.e., pretrained visual foundation models like CLIP~\cite{pmlr-v139-radford21a}, SigLIP~\cite{Zhai_2023_ICCV} and DINO~\cite{9709990}). The extract visual features are then 

\textbf{Training paradigm for MLLMs.} Typical MLLMs derive supervision by solely maximizing the log-likelihood of the text responses. The learning objective is formulated as Cross-Entropy loss over the text vocabulary as follows:
\begin{equation}
    \mathcal{L}_{\text{LM}}(\Theta;\bm{I}, \bm{x}) = - \frac{1}{T - P}\sum_{i=P+1}^{T}\log p_{\Theta}(\bm{x}_i | \bm{x}_{<i}, \bm{V}),
\end{equation}
where $P$ denotes prompt length (i.e., the prefix vision tokens and instruction tokens). Following LLaVA-OneVision 1.5~\cite{an2025llavaonevision15fullyopenframework}, we adopt a three-stage training receipt (Sec.~\ref{sec:expr_setup}).

% \textbf{Masked Image Modeling.} Masked Image Modeling (MIM) is a self-supervised learning approach where parts of an input image are masked, and the model is trained to reconstruct representations of the masked regions. Following iBOT~\cite{todo}, consider an input image $\bm{I}$, which is first divided into $N$ patches and encoded by a visual encoder $\mathcal{G}_{\upxi}$. Let $\mathcal{M} \in \{0,1\}^N$ denote the binary mask indicating the masked patches. iBOT utilizes a teacher-student framework to encourage the student to predict the teacher's representations of masked patches. The iBOT pretext task is formulated as minimizing the cross-entropy between the student outputs and the teacher targets for masked patches:
% \begin{equation}
%     \mathcal{L}_{\text{MIM}} = -\sum_{j \in \mathcal{P}_{\text{masked}}} \text{softmax}(\bm{z}^{\text{teacher}}_{j} / \tau) \cdot \log \text{softmax}(\bm{z}^{\text{student}}_{j}),
% \end{equation}
% where $\bm{z}^{\text{teacher}}_{j}$ and $\bm{z}^{\text{student}}_{j}$ denote the outputs for patch $j$ from the teacher and student, respectively, $\tau$ is a temperature parameter, and $\mathcal{P}_{\text{masked}}$ indexes the masked patches. This formulation drives the student model to learn semantic representations by reconstructing target representations for masked regions.

%% file: sec/4_method.tex
\section{Methodology}
\label{sec:method}

% In this section, we propose Laver, a novel MLLM training framework that produces more proper visual embeddings by jointly optimizing both language modeling and self-supervised visual objectives, eliminating the need for external embedding priors. 

This section elucidates how LaVer encourages MLLMs to learn discriminative visual representations, as depicted in Fig.~\ref{fig:method}. 
As shown in Fig.~\ref{fig:motivation}, \textit{progressive visual feature homogenization} indicates that MLLMs gradually discard meaningful visual semantics, utilizing only partial visual information for multimodal tasks.
To address this limitation, we force MLLMs to maintain visual structural information via masked image modeling, requiring the model to restore the missing visual information based on the visual context (Sec.~\ref{subsec:laver}). This process strengthens the model's intrinsic visual perception ability.
Nonetheless, models may exploit this objective by producing highly identical visual features for all vision tokens. Consequently, we introduce a regularizer to avert this shortcut (Sec.~\ref{subsec:inconsistency}).

\subsection{Latent Visual Reconstruction}
\label{subsec:laver}

% 1. method: latent visual reconstruction.

% 2. motivation: different from ROSS, which reconstructs the low-level pixels, we reconstruct the latent visual features in the LLM's semantic space.
% a. Pros 1: more alignment visual modeling and language modeling, as the model don't have to preserve the dense visual information in each layer of the model. Previous works also point out that different layers of MLLMs are responsible for different levels of visual information. \todo{Add references}
% b. Pros 2: more efficient training. Compared to ROSS, which trains an additional denoising network to reconstruct the raw pixels, we adopt a light-weight MLP as visual head to output the visual logits.

% 3. Technical details

\textbf{Masking.} We encode an image $\bm{I}$ into visual features $\bm{V} = \mathcal{H}_{\phi} \circ \mathcal{G}_{\upxi}(\bm{I}) \in \mathbb{R}^{N \times D}$. A binary mask $\mathcal{M} \in \{0,1\}^N$ is generated by randomly selecting a subset of vision positions at probability $r$ (mask ratio), where $\mathcal{M}_i = 1$ indicates the $i$-th token is masked. The masked vision tokens $\tilde{\bm{V}}$ are constructed by replacing the original features at masked positions with a learned mask token $\bm{e}_{\mathtt{[MASK]}} \in \mathbb{R}^D$: $\tilde{\bm{v}}_i = \mathcal{M}_i \cdot \bm{e}_{\mathtt{[MASK]}} + (1 - \mathcal{M}_i) \cdot \bm{v}_i$ for $i = 1, \ldots, N$. $\bm{e}_{\mathtt{[MASK]}}$ is learned along with the LLM's word embeddings.

\textbf{Vision head architecture.} We employ a 3-layer multilayer perceptron (MLP) as the vision head to project the visual output embeddings into visual logits, following~\cite{zhou2021ibot}. Let $\tilde{\bm{H}} = \{\tilde{\bm{h}}_1, \cdots, \tilde{\bm{h}}_N\} \in \mathbb{R}^{N\times D}$ denote the hidden state corresponding to the vision tokens $\tilde{\bm{V}}$, derived by $\tilde{\bm{H}} = \mathcal{F}_{\theta}(\tilde{\bm{V}})$. The vision head $\mathcal{V}_{\psi}$ with parameters $\psi$ transforms this representation  through the MLP with ReLU activations between layers. Formally, the transformation is defined as:
\begin{equation}
    \tilde{\bm{Z}} = \mathcal{V}_{\psi}(\tilde{\bm{H}}),
\end{equation}
where $\tilde{\bm{Z}} = \{\tilde{\bm{z}}_1, \cdots, \tilde{\bm{z}}_N\} \in \mathbb{R}^{N \times D_v}$ denotes the visual logits, and $D_v$ denotes the dimension of the visual logits.

\begin{figure*}
\centering
% \fbox{\rule{0pt}{2.5in} \rule{0.9\linewidth}{0pt}}
\begin{subfigure}{0.49\linewidth}
% \fbox{\rule{0pt}{1in} \rule{.9\linewidth}{0pt}}
\includegraphics[width=0.99\linewidth]{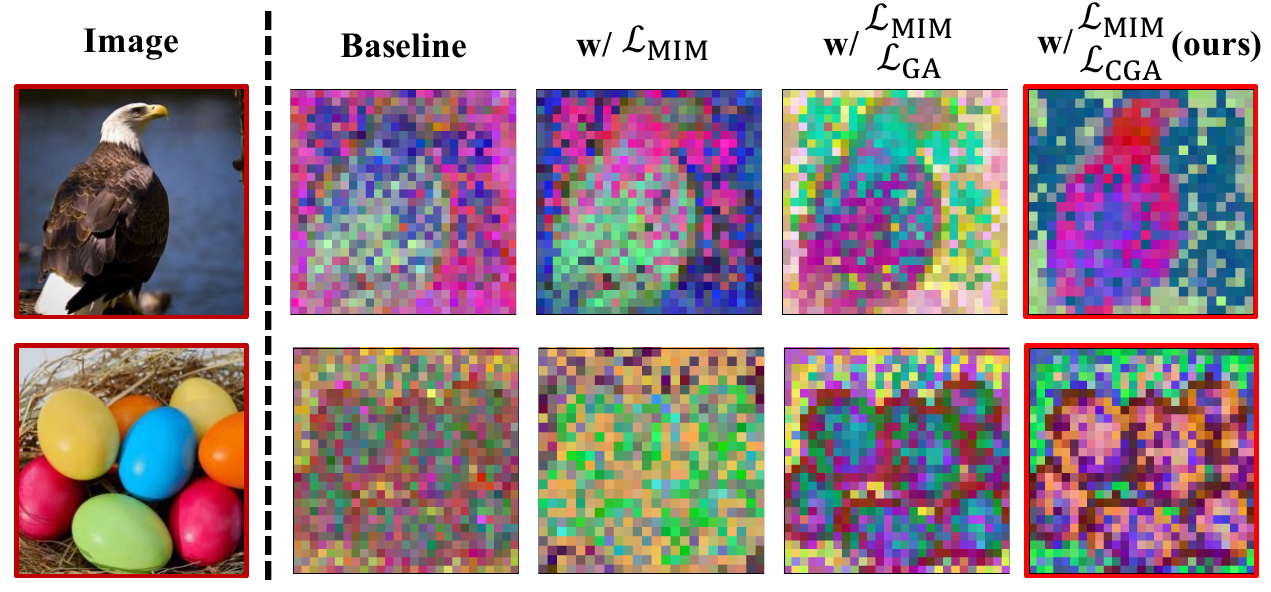}
\caption{PCA visualization of visual features.}
\label{fig:pca-a}
\end{subfigure}
\hfill
\begin{subfigure}{0.49\linewidth}
% \fbox{\rule{0pt}{1in} \rule{.95\linewidth}{0pt}}
\includegraphics[width=0.99\linewidth]{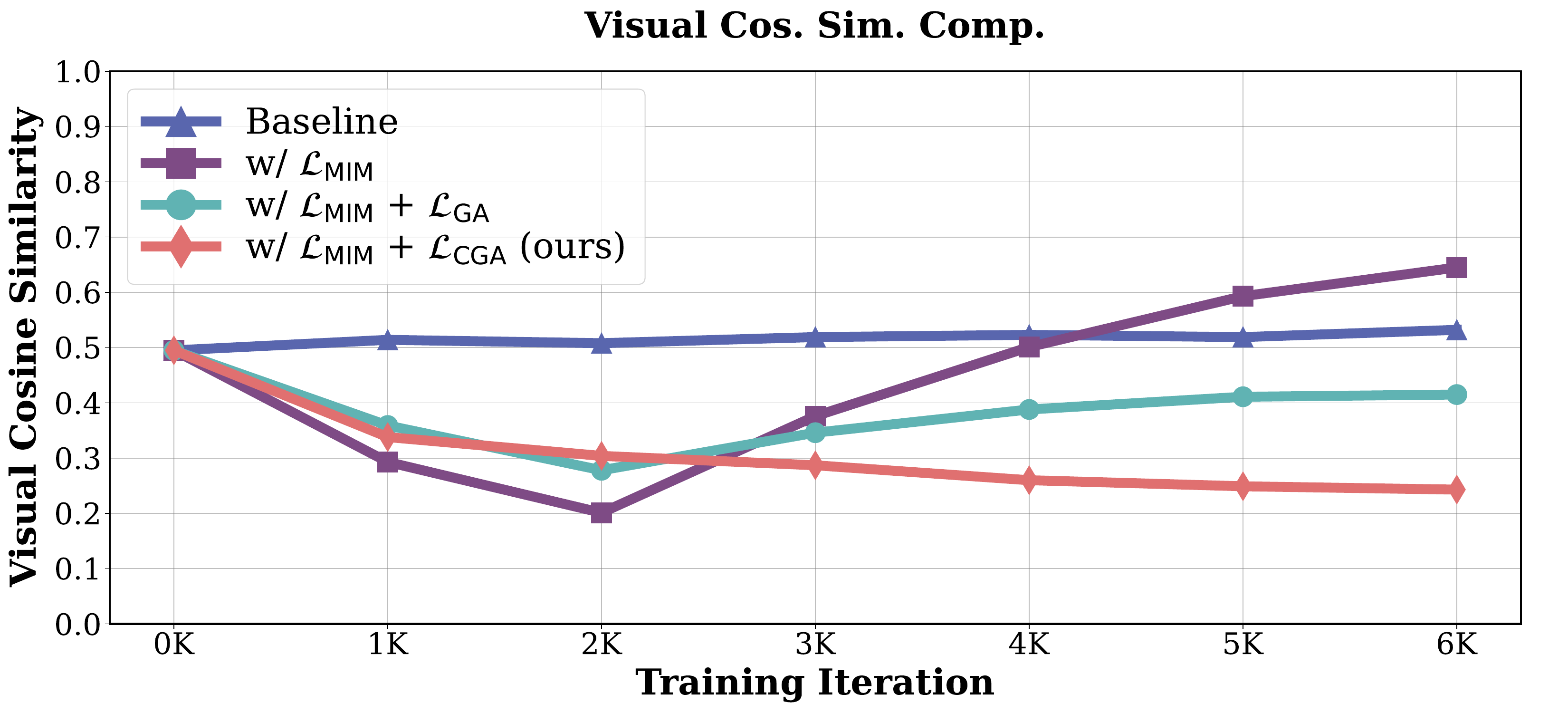}
\caption{Averaged cosine similarity between vision tokens.}
\label{fig:pca-b}
\end{subfigure}
\vspace{-8pt}
\caption{\textbf{Effects of MIM on visual feature consistency.} \textbf{(a)} illustrates PCA visualization of visual features with different components. \textbf{(b)} illustrates the averaged cosine similarity between vision tokens along training. Our method displays the most discrminative features.}
\label{fig:feature_inconsistency}
\vspace{-10pt}
\end{figure*}

\textbf{Training paradigm.} We employ a student-teacher framework where MLLMs are encouraged to predict the teacher's visual embeddings for the masked vision positions~\cite{zhou2021ibot}. The teacher model $\hat{\Phi} = \{\hat{\theta}, \hat{\psi}\}$ is maintained as an exponential moving average (EMA) of the student model $\Phi = \{\theta, \psi\}$ over training iterations. The teacher model processes the original unmasked vision tokens $\bm{V}$ to generate target visual logits $\hat{\bm{Z}} = \{\hat{\bm{z}}_1, \cdots, \hat{\bm{z}}_N\} \in \mathbb{R}^{N \times D_v}$ for the same positions, i.e., $\hat{\bm{Z}} = \mathcal{V}_{\hat{\psi}}\circ\mathcal{F}_{\hat{\theta}}(\bm{V})$. At updating step $t$, the teacher parameters are updated as:
% \begin{equation}
%     % \theta_{\text{teacher}}^{(t)} = \lambda \theta_{\text{teacher}}^{(t-1)} + (1 - \lambda) \theta_{\text{student}}^{(t)},
%     \hat{\theta}^{(t)} = \lambda \hat{\theta}^{(t-1)} + (1 - \lambda) \theta^{(t)},
%     \hat{\psi}^{(t)} = \lambda \hat{\psi}^{(t-1)} + (1 - \lambda) \psi^{(t)},
% \end{equation}
\begin{align}
    \begin{split}
    \hat{\theta}^{(t)} &= \lambda \hat{\theta}^{(t-1)} + (1 - \lambda) \theta^{(t)}, \\
    \hat{\psi}^{(t)} &= \lambda \hat{\psi}^{(t-1)} + (1 - \lambda) \psi^{(t)},
    \end{split}
\end{align}
where $\lambda \in [0,1]$ is the EMA decay rate.
% The student model processes the masked visual features $\tilde{\bm{v}}$ through the LLM backbone to obtain hidden states $\bm{h}_j$ for each vision token position $j$, which are then projected by the visual head to produce visual logits $\bm{z}_j^{\text{student}} = \mathcal{F}_{\psi}(\bm{h}_j)$.

Ideally, the student model $\Phi$ should be able to restore the masked vision tokens via the visual contextual information by leveraging the visual redundancy~\cite{9879206}, maintaining discriminative visual structural representations. Specifically, the distribution of the predicted visual logits $\tilde{\bm{Z}}$ should be close to the distribution of the target visual logits $\hat{\bm{Z}}$ that comes from clean representation process of the teacher $\hat{\Phi}$.

% The student model $\Phi$ is optimized to minimize the cross-entropy between its predictions $\tilde{\bm{z}}$ and the teacher's targets $\hat{\bm{z}}$ specifically for masked positions, encouraging the student to reconstruct the teacher's representations of the masked vision tokens.

\textbf{Learning objective.} The student model $\Phi$ is optimized to minimize the cross-entropy between its prediction $\tilde{\bm{z}}$ and the teacher's target $\hat{\bm{z}}$ for masked positions, encouraging the student to reconstruct the high-level visual structural information. Formally, the MIM loss function is formalized as:
\begin{equation}
\label{eq:mim_loss}
    \mathcal{L}_{\text{MIM}} = -\sum_{i \in \mathcal{P}_{\mathcal{M}}} \text{softmax}(\hat{\bm{z}}_i / \tau_{\text{tea.}}) \cdot \log \text{softmax}(\tilde{\bm{z}}_i / \tau_{\text{stu.}}),
\end{equation}
where $\mathcal{P}_{\mathcal{M}} = \{i \in \{1, \ldots, N\} \mid \mathcal{M}_i = 1\}$ denotes the set of indices corresponding to the masked positions, $\tau_{\text{tea.}} > 0$ and $\tau_{\text{stu.}} > 0$ are temperature scalers that control the sharpness of the softmax distribution for $\hat{\bm{z}}$ and $\tilde{\bm{z}}$, respectively. 
% This formulation employs the Kullback-Leibler divergence between the temperature-scaled softmax distributions of teacher and student outputs, encouraging the student to reconstruct the teacher's representation of masked visual patches.

\textbf{Improving Spatial Awareness.} MIM inherently requires the model to leverage spatial contextual information from neighboring regions to accurately reconstruct masked vision tokens~\cite{9879206,bao2022beitbertpretrainingimage,zhou2021ibot}, meaning that vision tokens need to attend to the entire image to effectively capture global spatial visual semantics~\cite{dosovitskiy2021imageworth16x16words,liu2021swintransformerhierarchicalvision}. Nonetheless, the standard causal attention mechanism~\cite{vaswani2023attentionneed,bai2023qwentechnicalreport} and the Rotary Position Embedding (RoPE)~\cite{10.1016/j.neucom.2023.127063,touvron2023llamaopenefficientfoundation}, which are designed primarily for sequential text processing, are fundamentally misaligned with the demands of visual modeling. Subsequently, we introduce the mixed attention mechanism~\cite{pmlr-v139-radford21a,Zhai_2023_ICCV,chen2025mocamodalityawarecontinualpretraining,lei2025sail}, which constructs bidirectional full attention for vision tokens and causal attention for text tokens (Fig.~\ref{fig:method-b}).
We also adopt 2D-RoPE~\cite{10.1007/978-3-031-72684-2_17,wang2024qwen2vlenhancingvisionlanguagemodels,bai2025qwen25vltechnicalreport} to better exploit the spatial structure of visual information by treating the image patch grid coordinates as positional index pairs for vision tokens (Fig.~\ref{fig:method-c}). The compatibility with the sequential processing of textual data is ensured by assigning identical row and column indices for text tokens.

\textbf{Independent Reconstruction.} 
To ensure independent visual reconstruction without interfering with the original learning paradigm, we leave the original multimodal sequence intact, and pack all the masked vision tokens to form a separate new sequence (Fig.~\ref{fig:method-a}). We adopt the diagonally blocked bidirectional attention and blocked 2D-RoPE to prevent information leakage between different images.

% 1. method: latent visual reconstruction.

% 2. motivation: different from ROSS, which reconstructs the low-level pixels, we reconstruct the latent visual features in the LLM's semantic space.
%     a. Pros 1: more alignment visual modeling and language modeling, as the model don't have to preserve the dense visual information in each layer of the model. Previous works also point out that different layers of MLLMs are responsible for different levels of visual information. \todo{Add references}
%     b. Pros 2: more efficient training. Compared to ROSS, which trains an additional denoising network to reconstruct the raw pixels, we adopt a light-weight MLP as visual head to output the visual logits.

% 3. Technical details

%     a. Masking: utilize a learnable mask embedding <mask> to substitute the encoded visual features, according to a random mask at certain mask ratio.
%     b. Visual head: a 3-layer MLP to project the visual output of the LLM backbone into the visual logits.
%     c. Training paradigm: student-teacher framework. Utilize an EMA-updated teacher model, and training the student model to predict the teacher's visual output for the masked visual tokens
%     d. Learning objective: cross-entropy loss over the visual logits.

\begin{table*}[t]
\centering
\caption{\textbf{Main results on various benchmarks with varied MLLM architectures.} (\%) We adopt the VLMEvalKit~\cite{10.1145/3664647.3685520} toolboxes to conduct the evaluation, following the official guideline to implement the missing benchmarks. LaVer consistently outperforms the baseline across most benchmarks, especially on benchmarks that require dense visual capabilities, including the OCR and vision-centric tasks.}
\vspace{-8pt}
\label{tab:main_results}
\resizebox{\linewidth}{!}{
\begin{tabular}{r|l| ccc ccc ccc| ccc ccc| ccc}
\toprule[1.5pt]
\multicolumn{2}{c|}{\multirow{3}{*}{\textbf{Benchmark}}}  & \multicolumn{9}{c|}{Fixed-Resolution Visual Encoder} & \multicolumn{6}{c|}{Native-Resolution Encoder} &  \multicolumn{3}{c}{Encoder-Free}\\
\cmidrule(rl){3-11} \cmidrule(lr){12-17} \cmidrule(lr){18-20}
\multicolumn{2}{c|}{} & \multicolumn{3}{c}{SigLIP2} & \multicolumn{3}{c}{CLIP} & \multicolumn{3}{c|}{DINOv2} & \multicolumn{3}{c}{AIMv2} & \multicolumn{3}{c|}{Qwen-ViT} & \multicolumn{3}{c}{MLP + Qwen 2.5}\\
\cmidrule(rl){3-5} \cmidrule(lr){6-8} \cmidrule(lr){9-11} \cmidrule(lr){12-14} \cmidrule(lr){15-17} \cmidrule(lr){18-20}
\multicolumn{2}{c|}{} & \textbf{Baseline} & \textbf{LaVer} & $\Delta_{\text{Baseline}}$ & \textbf{Baseline} & \textbf{LaVer} & $\Delta_{\text{Baseline}}$ & \textbf{Baseline} & \textbf{LaVer} & $\Delta_{\text{Baseline}}$ & \textbf{Baseline} & \textbf{LaVer} & $\Delta_{\text{Baseline}}$ & \textbf{Baseline} & \textbf{LaVer} & $\Delta_{\text{Baseline}}$ & \textbf{Baseline} & \textbf{LaVer} & $\Delta_{\text{Baseline}}$  \\
\midrule[1.0pt]
\multirow{7}{*}{\rotatebox[origin=c]{0}{\textbf{General VQA}}} & GQA~\cite{hudson2019gqanewdatasetrealworld} & 55.03 & \textbf{56.78} & \gain{1.75} & 51.51 & \textbf{54.77} & \gain{3.26} & 49.25 & \textbf{53.02} & \gain{3.77} & 53.02 & \textbf{57.04} & \gain{4.02} & \textbf{56.78} & \textbf{56.78} & \gain{0.00} & \textbf{33.17} & 32.91 & \loss{0.26} \\
& MMB$^\text{EN}$~\cite{10.1007/978-3-031-72658-3_13} & 73.97 & \textbf{75.60} & \gain{1.63} & 68.64 & \textbf{69.93} & \gain{1.29} & 59.19 & \textbf{61.68} & \gain{2.49} & 74.14 & \textbf{75.43} & \gain{1.29} & 77.66 & \textbf{79.04} & \gain{1.38} & 26.80 & \textbf{29.04} & \gain{2.24} \\
& SEED$^{\text{I}}$~\cite{li2023seedbenchbenchmarkingmultimodalllms} & 67.57 & \textbf{68.62} & \gain{1.05} & 64.36 & \textbf{65.20} & \gain{0.84} & 62.24 & \textbf{62.98} & \gain{0.74} & 66.96 & \textbf{67.04} & \gain{0.08} & \textbf{62.93} & 61.95 & \loss{0.98} & 40.62 & \textbf{43.70} & \gain{3.08} \\
& MME~\cite{fu2025mmecomprehensiveevaluationbenchmark} & 1510.73 & \textbf{1512.50} & \gain{1.77} & 1289.62 & \textbf{1474.65} & \gain{185.03} & 1166.51 & \textbf{1275.04} & \gain{108.53} & \textbf{1485.48} & 1446.67 & \loss{38.81} & 1556.08 & \textbf{1592.90} & \gain{36.82} & 1014.05 & \textbf{1138.67} & \gain{124.62} \\
& RWQA~\cite{xai2024grok15visionpreview} & 53.86 & \textbf{59.35} & \gain{5.49} & 54.25 & \textbf{56.47} & \gain{2.22} & 49.41 & \textbf{53.73} & \gain{4.32} & 56.21 & \textbf{57.78} & \gain{1.57} & 55.05 & \textbf{58.17} & \gain{3.12} & 44.18 & \textbf{48.76} & \gain{4.58} \\
& MMMU~\cite{yue2024mmmumassivemultidisciplinemultimodal} & 44.78 & \textbf{46.33} & \gain{1.55} & \textbf{44.56} & \textbf{44.56} & \gain{0.00} & 42.24 & \textbf{42.44} & \gain{0.20} & 44.56 & \textbf{45.00} & \gain{0.44} & 45.67 & \textbf{47.56} & \gain{1.89} & 39.78 & \textbf{40.67} & \gain{0.89} \\
& MM$^*$~\cite{chen2024rightwayevaluatinglarge} & 49.06 & \textbf{52.01} & \gain{2.95} & 43.17 & \textbf{45.45} & \gain{2.28} & 39.49 & \textbf{41.16} & \gain{1.67} & 47.52 & \textbf{49.80} & \gain{2.28} & 53.82 & \textbf{54.62} & \gain{0.80} & 30.99 & \textbf{31.99} & \gain{1.00} \\
\midrule
\multirow{4}{*}{\rotatebox[origin=c]{0}{\textbf{OCR VQA}}} & OCRB~\cite{Liu_2024} & 536 & \textbf{639} & \gain{103} & 306 & \textbf{365} & \gain{59} & 317 & \textbf{384} & \gain{67} & 399 & \textbf{412} & \gain{13} & 813 & \textbf{815} & \gain{2} & 144 & \textbf{153} & \gain{9} \\
& TVQA~\cite{singh2019vqamodelsread} & 67.87 & \textbf{69.82} & \gain{1.95} & 64.39 & \textbf{65.58} & \gain{1.19} & 60.08 & \textbf{61.34} & \gain{1.26} & 63.70 & \textbf{67.04} & \gain{3.34} & 62.87 & \textbf{69.89} & \gain{7.02} & 52.29 & \textbf{54.67} & \gain{2.38} \\
& CQA~\cite{masry-etal-2022-chartqa} & 62.06 & \textbf{63.93} & \gain{1.87} & 43.86 & \textbf{49.93} & \gain{6.07} & 41.91 & \textbf{42.47} & \gain{0.56} & 61.06 & \textbf{61.78} & \gain{0.72} & 72.01 & \textbf{74.55} & \gain{2.54} & 10.56 & \textbf{11.28} & \gain{0.72} \\
& AI2D~\cite{Kembhavi2016ADI} & 86.51 & \textbf{89.09} & \gain{2.58} & 81.61 & \textbf{83.30} & \gain{1.69} & 75.71 & \textbf{78.28} & \gain{2.57} & 86.02 & \textbf{87.46} & \gain{1.44} & 88.00 & \textbf{92.71} & \gain{4.71} & 74.17 & \textbf{76.28} & \gain{2.11} \\
\midrule
\multirow{2}{*}{\rotatebox[origin=c]{0}{\textbf{Vision-Centric}}} & MMVP~\cite{10655378} & 43.52 & \textbf{50.24} & \gain{6.72} & 27.36 & \textbf{39.36} & \gain{12.00} & 28.48 & \textbf{28.64} & \gain{0.16} & 35.28 & \textbf{37.68} & \gain{2.40} & 63.52 & \textbf{65.20} & \gain{1.68} & 18.08 & \textbf{19.04} & \gain{0.96} \\
& CV-B$^{\text{2D}}$~\cite{10.5555/3737916.3740687} & 52.20 & \textbf{55.60} & \gain{3.40} & 45.40 & \textbf{48.90} & \gain{3.50} & 42.10 & \textbf{44.90} & \gain{2.80} & 50.60 & \textbf{51.10} & \gain{0.50} & 59.90 & \textbf{61.60} & \gain{1.70} & 37.40 & \textbf{38.00} & \gain{0.60} \\
\midrule
\multirow{2}{*}{\rotatebox[origin=c]{0}{\textbf{Math \& Know.}}} & SQA~\cite{lu2022learnexplainmultimodalreasoning} & 73.74 & \textbf{75.55} & \gain{1.81} & 66.00 & \textbf{70.40} & \gain{4.40} & 65.67 & \textbf{65.87} & \gain{0.20} & 73.15 & \textbf{73.80} & \gain{0.65} & 76.36 & \textbf{77.53} & \gain{1.17} & 60.33 & \textbf{63.20} & \gain{2.87} \\
& MathV~\cite{lu2024mathvistaevaluatingmathematicalreasoning} & 56.05 & \textbf{58.04} & \gain{1.99} & 53.42 & \textbf{55.95} & \gain{2.53} & 47.73 & \textbf{48.26} & \gain{0.53} & 56.05 & \textbf{56.31} & \gain{0.26} & 57.10 & \textbf{62.36} & \gain{5.26} & \textbf{46.48} & 46.37 & \loss{0.11} \\
\midrule
\multirow{2}{*}{\rotatebox[origin=c]{0}{\textbf{Hallucination}}} & Hallu~\cite{guan2024hallusionbenchadvanceddiagnosticsuite} & 69.00 & \textbf{70.33} & \gain{1.33} & 60.00 & \textbf{64.00} & \gain{4.00} & 62.00 & \textbf{62.67} & \gain{0.67} & 74.33 & \textbf{75.00} & \gain{0.67} & 63.00 & \textbf{66.00} & \gain{3.00} & 52.00 & \textbf{53.00} & \gain{1.00} \\
& POPE~\cite{li-etal-2023-evaluating} & 90.90 & \textbf{91.23} & \gain{0.33} & \textbf{90.50} & 90.30 & \loss{0.20} & 88.30 & \textbf{88.53} & \gain{0.23} & 86.27 & \textbf{86.37} & \gain{0.10} & 89.10 & \textbf{89.57} & \gain{0.47} & 50.93 & \textbf{52.03} & \gain{1.10} \\
\midrule
\multicolumn{2}{c|}{Average} & 55.72 & \textbf{57.87} & \gain{2.15} & 50.58 & \textbf{53.24} & \gain{2.66} & 47.92 & \textbf{49.23} & \gain{1.31} & 54.70 & \textbf{55.86} & \gain{1.16} & 57.96 & \textbf{59.94} & \gain{1.99} & 36.37 & \textbf{37.74} & \gain{1.37} \\
\bottomrule[1.5pt]
\end{tabular}
}
\vspace{-8pt}
\end{table*}

\subsection{Mitigating Visual Feature Inconsistency}
\label{subsec:inconsistency}

\textbf{Visual Feature Inconsistency.} We observe that learning solely via MIM induces \textit{visual feature inconsistency}, wherein vision tokens exhibit high cosine similarities despite containing substantially different visual semantics, similar to observations in~\cite{oquab2024dinov,fan2025scalinglanguagefreevisualrepresentation,siméoni2025dinov3}. This phenomenon is illustrated in Fig.~\ref{fig:pca-a}, where visual features become entangled, leading to collapsed local visual structural information. As shown in Fig.~\ref{fig:pca-b}, the averaged cosine similarity during training reveals that MIM alone causes similarity to decrease rapidly in early stages but ultimately converges to values exceeding the baseline. This behavior indicates that without proper regularization, visual features can freely drift and hack the MIM loss. The underlying cause is that the loss function in Eqn.~(\ref{eq:mim_loss}) only enforces token-wise distribution matching between student and teacher for individual masked tokens, while failing to preserve the structural diversity across the entire set of vision tokens.

% \textbf{Visual Feature Inconsistency.} We observe that learning solely via MIM leads to \textit{visual feature inconsistency}, where vision tokens display high cosine similarities though containing significantly different visual semantic, similar to observations in~\cite{oquab2024dinov,fan2025scalinglanguagefreevisualrepresentation,siméoni2025dinov3}. The inconsistency is illustrated in Fig.~\ref{fig:pca-a}, where visual features are mixed-up, resulting in collapsed local visual strutural information. We present the averaged cosine similarity along the training process in Fig.~\ref{fig:pca-b} and find that singularly applying MIM will make the cosine similarity goes down rapidly but ends up high similarity even larger than the baseline, indicating that without regularization, the visual features can move freely and quickly hack the MIM loss. This inconsistency occurs because the loss function in Eqn.~(\ref{eq:mim_loss}) only constrains the student model to match the teacher's distribution for individual masked tokens without maintaining the structural diversity across different vision tokens.
% Consequently, the model can achieve low loss values by producing homogeneous representations $\bm{z}^{\text{student}}_{j} = \bm{c}$ for all $j$, where $\bm{c} \in \mathbb{R}^{D_v}$ is a constant vector that approximates the average teacher distribution. This degeneracy undermines the representational capacity of the visual features and prevents the model from learning discriminative visual semantics.

\textbf{Gram-Anchoring.} To mitigate visual feature inconsistency, we explore the \textit{Gram-Anchoring} (GA) mechanism~\cite{siméoni2025dinov3} that enforces the relative structure of visual features from student to match the teacher's while making the features free to move. Specifically, the Gram matrices $\bm{G} \in \mathbb{R}^{N \times N}$ for visual logits $\bm{Z}$ is calculated as follows:
\begin{equation}
    \bm{G}(\bm{Z}) = \text{Norm}(\bm{Z}) \cdot \text{Norm}(\bm{Z})^\top, \bm{Z} \in \mathbb{R}^{N \times D_v},
\end{equation}
where $\text{Norm}(\cdot) = \nicefrac{\cdot}{\|\cdot\|}$ denotes the $\text{L}_2$-normalization along the feature dimension. The GA loss is then formulated as:
\begin{equation}
    \label{eq:ga_loss}
    \mathcal{L}_{\text{GA}} = \Vert \bm{G}(\tilde{\bm{Z}}) - \bm{G}(\hat{\bm{Z}}) \Vert_F^2.
\end{equation}
% The Gram-Anchoring enforces the relative structure of visual representations from student to match the teacher's.

% The Gram matrix elements are defined as:
% For both teacher and student models, we compute the Gram matrix $\bm{G} \in \mathbb{R}^{N \times N}$ that captures the pairwise cosine similarities between all visual logits. Specifically, for a set of visual logits $\{\bm{z}_j\}_{j=1}^N$, the Gram matrix elements are defined as:
% \begin{equation}
%     \bm{G}_{jk} = \frac{\bm{z}_j \cdot \bm{z}_k}{\|\bm{z}_j\| \|\bm{z}_k\|}, \quad \text{for } j,k = 1,\ldots,N,
% \end{equation}
% where $\cdot$ denotes the dot product and $\|\cdot\|$ denotes the L2 norm. We compute separate Gram matrices $\bm{G}^{\text{teacher}}$ and $\bm{G}^{\text{student}}$ for the teacher and student outputs respectively. The gram-anchoring loss is then formulated as the mean squared error (MSE) between these two Gram matrices:
% \begin{equation}
%     \mathcal{L}_{\text{gram}} = \frac{1}{N^2} \sum_{j=1}^{N} \sum_{k=1}^{N} \left( \bm{G}^{\text{teacher}}_{jk} - \bm{G}^{\text{student}}_{jk} \right)^2.
% \end{equation}
% This loss function enforces that the relative similarity structure among visual tokens in the student's output matches that of the teacher's, thereby preventing feature inconsistency by maintaining the diversity and discriminative relationships between different visual representations.

\textbf{Clipped Gram-Anchoring.} While GA effectively mitigates visual feature inconsistency through structural similarity preservation, it exhibits limited capacity for learning discriminative representations. As evidenced in Fig.~\ref{fig:pca-b}, cosine similarity initially decreases but gradually increases during training, revealing that GA inadequately penalizes feature inconsistency. This limitation stems from the symmetric nature of Eqn.~(\ref{eq:ga_loss}), which equally penalizes all deviations between teacher and student Gram matrices irrespective of deviation direction. 
Critically, when $\bm{G}(\tilde{\bm{Z}}) < \bm{G}(\hat{\bm{Z}})$, indicating the student produces more discriminative features than the teacher, Eqn.~(\ref{eq:ga_loss}) inadvertently suppresses this desirable behavior. We therefore introduce asymmetric \textit{Clipped Gram-Anchoring} (CGA):
\begin{equation}
    \label{eq:cga_loss}
    \mathcal{L}_{\text{CGA}} = \Vert \text{Clip}(\bm{G}(\tilde{\bm{Z}}) - \bm{G}(\hat{\bm{Z}})) \Vert_F^2,
\end{equation}
where $\text{Clip}(\cdot) = \max(0, \cdot)$ denotes element-wise clipping.
CGA selectively penalizes only undesirable deviations where student features become less discriminative than the teacher, thereby encouraging the development of superior representations.
CGA consistently yields discriminative features with decreasing cosine similarity (Fig.~\ref{fig:pca-b}).

\textbf{Final objective.} The objective for our LaVer is to jointly minimize the sum of the LM, the MIM, and the CGA loss:
\begin{equation}
    \mathcal{L}_{\text{Laver}} = \mathcal{L}_{LM} + \omega_{\text{MIM}}\mathcal{L}_{\text{MIM}} + \omega_{\text{CGA}}\mathcal{L}_{\text{CGA}},
\end{equation}
where $\omega_{\text{MIM}}$ and $\omega_{\text{CGA}}$ are trade-off parameters default to 1.0. This formulation integrates the MIM loss for visual reconstruction and the clipped gram-anchoring loss for feature diversity preservation, ensuring that the model learns both discriminative visual representations and semantic consistency across different vision tokens.

%% file: sec/5_expr.tex
\begin{table*}
\centering
\caption{\textbf{Performance on complex visual task and ablation studies on spatial awareness and loss components.} (\%) \textbf{(a)} Performance on complex visual task. \textbf{(b)} Ablation study on spatial awareness. \textbf{(c)} Ablation study on loss components.}
\vspace{-8pt}
\label{tab:ablations_combined}
\resizebox{0.265\linewidth}{!}{
\begin{tabular}{c|c|ccc}
\toprule[1.5pt]
\multicolumn{5}{c}{\textbf{(a) Performance on complex visual task}} \\
\midrule[1.0pt]
\multicolumn{2}{c|}{\multirow{2}{*}{\rotatebox[origin=c]{0}{\textbf{Methods}}}} & \multicolumn{3}{c}{ReasonSeg} \\
\cmidrule(lr){3-5} 
\multicolumn{2}{c|}{} & Short & Long & Overall \\
\midrule[1.0pt]
\multirow{2}{*}{SigLIP 2} & Baseline & 38.25 & 52.76 & 51.53  \\
& LaVer & \textbf{39.21} & \textbf{53.63} & \textbf{52.89} \\
\midrule[1.0pt]
\multirow{2}{*}{CLIP} & Baseline & 35.07 & 50.80 & 48.26  \\
& LaVer & \textbf{37.98} & \textbf{52.11} & \textbf{49.43} \\
\bottomrule[1.5pt]
\end{tabular}
}
\hspace{0.01\linewidth}
\resizebox{0.33\linewidth}{!}{
\begin{tabular}{ccc|cc}
\toprule[1.5pt]
\multicolumn{5}{c}{\textbf{(b) Ablation study on spatial awareness}} \\
\midrule[1.0pt]
    LaVer & Mixed Attn. & 2D-RoPE & SigLIP 2 & CLIP \\
\midrule[1.0pt]
\textcolor{Grey}{\ding{55}} &  \textcolor{Grey}{\ding{55}} & \textcolor{Grey}{\ding{55}} &  55.72 & 50.58\\
\textcolor{Grey}{\ding{55}} & \ding{51} & \textcolor{Grey}{\ding{55}} & 56.78 & 51.40 \\
\textcolor{Grey}{\ding{55}} & \textcolor{Grey}{\ding{55}} & \ding{51} & 55.57 & 50.59 \\
\textcolor{Grey}{\ding{55}} & \ding{51} & \ding{51} & 56.43 & 51.99 \\
\ding{51} & \ding{51} & \ding{51} & \textbf{57.87} & \textbf{53.24} \\
\bottomrule[1.5pt]
\end{tabular}
}
\hspace{0.01\linewidth}
\resizebox{0.355\linewidth}{!}{
\begin{tabular}{ccc|cc}
\toprule[1.5pt]
\multicolumn{5}{c}{\textbf{(c) Ablation study on loss components}} \\
\midrule[1.0pt]
w/ $\mathcal{L}_{\text{MIM}}$ & w/ $\mathcal{L}_{\text{GA}}$ & w/ $\mathcal{L}_{\text{CGA}}$ & SigLIP 2 & CLIP \\
\midrule[1.0pt]
\textcolor{Grey}{\ding{55}}  & \textcolor{Grey}{\ding{55}}  & \textcolor{Grey}{\ding{55}} & 55.72 & 50.58\\
\ding{51} & \textcolor{Grey}{\ding{55}}  & \textcolor{Grey}{\ding{55}}  & 53.76 & 49.71 \\
\ding{51} & \ding{51} & \textcolor{Grey}{\ding{55}} & 56.46 & 52.01 \\
\ding{51} & \textcolor{Grey}{\ding{55}}  & \ding{51} & \textbf{57.87} & \textbf{53.24} \\
\bottomrule[1.5pt]
\end{tabular}
}

\end{table*}
    
\begin{figure*}
\centering
\begin{subfigure}{0.241\linewidth}
    % \fbox{\rule{0pt}{2.25in} \rule{.9\linewidth}{0pt}}
    \includegraphics[width=0.99\linewidth]{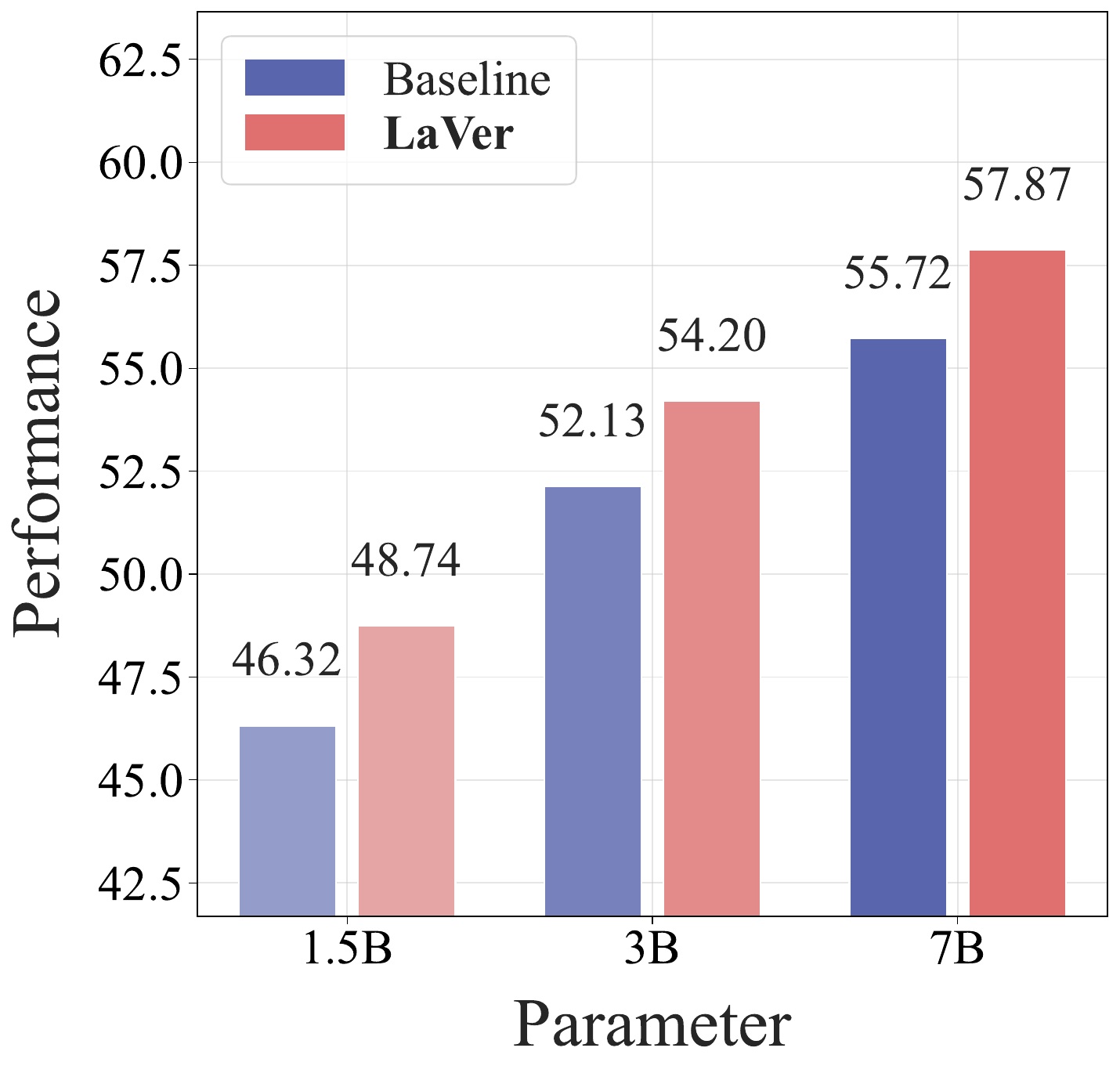}
    \caption{Parameter Scaling of LaVer.}
    \label{fig:abla-a}
\end{subfigure}
\hfill
\begin{subfigure}{0.232\linewidth}
    % \fbox{\rule{0pt}{2.25in} \rule{.9\linewidth}{0pt}}
    \includegraphics[width=0.99\linewidth]{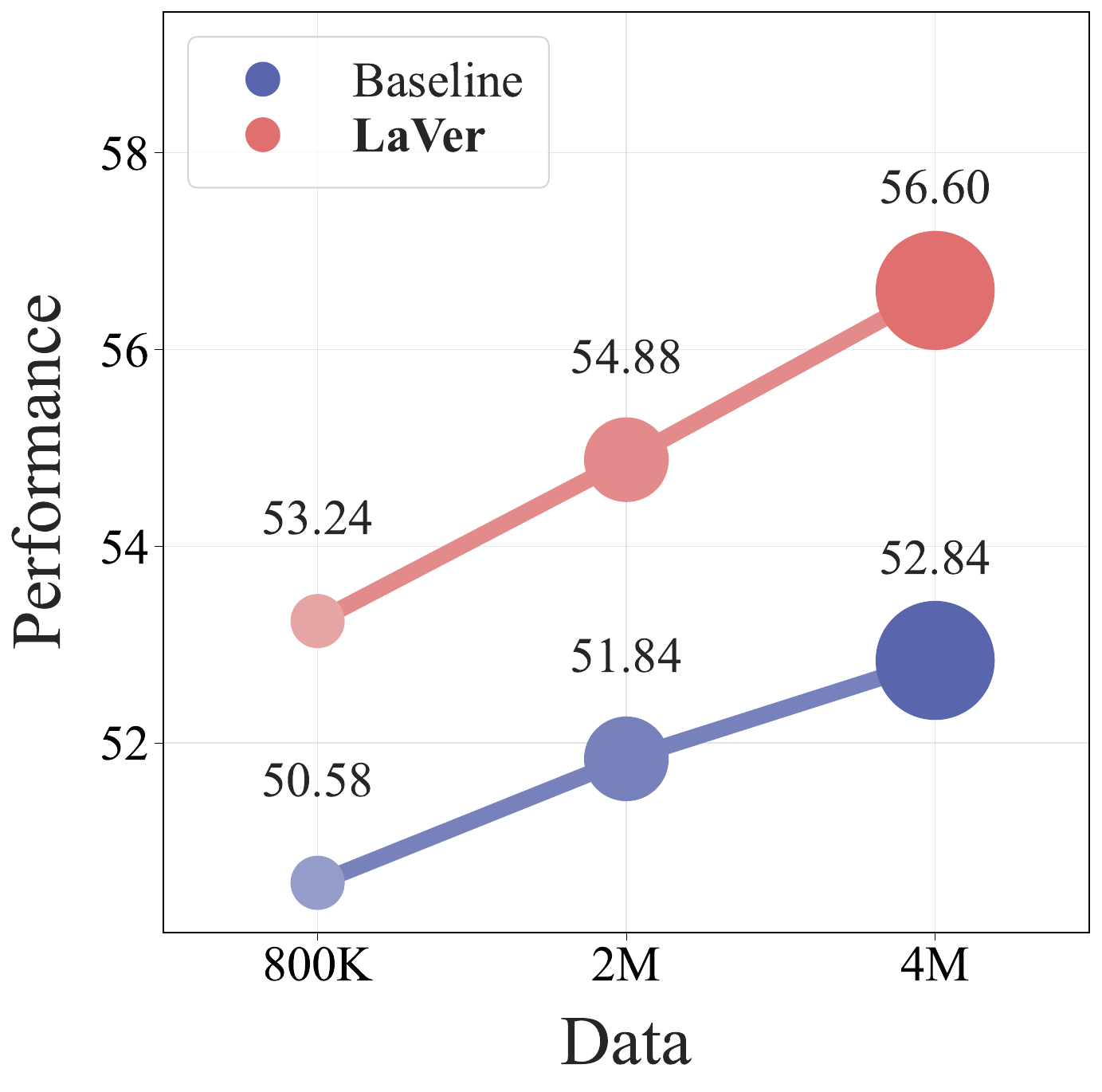}
    \caption{Data Scaling of LaVer.}
    \label{fig:abla-b}
\end{subfigure}
\hfill
\begin{subfigure}{0.234\linewidth}
    % \fbox{\rule{0pt}{2.25in} \rule{.9\linewidth}{0pt}}
    \includegraphics[width=0.99\linewidth]{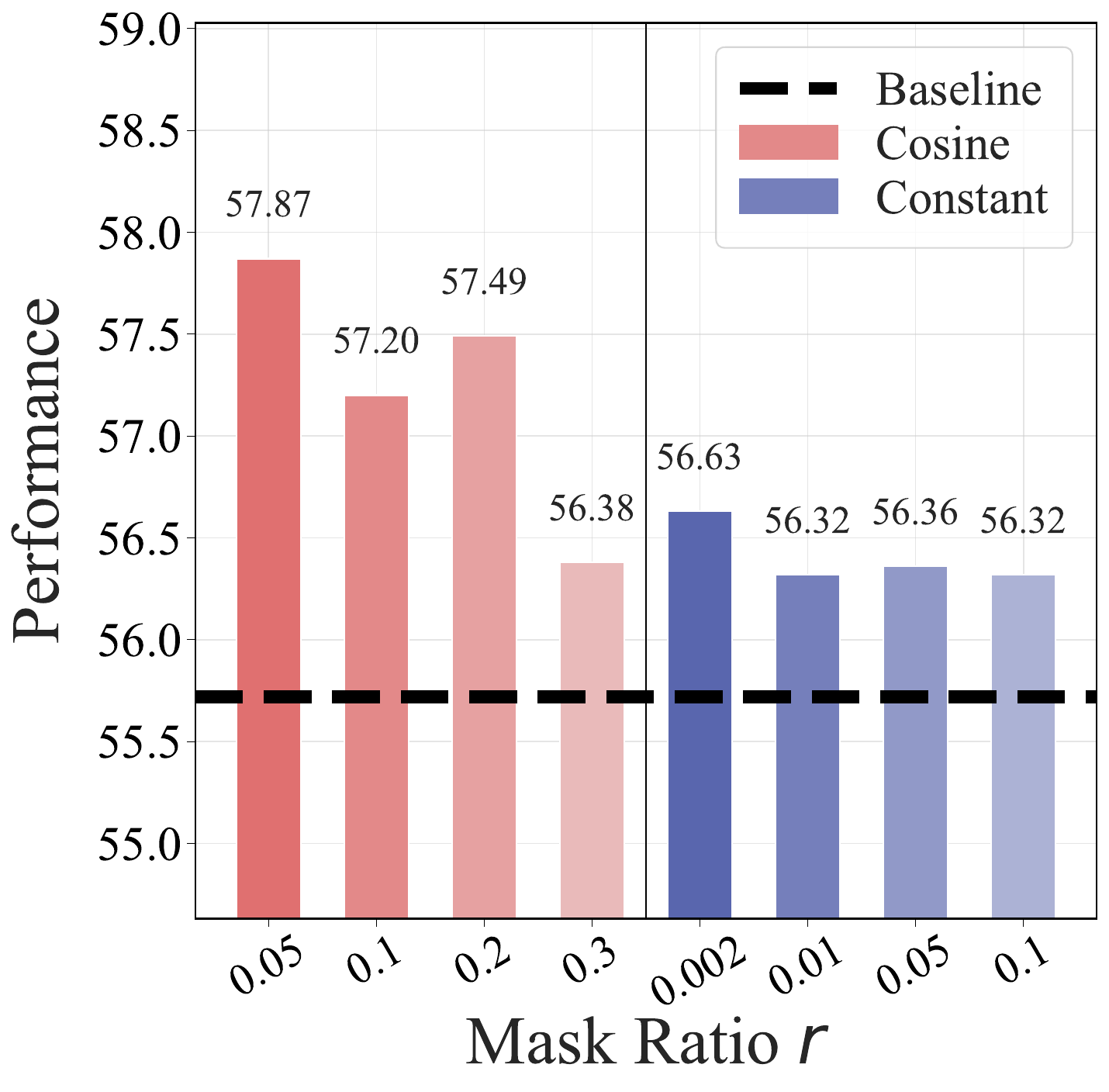}
    \caption{Ablation on masking strategies.}
    \label{fig:abla-c}
\end{subfigure}
\hfill
\begin{subfigure}{0.23\linewidth}
    % \fbox{\rule{0pt}{2.25in} \rule{.9\linewidth}{0pt}}
    \includegraphics[width=0.99\linewidth]{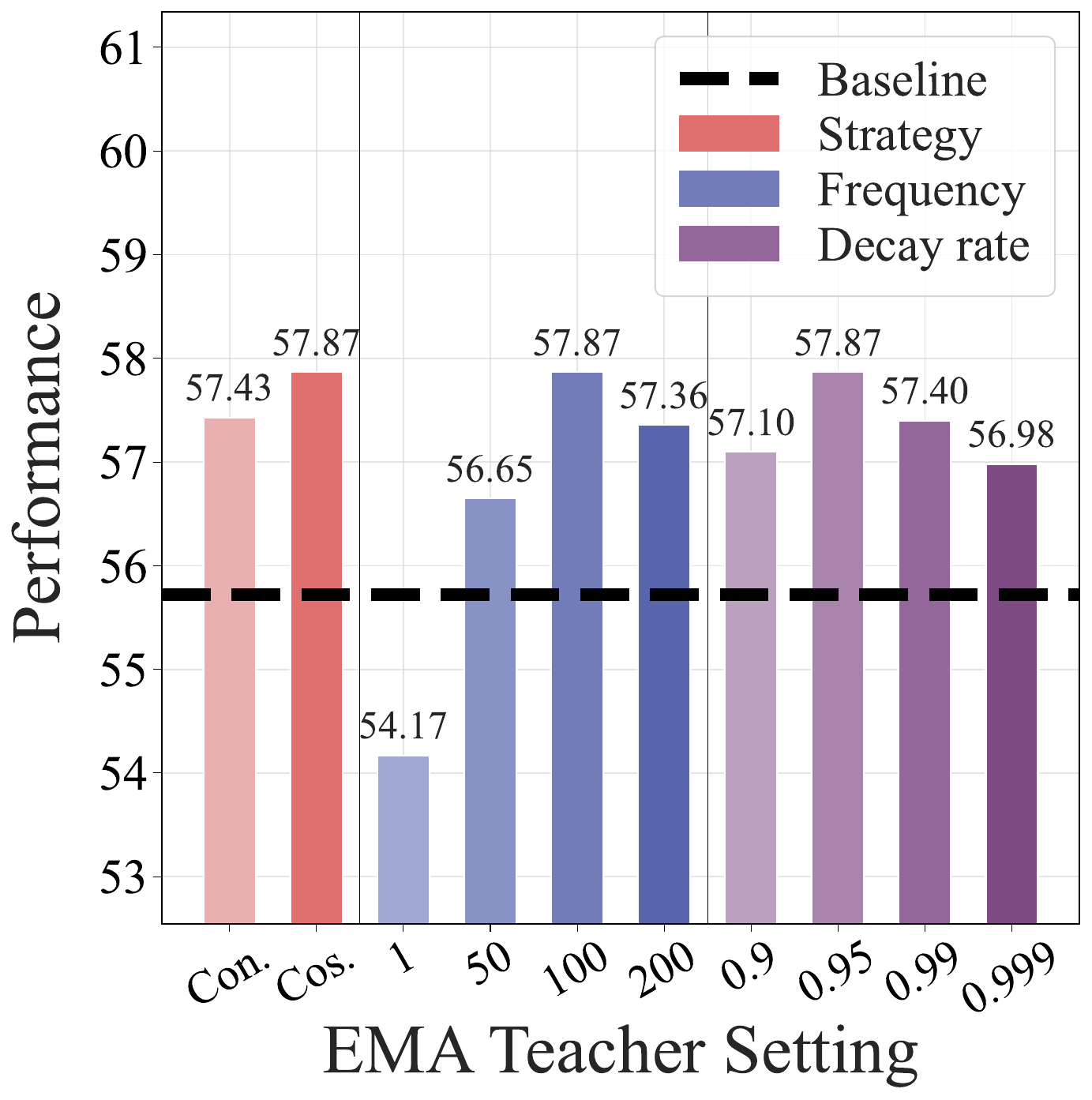}
    \caption{Ablation on EMA strategies.}
    \label{fig:abla-d}
\end{subfigure}
\vspace{-8pt}
\caption{\textbf{Scaling properties \& Ablation studies.} \textbf{(a)} Parameter Scaling of LaVer. \textbf{(b)} Data Scaling of LaVer. \textbf{(c)} Ablation on masking strategies. \textbf{(d)} Ablation on EMA updating strategies. LaVer displays significant scaling properties and robustness of hyperparameters.}
\label{fig:abla}
\vspace{-10pt}
\end{figure*}

\section{Experiments}
\label{sec:expr}

\subsection{Experimental Setup}
\label{sec:expr_setup}

\textbf{Implementation Details.} We employ Qwen2.5-7B-Instruct~\cite{qwen2025qwen25technicalreport} as the language model backbone.
To comprehensively evaluate LaVer across diverse visual encoding paradigms, we experiment with multiple vision encoders: fixed-resolution encoders (SigLIP 2-ViT-SO400M/14@384~\cite{tschannen2025siglip2multilingualvisionlanguage}, CLIP-ViT-L/14@336~\cite{pmlr-v139-radford21a}, DINOv2-Large/14@224~\cite{oquab2024dinov}), native-resolution encoders (AIMv2-Large/14~\cite{fini2025multimodal}, Qwen-ViT from Qwen2.5-VL-7B-Instruct~\cite{bai2025qwen25vltechnicalreport}), and an encoder-free architecture employing a single MLP to project raw pixels directly into vision tokens~\cite{chen2024a,lei2025sail}. All experiments are conducted on 16 NVIDIA A100 (80GB) GPUs. Additional implementation details are provided in the supplementary material.

% \textbf{Implementation Details.} We employ Qwen 2.5 Instruct~\cite{qwen2025qwen25technicalreport} as our language model backbone, leveraging its strong capabilities in language tasks.
% We adopt various vision encoders to comprehensively evaluate our approach across different visual encoding paradigms.
% Specifically, we experiment with fixed-resolution encoders including SigLIP 2-ViT-SO400M/14@384~\cite{tschannen2025siglip2multilingualvisionlanguage}, CLIP-ViT-L/14@336~\cite{pmlr-v139-radford21a}, and DINOv2-Large/14@224~\cite{oquab2024dinov}, native-resolution encoders such as AIMv2-Large/14~\cite{fini2025multimodal} and Qwen-ViT from Qwen2.5-VL-7B-Instruct~\cite{bai2025qwen25vltechnicalreport}, and an encoder-free architecture that utilizes a single MLP to project raw pixels into vision tokens~\cite{chen2024a,lei2025sail}. More details are provided in the supplementary material.

\textbf{Training Recipe.} Following LLaVA-OneVision 1.5~\cite{an2025llavaonevision15fullyopenframework}, we adopt a three-stage training procedure to progressively develop multimodal capabilities.
Stage 1 performs connector initialization for vision-language alignment using LLaVA-558K~\cite{liu2023llava}.
Stage 2 applies LaVer for intrinsic visual modeling and knowledge injection, learning discriminative representations from 800K pairs subsampled from FineVision 23M~\cite{wiedmann2025finevisionopendataneed}.
Stage 3 conducts visual instruction tuning~\cite{liu2023llava} using 800K samples from LLaVA-OneVision 4M~\cite{li2025llavaonevision}. More settings are in the supplementary material.

\textbf{Evaluation.} We conduct comprehensive evaluations across diverse multimodal understanding benchmarks using VLMEvalKit~\cite{10.1145/3664647.3685520}.
Our evaluation suite encompasses (1) general VQA benchmarks: GQA~\cite{hudson2019gqanewdatasetrealworld}, MMBench (MMB$^{\text{EN}}$)~\cite{10.1007/978-3-031-72658-3_13}, SEED-Image (SEED$^{\text{I}}$)~\cite{li2023seedbenchbenchmarkingmultimodalllms}, MME~\cite{fu2025mmecomprehensiveevaluationbenchmark}, RealWorldQA (RWQA)~\cite{xai2024grok15visionpreview}, MMMU~\cite{yue2024mmmumassivemultidisciplinemultimodal}, and MMStar (MM$^*$)~\cite{chen2024rightwayevaluatinglarge}; (2) OCR benchmarks: OCR-Bench (OCRB)~\cite{Liu_2024}, TextVQA (TVQA)~\cite{singh2019vqamodelsread}, ChartQA (CQA)~\cite{masry-etal-2022-chartqa}, and AI2D~\cite{Kembhavi2016ADI}; (3) vision-centric benchmarks: CV-Bench-2D (CV-B$^{\text{2D}}$)~\cite{10.5555/3737916.3740687} and MMVP~\cite{10655378}; (4) knowledge and math benchmarks: ScienceQA (SQA)~\cite{lu2022learnexplainmultimodalreasoning} and MathVista (MathV)~\cite{lu2024mathvistaevaluatingmathematicalreasoning}; and (5) hallucination benchmarks: HallucinationBench (Hallu)~\cite{guan2024hallusionbenchadvanceddiagnosticsuite} and POPE~\cite{li-etal-2023-evaluating}. Average scores are computed across all benchmarks, with MME~\cite{fu2025mmecomprehensiveevaluationbenchmark} and OCRB~\cite{Liu_2024} normalized to $[0.0, 1.0]$.
Evaluation details are in the supplementary material.

\subsection{Main Results}
\label{sec:expr_main_results}

\textbf{Consistent improvements across diverse vision encoders.} Table~\ref{tab:main_results} demonstrates that LaVer consistently outperforms baselines across nearly all benchmarks, with particularly pronounced gains on tasks requiring dense visual information comprehension, such as OCR~\cite{Liu_2024,masry-etal-2022-chartqa} and vision-centric benchmarks~\cite{10655378,10.5555/3737916.3740687}.
With SigLIP 2, LaVer achieves remarkable improvements of \textbf{103} points (\textbf{19.22\%}) on OCR-Bench~\cite{Liu_2024} and \textbf{6.72\%} on MMVP~\cite{10655378}.
CLIP exhibits similar substantial gains: \textbf{6.07\%} on ChartQA~\cite{masry-etal-2022-chartqa} and \textbf{12.00\%} on MMVP.
Native-resolution encoders also demonstrate significant enhancements, with AIMv2 and Qwen-ViT achieving \textbf{3.34\%} and \textbf{7.02\%} improvements on TextVQA~\cite{singh2019vqamodelsread}, respectively.
Even the encoder-free architecture, i.e., employing a single MLP to project raw pixels into visual tokens, yields consistent \textbf{1.37\%} overall gains.
These improvements across diverse architectures validate LaVer's effectiveness in enhancing discriminative visual representations within the model's latent space, without imposing architectural constraints.
Additional results are provided in the supplementary material.

\textbf{Effectiveness on complex visual reasoning tasks.} To further assess LaVer's capabilities, we examine its performance on \textit{Reasoning Segmentation} (ReasonSeg)~\cite{lai2024lisareasoningsegmentationlarge}, a challenging task requiring MLLMs to amalgamate language reasoning with integrated visual perception.
In accordance with~\cite{tang2025ufounifiedapproachfinegrained}, we initialize both baseline and LaVer models from stage-2 checkpoints and fine-tune them on segmentation datasets~\cite{yu2016modelingcontextreferringexpressions,caesar2018cocostuffthingstuffclasses,caesar2020nuscenesmultimodaldatasetautonomous,8100027,8237796} combined with LLaVA-665K~\cite{liu2023improvedllava}.
Zero-shot evaluation on the test set~\cite{lai2024lisareasoningsegmentationlarge,tang2025ufounifiedapproachfinegrained} reveals that LaVer-initialized model consistently outperforms its baseline counterpart, achieving improvements of \textbf{1.36 \%} and \textbf{1.17 \%} on gIoU with SigLIP 2 and CLIP, respectively (Table~\ref{tab:ablations_combined}(a)).
These gains demonstrate LaVer's power to integrate visual perception with language reasoning, particularly benefiting tasks requiring sophisticated cross-modal coordination.
See details in supplementary material.

\textbf{Scalability of LaVer.} We investigate LaVer's scaling properties across both model and data dimensions.
For parameter scaling, we evaluate Qwen2.5-Instruct~\cite{qwen2025qwen25technicalreport} models at 1.5B, 3B, and 7B parameters with SigLIP 2 and CLIP, maintaining consistent training data as specified in Sec.~\ref{sec:expr_setup}. As shown in Fig.~\ref{fig:abla-a}, LaVer consistently outperforms the baseline across all model sizes with SigLIP 2, demonstrating robust scalability. 
For data scaling, we vary stage-2 training volume where LaVer is applied, while fixing stages 1 and 3, uniformly subsampling 800K, 2M, and 4M samples from FineVision 23M~\cite{wiedmann2025finevisionopendataneed} and training for one epoch at each scale using Qwen2.5-7B-Instruct with SigLIP 2 and CLIP. Fig.~\ref{fig:abla-b} presents averaged results with CLIP, revealing that LaVer maintains substantial gains across all data scales, confirming its effectiveness under diverse training regimes. Complete results appear in the supplementary material.

% \textbf{Scalability of LaVer.} To investigate the scaling properties of LaVer, we evaluate its performance relative to the baseline across varying model and data scales.
% For parameter scaling, we employ the Qwen2.5 series~\cite{qwen2025qwen25technicalreport} with model sizes of 1.5B, 3B, and 7B parameters, using SigLIP 2 and CLIP as vision encoders while maintaining the training data specified in Sec.~\ref{sec:expr_setup}. Figure~\ref{fig:abla-a} presents the averaged results with SigLIP 2, demonstrating that LaVer consistently delivers substantial improvements over the baseline across all model sizes. Complete results are provided in the supplementary material.
% For data scaling, we vary the training data volume in stage-2 where LaVer is applied, while maintaining fixed data for stages 1 and 3. Using Qwen2.5-7B-Instruct as the language backbone with SigLIP 2 and CLIP as vision encoders, we uniformly subsample 800K, 2M, and 4M samples from FineVision 23M~\cite{wiedmann2025finevisionopendataneed}, training for one epoch at each scale. Figure~\ref{fig:abla-b} presents the averaged results with CLIP, revealing that LaVer consistently yields significant improvements across all data scales.

\subsection{Ablation Studies}
\label{sec:expr_ablation_studies}

\textbf{Ablation on masking strategies.} We examine two masking schedules: \textit{constant}, fixing $r$ throughout training, and \textit{cosine}, gradually increasing $r$ from 0.0 to the target value following a cosine schedule. Fig.~\ref{fig:abla-c} presents results with SigLIP 2. We find that even modest mask ratios (e.g., $r=0.05$) yield substantial improvements, while higher ratios under cosine scheduling degrade performance, suggesting that higher ratios may require longer training process to converge effectively, as MIM-based models~\cite{9709990,oquab2024dinov,siméoni2025dinov3} typically require extensive pretraining to manifest emergent properties. The cosine schedule consistently outperforms constant scheduling, as the latter either limits reconstruction capacity or imposes excessive cold-start with fixed $r$. Notably, LaVer surpasses the baseline across all configurations, demonstrating robustness to masking strategies.
%Complete results appear in the supplementary material.

% \textbf{Ablation on masking strategies.} As a core component of LaVer, we investigate two schedules towards mask ratio $r$: \textit{constant}, which maintains a fixed mask ratio throughout training, and \textit{cosine}, which gradually increases the mask ratio from 0.0 to the target value following a cosine schedule. Figure~\ref{fig:abla-c} presents results using SigLIP 2 as the vision encoder. We observe that even modest mask ratios yield substantial improvements, while performance degrades with higher ratios under the cosine schedule. This degradation likely stems from insufficient training iterations, as prior MIM-based models~\cite{9709990,oquab2024dinov,siméoni2025dinov3} typically require extensive image data to manifest emergent properties. The cosine schedule generally outperforms the constant schedule, potentially because the latter provides inadequate task difficulty progression, i.e., either limiting the model's capacity to reconstruct masked tokens or imposing excessive reconstruction difficulty at early iterations. Despite variations across schedules and ratios, LaVer consistently surpasses the baseline across all configurations. Full results are found in the supplementary material.

\textbf{Ablation on EMA teacher updating strategies.} We examine EMA updating strategies using SigLIP 2 in Fig.~\ref{fig:abla-d}. We compare two decay schedules: \textit{constant}, fixing $\lambda$, and \textit{cosine}, gradually increasing $\lambda$ to 1.0. The cosine schedule marginally outperforms constant scheduling, with both substantially surpassing the baseline, confirming robustness across strategies. For update frequency, excessively frequent updates (every step) degrade performance, likely because rapid teacher updates destabilize learning by propagating feature inconsistencies inadvertently. Optimal performance emerges at approximately 100-step intervals (0.02 frequency over 6K iterations). Varying decay rates reveals $\lambda=0.95$ as optimal, with all configurations outperforming the baseline. Full results are in the supplementary material.

\begin{figure}[t]
\centering
% \fbox{\rule{0pt}{2.5in} \rule{0.9\linewidth}{0pt}}
    %\includegraphics[width=0.8\linewidth]{egfigure.eps}
    \includegraphics[width=0.8\linewidth]{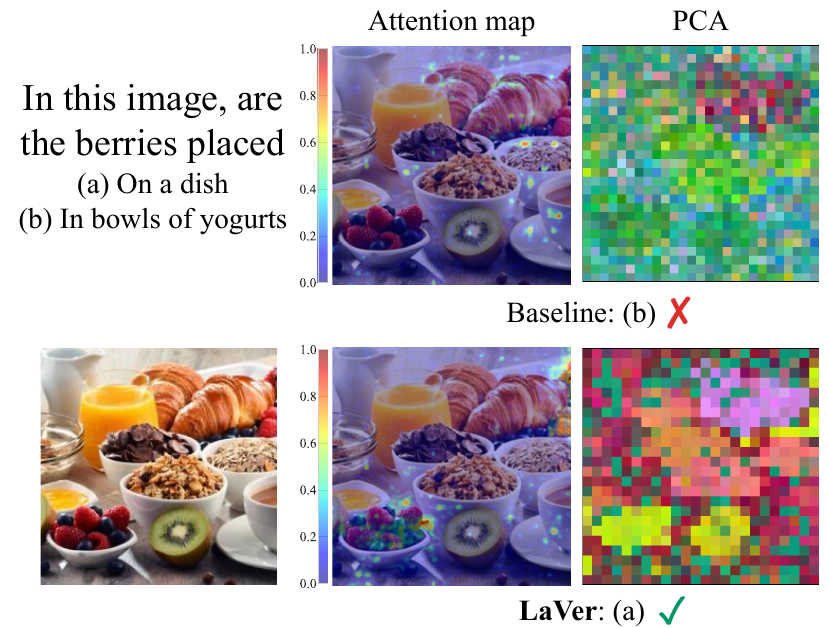}
    \vspace{-10pt}
    \caption{\textbf{Qualitative comparison.} LaVer allocates higher attention values with specific focus on corresponding visual regions, with more discriminative visual representations.}
    \label{fig:qualitative}
    \vspace{-10pt}
\end{figure}

\textbf{Ablation on spatial awareness.} Table~\ref{tab:ablations_combined}(b) examines spatial awareness components, i.e., mixed attention and 2D-RoPE, with averaged results.
Mixed attention alone improves over the baseline (\textbf{56.78\%} vs. \textbf{55.72\%} for SigLIP 2) but underperforms full LaVer (\textbf{57.87\%}), demonstrating that LaVer's gains stem from enhanced visual representation learning rather than merely attending to all vision tokens. 
2D-RoPE exhibits minimal impact (\textbf{55.57\%} for SigLIP 2), potentially requiring longer training to realize its benefits.
The complete LaVer framework consistently surpasses all ablated variants, confirming that spatial awareness components effectively complement the core methodology.

\textbf{Ablation on loss components.} Table~\ref{tab:ablations_combined}(c) dissects the contribution of individual loss components. Applying $\mathcal{L}_{\text{MIM}}$ alone degrades performance (e.g., \textbf{53.76\%} vs. \textbf{55.72\%} for SigLIP 2), empirically confirming the \textit{visual feature inconsistency} phenomenon identified in Sec.~\ref{subsec:inconsistency}, where visual information loss manifests as increased vision-token cosine similarity, diminishing visual structural information. While $\mathcal{L}_{\text{GA}}$ preserves structure, it constrains discriminative learning. Conversely, $\mathcal{L}_{\text{CGA}}$ achieves optimal performance by balancing structural fidelity with feature discrimination.

\subsection{Discussion}

\begin{table}
\centering
\caption{\textbf{Performance Comparison with the reconstruction-based baseline.} (\%) For fair comparison, we use the same model and datasets in~\cite{wang2025reconstructive}. LaVer outperforms ROSS on average.}
\vspace{-8pt}
\label{tab:ross_cmp}
\resizebox{\linewidth}{!}{
\begin{tabular}{l|ccc|ccc}
    \toprule[1.5pt]
    \multirow{2}{*}{\rotatebox[origin=c]{0}{\textbf{Benchmark}}} & \multicolumn{3}{c|}{CLIP-ViT-L/14@336} & \multicolumn{3}{c}{SigLIP-ViT-SO400M/14@384} \\
    \cmidrule(lr){2-4} \cmidrule(lr){5-7}
    & \textbf{Baseline} & \textbf{ROSS} & \textbf{LaVer} & \textbf{Baseline} & \textbf{ROSS} & \textbf{LaVer} \\
    \midrule[1.0pt]
    MMB$^{\text{EN}}$ & 66.15 & \gain{1.41} & \textbf{\gain{4.55}} & 70.96 & \gain{0.63} & \textbf{\gain{1.78}} \\
    SEED$^{\text{I}}$ & 60.42 & \gain{0.43} & \textbf{\gain{2.46}} & 65.31 & \loss{0.18} & \textbf{\gain{0.47}} \\
    RWQA & 53.07 & \textbf{\gain{0.69}} & \gain{0.09} & 57.39 & \gain{1.22} & \textbf{\gain{3.26}} \\
    MMMU & 36.78 & \gain{0.70} & \textbf{\gain{4.44}} & 37.56 & \textbf{\gain{1.00}} & {\gain{0.66}} \\
    OCRB & 363 & \textbf{\gain{18}} & {\gain{10}} & 376 & \gain{16} & \textbf{\gain{51}} \\
    TVQA & 38.05 & \gain{2.10} & \textbf{\gain{4.24}} & 52.56 & \gain{1.26} & \textbf{\gain{3.19}} \\
    CQA & 27.04 & \gain{3.13} & \textbf{\gain{4.32}} & 34.40 & \textbf{\gain{1.87}} & {\gain{0.72}} \\
    AI2D & 58.16 & \gain{1.41} & \textbf{\gain{1.67}} & 63.02 & \gain{0.48} & \textbf{\gain{0.83}} \\
    MMVP & 58.67 & \textbf{\gain{12.71}} & \gain{10.53} & 68.00 & \textbf{\gain{8.60}} & \gain{5.00} \\
    Hallu & 47.53 & \gain{1.68} & \textbf{\gain{1.79}} & 48.37 & \gain{0.85} & \textbf{\gain{3.37}} \\
    POPE & 88.43 & \gain{0.55} & \textbf{\gain{1.20}} & 88.97 & \gain{0.23} & \textbf{\gain{0.26}} \\
    \midrule
    Average & 48.59 & \gain{2.25} & \textbf{\gain{2.91}} & 53.36 & \gain{1.45} & \textbf{\gain{1.78}} \\
    \bottomrule[1.5pt]
\end{tabular}
}
\vspace{-8pt}
\end{table}

\textbf{Comparison with reconstruction-based baseline.} We compare LaVer against ROSS~\cite{wang2025reconstructive}, a reconstruction-based approach, under identical settings: Qwen2-7B-Instruct~\cite{yang2024qwen2technicalreport} with CLIP-ViT-L/14@336~\cite{pmlr-v139-radford21a} and SigLIP-ViT-SO400M/14@384~\cite{Zhai_2023_ICCV}, trained on LLaVA-558K~\cite{liu2023llava} and Cambrian-737K~\cite{10.5555/3737916.3740687}.
ROSS reconstructs low-level visual features, lacking semantic alignment with the language model's embedding space.
LaVer instead learns unified discriminative representations across modalities, directly mitigating modality imbalance.
Table~\ref{tab:ross_cmp} shows LaVer achieves superior gains of \textbf{2.91\%} and \textbf{1.78\%} with CLIP and SigLIP respectively, outperforming ROSS's \textbf{2.25\%} and \textbf{1.45\%}, with notable improvements on MMMU (+4.44\% vs. +0.70\%) and MMB$^{\text{EN}}$ (+4.55\% vs. +1.41\%) using CLIP. Additional details are in the supplementary material.

\textbf{Enhanced visual attention allocation.} LaVer substantially increases visual attention allocation (Fig.~\ref{fig:qualitative}), demonstrating improved visual information utilization and strengthened synergy between visual and textual modalities in the unified latent space. Additional quantitative and qualitative analyses are provided in the supplementary material.

% \textbf{Additional discussion.} We further provide empirical validation to \textit{progressive homogenization of visal features} under various scenarios, as well as additional discussion on \textit{visual feawture inconsistency}. The details are found in the supplementary material due to space limit.

% \textbf{Computation cost analysis.} The implementation of LaVer introduces additional computational overhead, primarily stemming from the forward passes required for both student and teacher models to process image tokens, as well as the exponential moving average (EMA) updates for the teacher model parameters. These operations collectively result in increased training time and memory consumption compared to the baseline approach. Nevertheless, the empirical benefits demonstrated by LaVer substantially justify this additional computational cost. As evidenced by our comprehensive experiments, LaVer consistently delivers performance improvements across MLLMs with diverse architectural configurations and achieves superior results on a wide range of benchmarks. A detailed computational cost analysis, including quantitative comparisons of training time and memory usage, is provided in the supplementary material.

% the additional image token forward for the student and teacher model, and teacher EMA updating increase the training time and memory usage. However, the benefits of LaVer outweigh the additional cost, as it consistently improve the MLLM with different architectures on various benchmarks. More computational analysis is found in the supplementary material.

%% file: sec/6_conclude.tex
\section{Conclusion}
\label{sec:conclude}

% limitations:
% 1. scalability is not fully validated on large-scale datasets, lack computation resource, with scaling, LaVer can potentially achieve sota performance compared with close-source commercial MLLMs with such self-evolution paradigm.
% 2. we don't include other modalities like video or audio, left for future work

% conclusion:
% we propose LaVer, a novel multimodal training framework that enable MLLMs to develop unified multimodal embeddings in the high-level semantic space, with direct visual supervisory signals provided online teacher, thus increasing visual information utilization and mitigating modality imbalance.

% While our work demonstrates promising results, several limitations warrant discussion.
% First, due to computational constraints, we have not fully validated the scalability of LaVer on large-scale datasets.
% We hypothesize that with increased scale, LaVer's self-evolution paradigm could potentially achieve state-of-the-art performance comparable to closed-source commercial MLLMs, though this remains to be empirically verified.
% Second, our current framework focuses exclusively on vision-language modalities.
% The extension to other modalities, such as video and audio, represents an important direction for future work.

This paper presents LaVer, a novel multimodal training framework that enables MLLMs to develop intrinsic visual representations within a unified high-level semantic space.  A new regularizer is introduced to offer asymmetric guidances towards the visual feature deviation.
With direct visual supervision from an online teacher, LaVer enhances visual information utilization and effectively addresses the modality imbalance prevalent in existing MLLMs.
Our experiments demonstrate that explicit visual guidance during training substantially improves multimodal representations, yielding more robust and capable multimodal systems.
We believe LaVer's self-supervised paradigm offers a promising direction for advancing multimodal learning.

% This paper presents LaVer, a novel multimodal training framework that enables MLLMs to develop intrinsic visual representations within a high-level semantic space. A new regularizer is introduced to offer correct guidances towards the visual feature deviation.
% Through direct visual supervisory signals, LaVer enhances visual information utilization and effectively mitigates the modality imbalance problem that commonly afflicts MLLMs.
% Our approach demonstrates that explicit visual guidance during training can substantially improve the multimodal representation with better utiliztion of each modality, leading to more robust and capable multimodal systems.
% We believe that LaVer's self-evolution paradigm offers a promising direction for advancing the development of more balanced and effective MLLMs.

%% file: sec/X_suppl.tex
\clearpage
\setcounter{page}{1}
\maketitlesupplementary

\renewcommand{\thesection}{\Alph{section}}
\setcounter{section}{0}

\begin{figure*}[t]
\centering
% \fbox{\rule{0pt}{2.5in} \rule{0.9\linewidth}{0pt}}
\begin{subfigure}{0.33\linewidth}
% \fbox{\rule{0pt}{1in} \rule{.9\linewidth}{0pt}}
\includegraphics[width=0.99\linewidth]{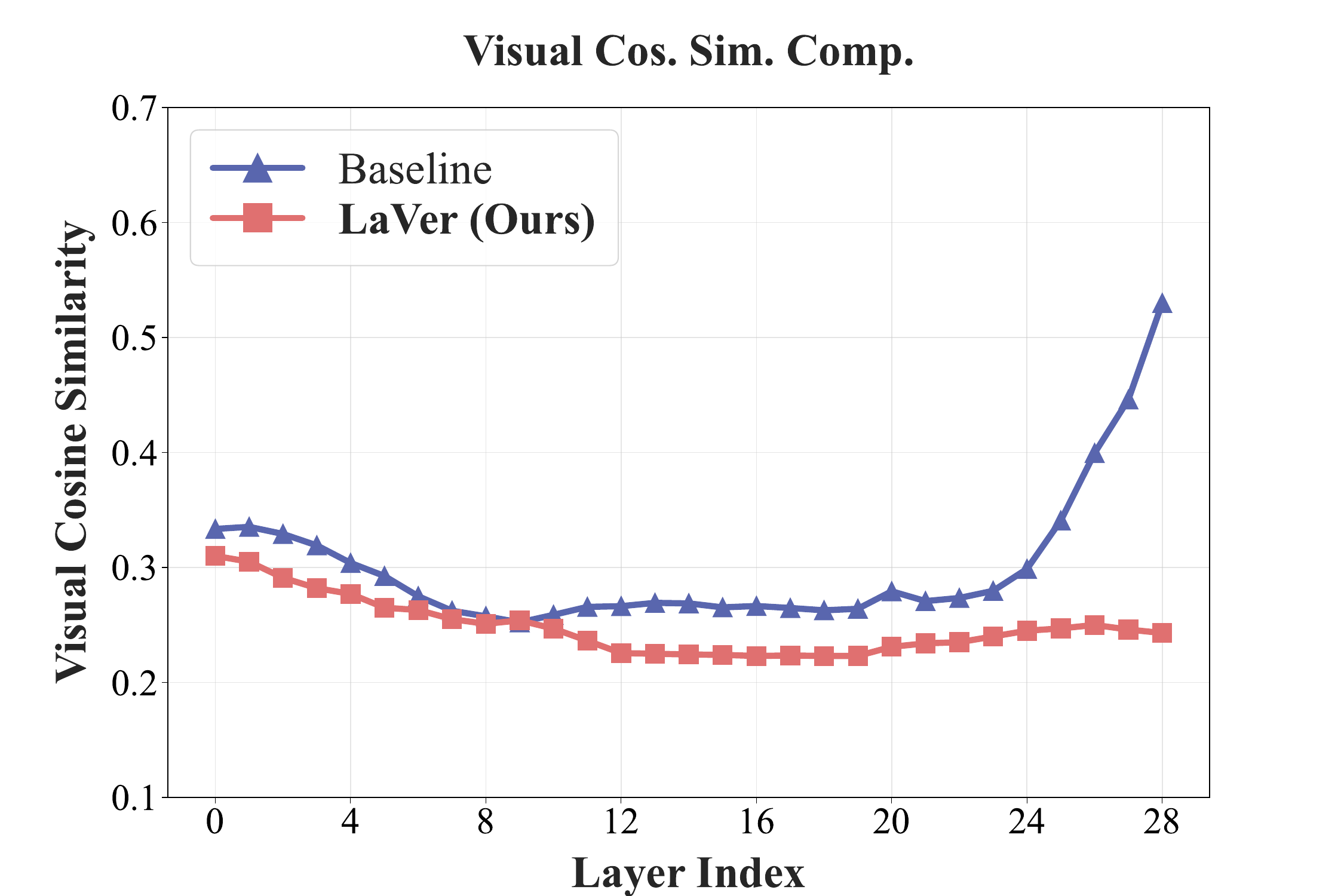}
\caption{Visual cosine similarity for SigLIP 2.}
\label{fig:supp_cos_sim_siglip}
\end{subfigure}
\hfill
\begin{subfigure}{0.33\linewidth}
% \fbox{\rule{0pt}{1in} \rule{.95\linewidth}{0pt}}
\includegraphics[width=0.99\linewidth]{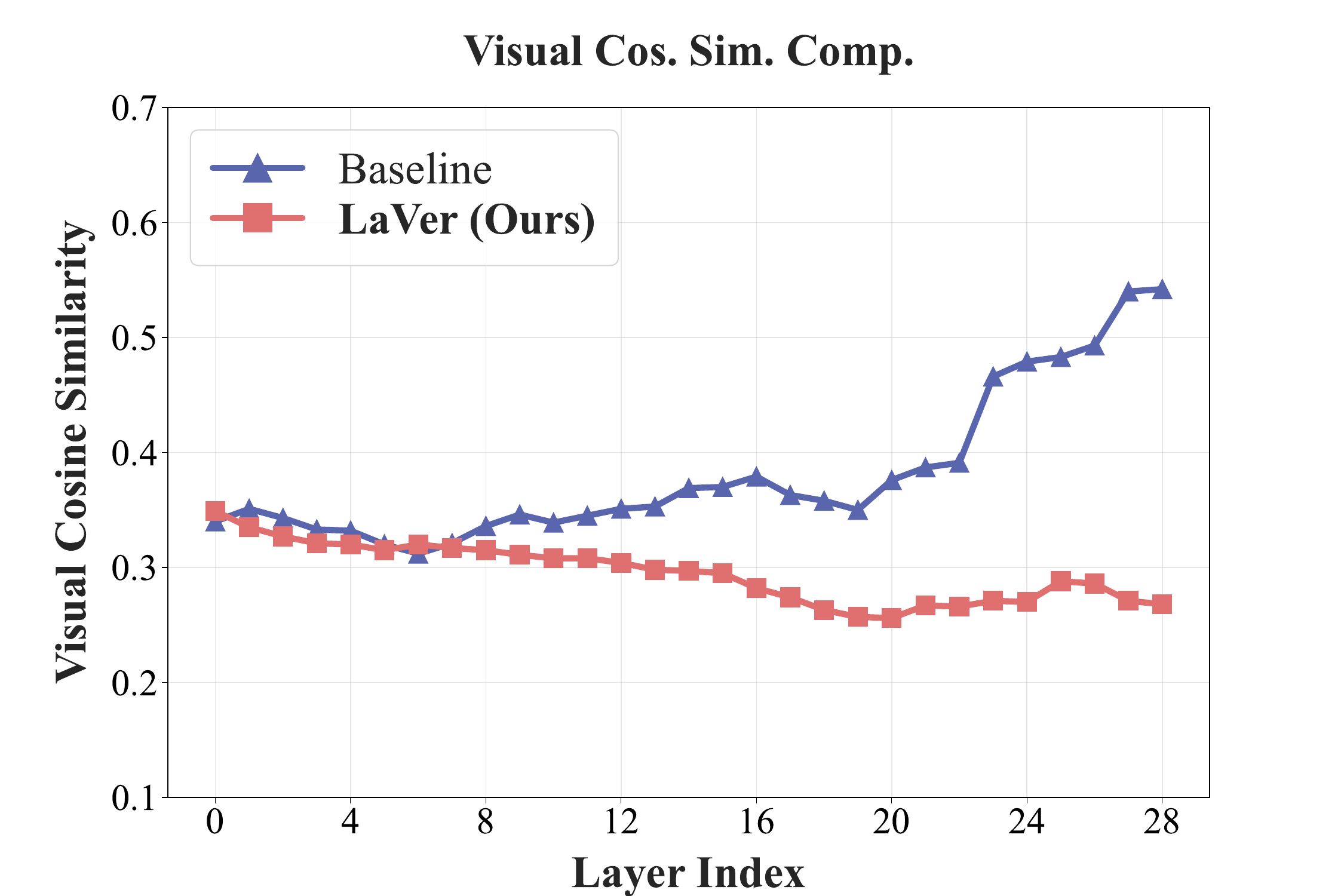}
\caption{Visual cosine similarity for CLIP.}
\label{fig:supp_cos_sim_clip}
\end{subfigure}
\hfill
\begin{subfigure}{0.33\linewidth}
% \fbox{\rule{0pt}{1in} \rule{.95\linewidth}{0pt}}
\includegraphics[width=0.99\linewidth]{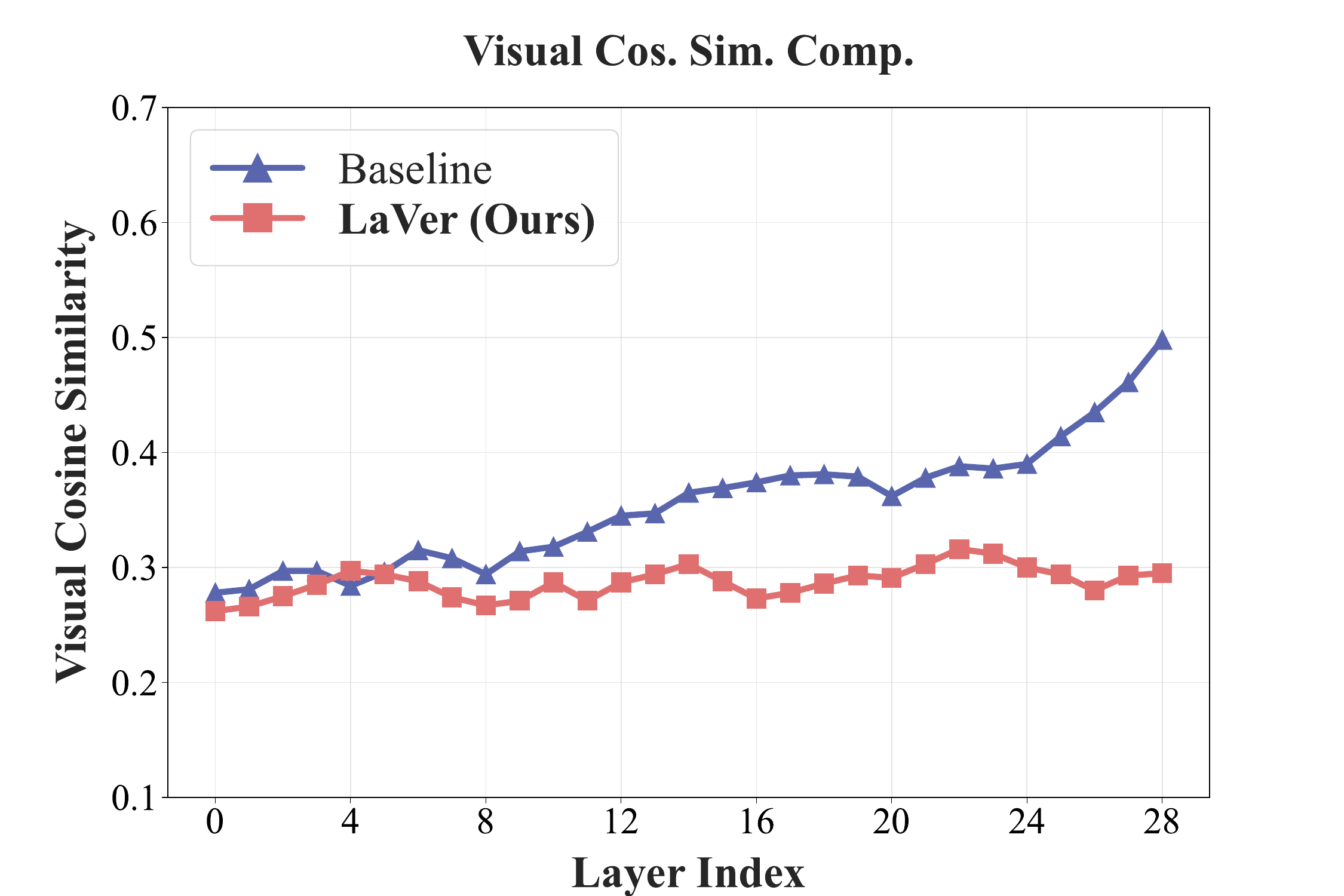}
\caption{Visual cosine similarity for DINOv2.}
\label{fig:supp_cos_sim_dino}
\end{subfigure}
\vspace{-20pt}
\caption{\textbf{Averaged visual cosine similarity across the layers.} \textbf{(a)} shows the visual cosine similarity for SigLIP 2~\cite{tschannen2025siglip2multilingualvisionlanguage}. \textbf{(b)} shows the visual cosine similarity for CLIP~\cite{pmlr-v139-radford21a}. \textbf{(c)} shows the visual cosine similarity for DINOv2~\cite{oquab2024dinov}.}
\label{fig:supp_cos_sim}
% \vspace{-10pt}
\end{figure*}

\begin{figure*}[t]
\centering
% \fbox{\rule{0pt}{2.5in} \rule{0.9\linewidth}{0pt}}
\begin{subfigure}{0.33\linewidth}
% \fbox{\rule{0pt}{1in} \rule{.9\linewidth}{0pt}}
\includegraphics[width=0.99\linewidth]{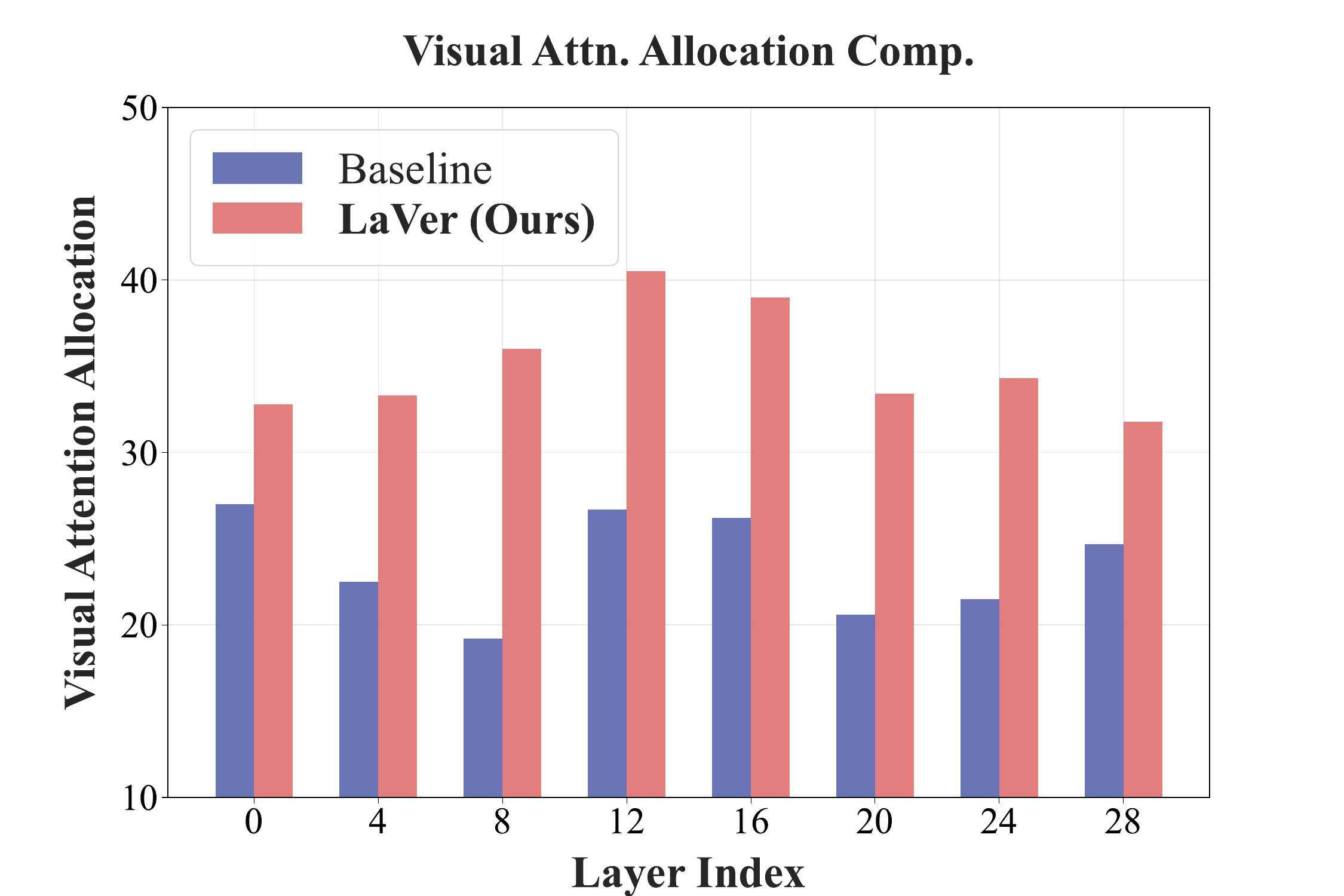}
\caption{Visual attention allocation for SigLIP 2.}
\label{fig:supp_attn_allocation_siglip}
\end{subfigure}
\hfill
\begin{subfigure}{0.33\linewidth}
% \fbox{\rule{0pt}{1in} \rule{.95\linewidth}{0pt}}
\includegraphics[width=0.99\linewidth]{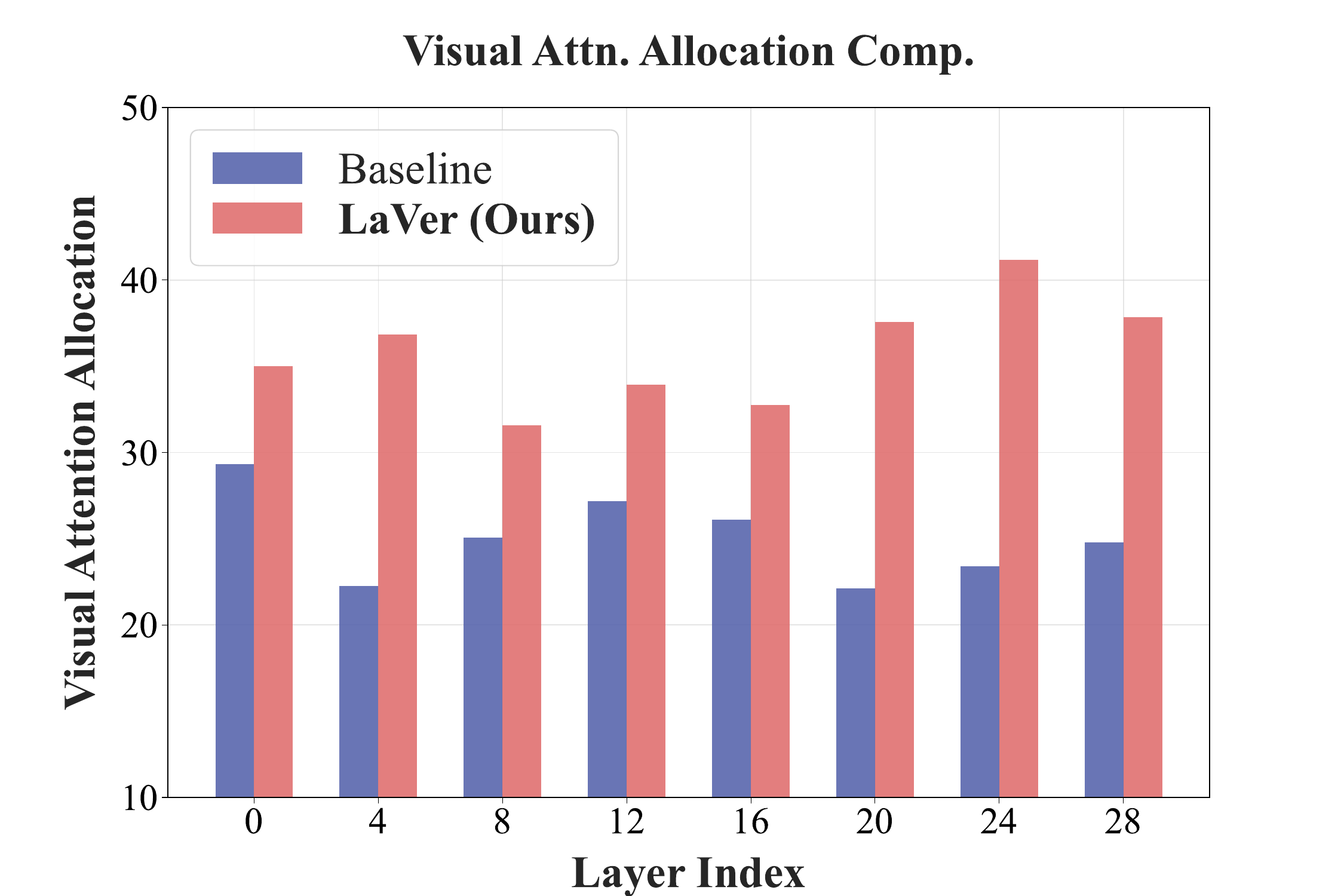}
\caption{Visual attention allocation for CLIP.}
\label{fig:supp_attn_allocation_clip}
\end{subfigure}
\hfill
\begin{subfigure}{0.33\linewidth}
% \fbox{\rule{0pt}{1in} \rule{.95\linewidth}{0pt}}
\includegraphics[width=0.99\linewidth]{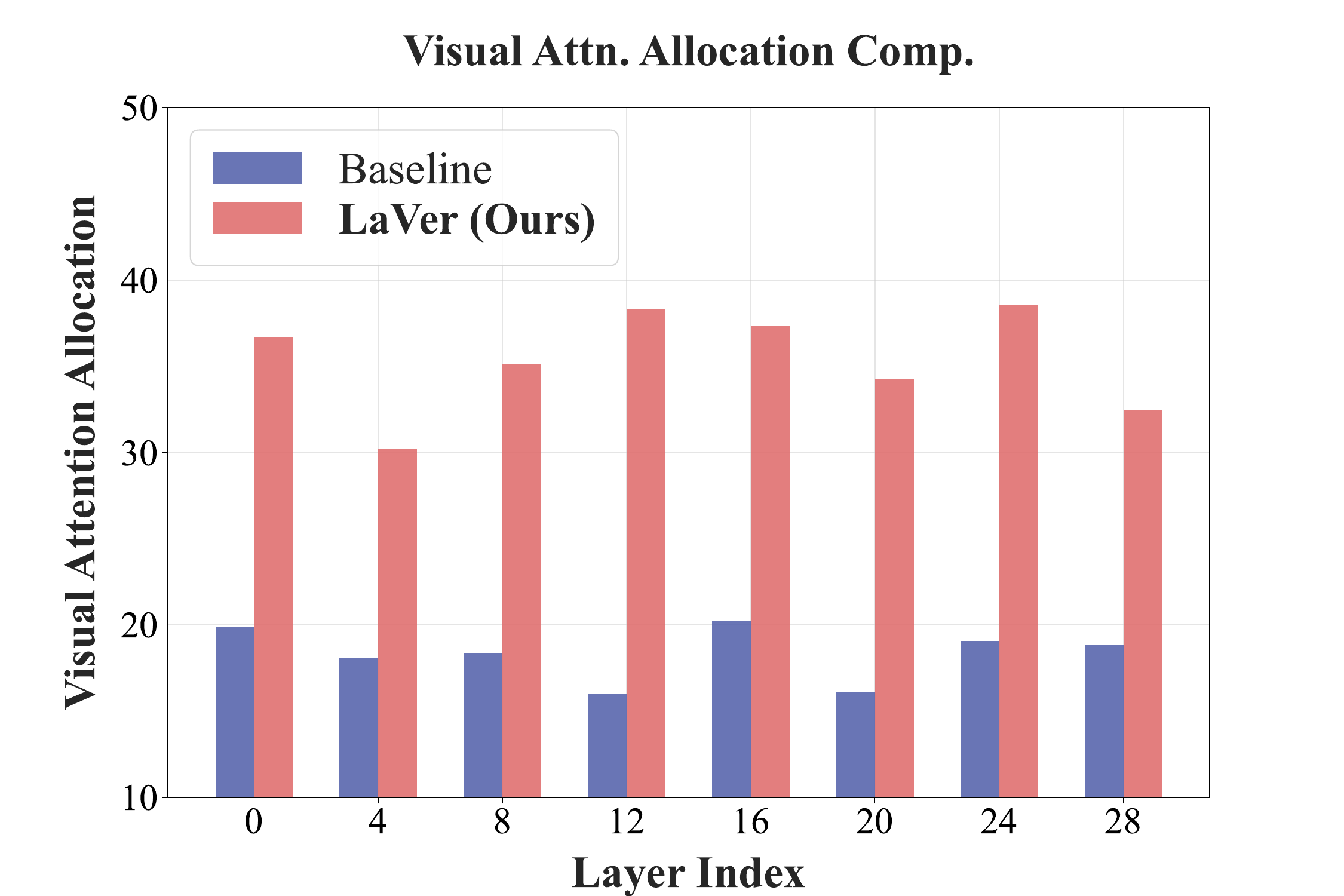}
\caption{Visual attention allocation for DINOv2.}
\label{fig:supp_attn_allocation_dino}
\end{subfigure}
\vspace{-20pt}
\caption{\textbf{Averaged visual attention allocation across the layers.} \textbf{(a)} shows the visual attention allocation for SigLIP 2~\cite{tschannen2025siglip2multilingualvisionlanguage}. \textbf{(b)} shows the visual attention allocation for CLIP~\cite{pmlr-v139-radford21a}. \textbf{(c)} shows the visual attention allocation for DINOv2~\cite{oquab2024dinov}.}
\label{fig:supp_attn_allocation}
% \vspace{-10pt}
\end{figure*}

\begin{figure*}[t]
\centering
% \fbox{\rule{0pt}{2.5in} \rule{0.9\linewidth}{0pt}}
\begin{subfigure}{0.33\linewidth}
% \fbox{\rule{0pt}{1in} \rule{.9\linewidth}{0pt}}
\includegraphics[width=0.99\linewidth]{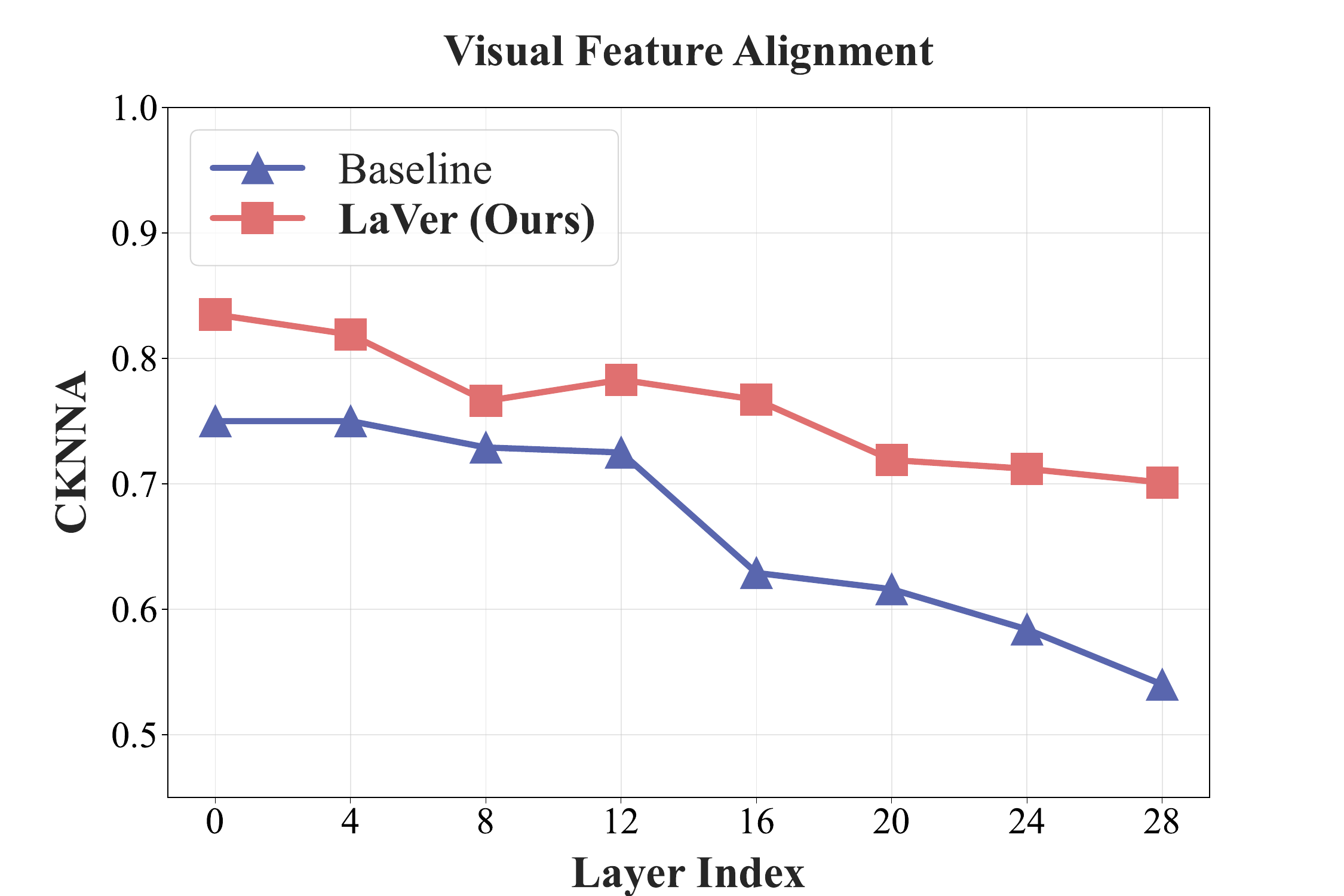}
\caption{CKNNA for SigLIP 2.}
\label{fig:supp_cknna_siglip}
\end{subfigure}
\hfill
\begin{subfigure}{0.33\linewidth}
% \fbox{\rule{0pt}{1in} \rule{.95\linewidth}{0pt}}
\includegraphics[width=0.99\linewidth]{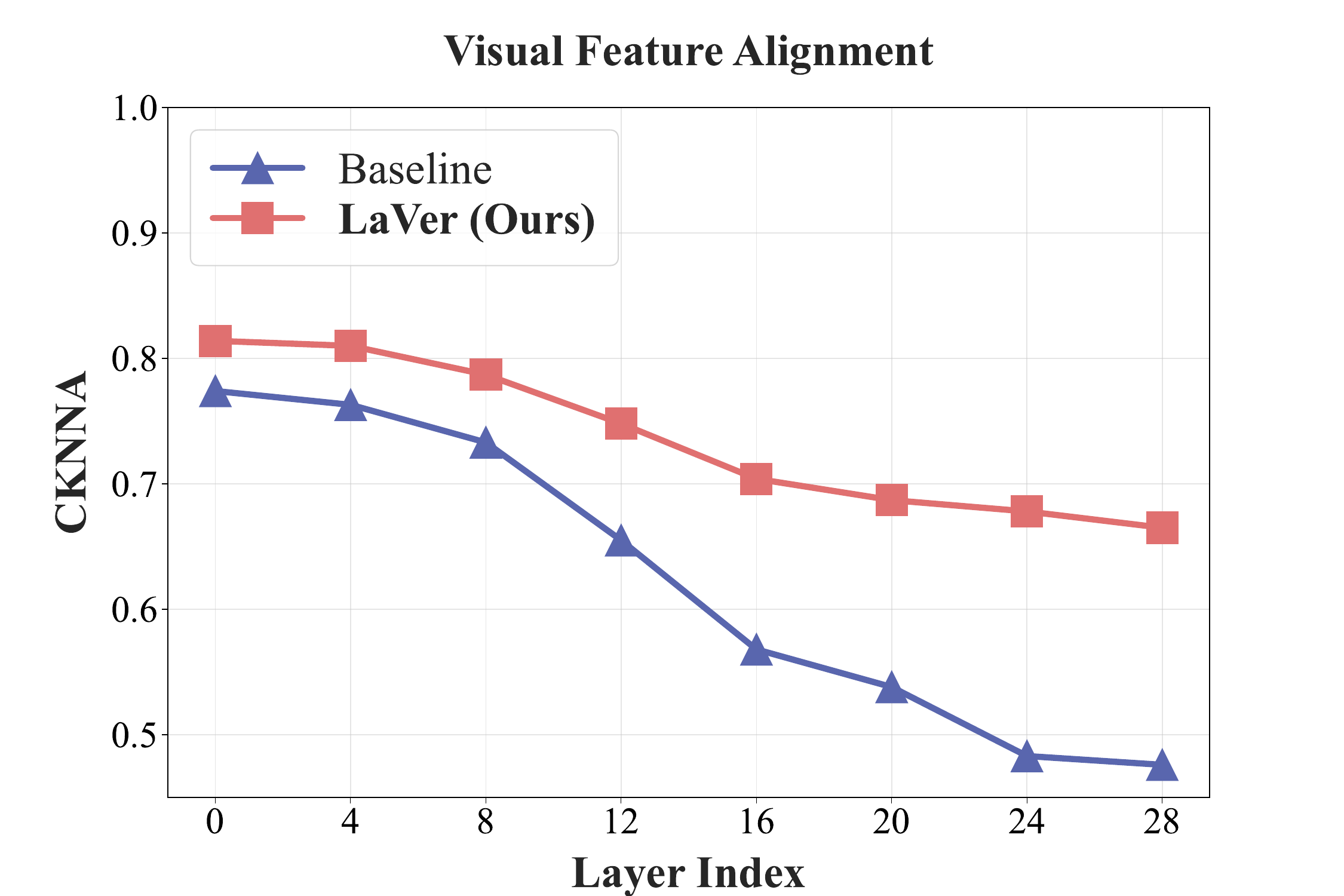}
\caption{CKNNA for CLIP.}
\label{fig:supp_cknna_clip}
\end{subfigure}
\hfill
\begin{subfigure}{0.33\linewidth}
% \fbox{\rule{0pt}{1in} \rule{.95\linewidth}{0pt}}
\includegraphics[width=0.99\linewidth]{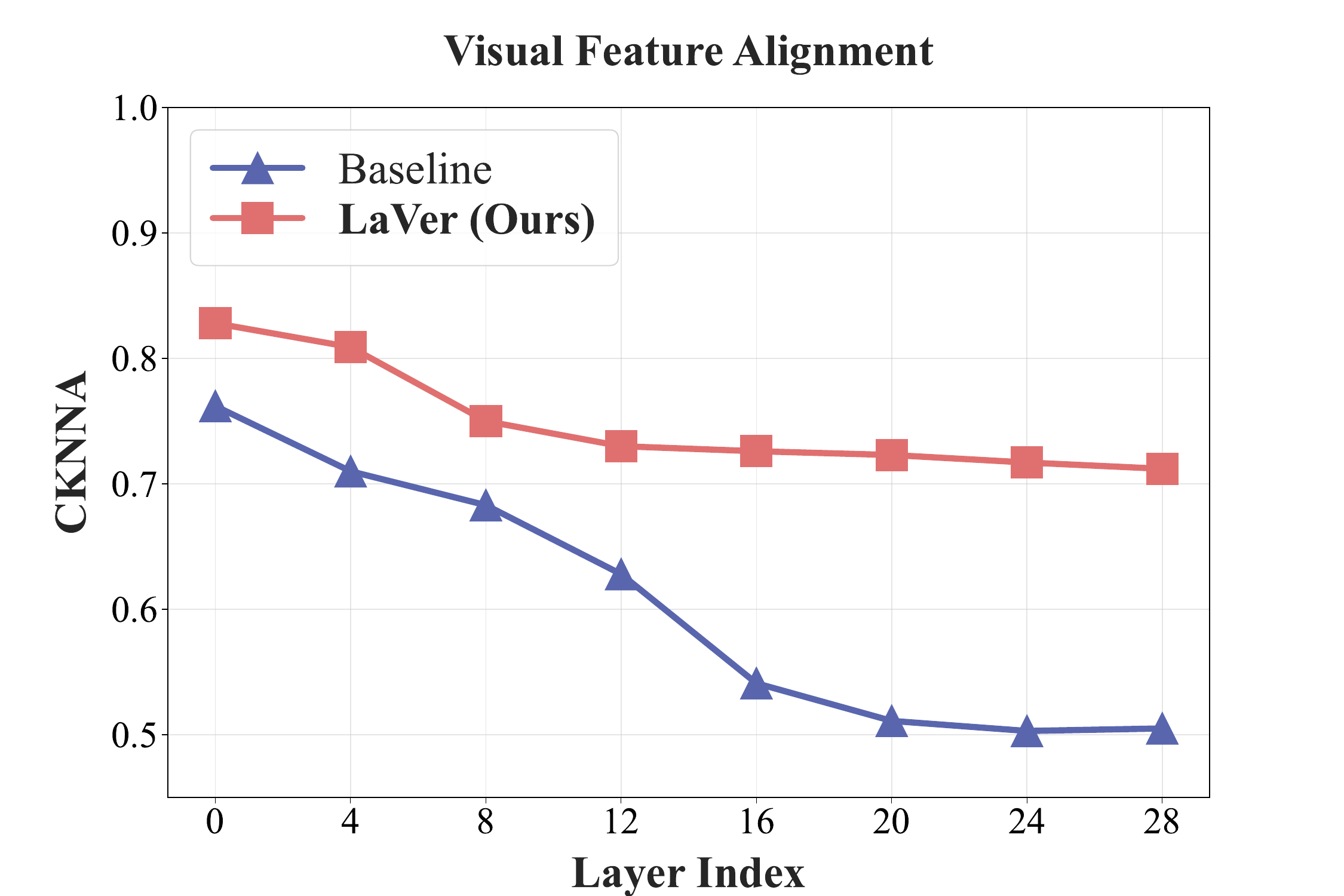}
\caption{CKNNA for DINOv2.}
\label{fig:supp_cknna_dino}
\end{subfigure}
\vspace{-20pt}
\caption{\textbf{CKNNA metrics across the layers.} \textbf{(a)} displays the CKNNA metric for SigLIP 2~\cite{tschannen2025siglip2multilingualvisionlanguage}. \textbf{(b)} displays the CKNNA metric for CLIP~\cite{pmlr-v139-radford21a}. \textbf{(c)} displays the CKNNA metric for DINOv2~\cite{oquab2024dinov}.}
\label{fig:supp_cknna}
% \vspace{-10pt}
\end{figure*}

\section{Additional Preliminaries}
\label{sec:add_preliminaries}

\textbf{Masked Image Modeling.} Masked Image Modeling (MIM) is a self-supervised learning paradigm wherein a model learns to reconstruct representations of masked image regions. Following iBOT~\cite{zhou2021ibot}, an input image $\bm{I}$ is partitioned into $N$ patches and encoded via a vision encoder $\mathcal{G}_{\upxi}$. A binary mask $\mathcal{M} \in \{0,1\}^N$ indicates which patches are masked. iBOT employs a teacher-student framework where the student predicts the teacher's representations at masked positions. The objective minimizes the cross-entropy between student predictions and teacher targets:
\begin{equation}
\label{eq:mim}
    \mathcal{L}_{\text{MIM}} = -\sum_{i \in \mathcal{P}_{\mathcal{M}}} \text{softmax}(\hat{\bm{z}}_i / \tau_{\text{tea.}}) \cdot \log \text{softmax}(\tilde{\bm{z}}_i / \tau_{\text{stu.}}),
\end{equation}
where $\mathcal{P}_{\mathcal{M}} = \{i \in \{1, \ldots, N\} \mid \mathcal{M}_i = 1\}$ denotes the masked position indices, and $\tau_{\text{tea.}}, \tau_{\text{stu.}}$ are temperature parameters controlling the softmax sharpness for teacher and student distributions, respectively. We adopt $\tau_{\text{tea.}} = 0.04$ and $\tau_{\text{stu.}} = 0.1$ by default. This formulation enables the student to learn discriminative semantic representations through reconstruction of masked region features.

\textbf{CKNNA.} \textit{Centered Kernel Nearest-Neighbor Alignment} (CKNNA)~\cite{huh2024platonicrepresentationhypothesis} is a metric for measuring feature alignment between models, derived as a relaxed variant of \textit{Centered Kernel Alignment} (CKA)~\cite{kornblith2019similarityneuralnetworkrepresentations}. Given two feature sets, CKA quantifies their global similarity through kernel matrices as:
\begin{equation}
    \label{eq:cka}
    \text{CKA}(\textbf{K}, \textbf{L}) = \frac{\text{HSIC}(\textbf{K}, \textbf{L})}{\sqrt{\text{HSIC}(\textbf{K}, \textbf{K}) \text{HSIC}(\textbf{L}, \textbf{L})}},
\end{equation}
where $\textbf{K}$ and $\textbf{L}$ denote kernel matrices computed from the feature sets, and $\text{HSIC}(\cdot, \cdot)$ represents the Hilbert-Schmidt Independence Criterion measuring feature dependence. The kernel matrices are defined as $\textbf{K}_{ij} = \kappa(\bm{k}_i, \bm{k}_j)$ and $\textbf{L}_{ij} = \kappa(\bm{l}_i, \bm{l}_j)$, where $\kappa(\cdot, \cdot)$ is the kernel function and $\bm{k}_i$, $\bm{l}_i$ are feature vectors. Using the inner product kernel, HSIC is formulated as:
\begin{equation}
    \label{eq:hsic}
    \begin{split}
    \text{HSIC}(\textbf{K}, \textbf{L}) & = \frac{1}{(N-1)^2} \Big ( \sum_{i=1}^N \sum_{j=1}^N \big (\langle \bm{k}_i, \bm{k}_j\rangle - \mathbb{E}[\langle \bm{k}_i, \bm{k}_j\rangle] \big ) \\
    & \big (\langle \bm{l}_i, \bm{l}_j\rangle - \mathbb{E}[\langle \bm{l}_i, \bm{l}_j\rangle] \big) \Big ).
    \end{split}
\end{equation}
CKNNA refines CKA by restricting alignment to nearest neighbors, replacing $\text{HSIC}(\cdot, \cdot)$ with $\text{HSIC}_{\text{kNN}}(\cdot, \cdot)$:
\begin{equation}
    \label{eq:hsic_knn}
    \begin{split}
    \text{HSIC}_{\text{kNN}}&(\textbf{K}, \textbf{L})  = \frac{1}{(N-1)^2} \Big ( \sum_{i=1}^N \sum_{j=1}^N \mathbb{I}(i,j)\big (\langle \bm{k}_i, \bm{k}_j\rangle \\
    & - \mathbb{E}[\langle \bm{k}_i, \bm{k}_j\rangle] \big ) \big (\langle \bm{l}_i, \bm{l}_j\rangle - \mathbb{E}[\langle \bm{l}_i, \bm{l}_j\rangle]\big ) \Big ),
    \end{split}
\end{equation}
where $\mathbb{I}(i,j)$ is the nearest-neighbor indicator:
\begin{equation}
    \mathbb{I}(i,j) = \left\{\begin{array}{ll}
    1 & \text{if } i\neq j, \bm{k}_i \in \text{kNN}(\bm{l}_j; k), \bm{l}_j \in \text{kNN}(\bm{k}_i; k), \\
    0 & \text{otherwise},
    \end{array}
    \right.
\end{equation}
with $\text{kNN}(\bm{x}; k)$ denoting the $k$-nearest neighbors of $\bm{x}$. We set $k=10$ by default.

\section{Additional Discussion on Visual Feature Homogenization}

MLLMs exhibit \textit{modality imbalance}~\cite{10.1145/3746027.3755364,zheng2025mllmsdeeplyaffectedmodality,10.1016/j.patcog.2025.111670}, systematically biasing toward textual information over visual inputs~\cite{10.1145/3746027.3755364,zheng2025mllmsdeeplyaffectedmodality,10.1609/aaai.v39i19.34183,liu2025faithfulnessvisualthinkingmeasurement,jia2024symdpoboostingincontextlearning,leng2024cursemultimodalitiesevaluatinghallucinations}, with more allocated attention scores and predictions predominantly grounded in text modality~\cite{wu2025languageoverrulesrevealingtext,10.1609/aaai.v39i19.34183}.

To further validate the \textit{progressive visual feature homogenization} phenomenon illustrated in Fig. 2 of the main text, we conduct comprehensive empirical analyses across multiple vision encoders and evaluation metrics.
This phenomenon reveals a critical insight: MLLMs progressively discard rich visual information throughout their layers, potentially retaining only those visual features directly relevant to textual reasoning tasks.
We posit that this mechanism fundamentally stems from the \textit{next-text-token-prediction} objective, which inherently emphasizes the text modality and provides only indirect, implicit supervisory signals for developing intrinsic visual modeling capabilities.
While this text-centric training paradigm proves effective for tasks demanding sophisticated language generation competence, it is suboptimal for training MLLMs that should seamlessly integrate multimodal information without disproportionately favoring any particular modality~\cite{chen2024quantifyingmitigatingunimodalbiases,10655378}.
The limited performance of MLLMs on dense visual understanding tasks serves as compelling evidence of this text-modality bias.

We provide extensive empirical validation of the \textit{progressive visual feature homogenization} phenomenon in Fig.~\ref{fig:supp_cos_sim}, Fig.~\ref{fig:supp_attn_allocation}, and Fig.~\ref{fig:supp_cknna}, examining diverse vision encoders including Qwen2.5-7B-Instruct~\cite{qwen2025qwen25technicalreport} paired with SigLIP 2~\cite{tschannen2025siglip2multilingualvisionlanguage}, CLIP~\cite{pmlr-v139-radford21a}, and DINOv2~\cite{oquab2024dinov}, demonstrating that this phenomenon persists across different architectural configurations.
Specifically, we employ the CKNNA metric to quantify feature alignment between intermediate visual representations and input visual features, with detailed formulation provided in Sec.~\ref{sec:add_preliminaries}.
We construct visual feature sets using images from MMVP~\cite{10655378} and set $k=10$ as the default neighborhood size.

Our results reveal several critical observations. First, visual feature homogenization intensifies in deeper layers, as evidenced by the progressively increasing averaged visual feature-wise cosine similarity shown in Fig.~\ref{fig:supp_cos_sim}.
Second, the substantially diminished visual attention allocation illustrated in Fig.~\ref{fig:supp_attn_allocation} demonstrates that models predominantly leverage information from text tokens, leaving abundant visual information severely underutilized. Notably, our empirical findings indicate that this underutilization of visual information persists consistently across all layers.
Third, the gradually decreasing CKNNA metric depicted in Fig.~\ref{fig:supp_cknna} indicates progressive misalignment of intermediate visual features from their original representations. In stark contrast to the lower CKNNA similarity exhibited by baseline models, our proposed LaVer consistently maintains substantially higher CKNNA similarity scores.
This demonstrates that LaVer effectively enables models to preserve rich visual information from vision encoders and cultivate robust intrinsic visual modeling capabilities.

The comprehensive evaluation across diverse benchmarks conclusively demonstrates that by introducing explicit vision-centric supervisory signals, models' multimodal capabilities can be significantly enhanced, particularly on tasks demanding dense visual information comprehension and fine-grained visual understanding.

\begin{table*}[t]
\centering
\caption{\textbf{Hyperparameters of training stages.}}
\label{tab:hyper}
\resizebox{\linewidth}{!}{
\begin{tabular}{l|ccc}
\toprule[1.5pt]
Configuration & Stage 1 & Stage 2 & Stage 3 \\
\midrule[1pt]
Trainable parameters & Connector $\mathcal{H}_{\phi}$ & Connector $\mathcal{H}_{\phi}$, LLM $\mathcal{F}_{\theta}$, Vision Head $\mathcal{V}_{\psi}$ & Connector $\mathcal{H}_{\phi}$, LLM $\mathcal{F}_{\theta}$ \\
Frozen parameters & Vision Encoder $\mathcal{G}_{\upxi}$, LLM $\mathcal{F}_{\theta}$ & Vision Encoder $\mathcal{G}_{\upxi}$, Teacher LLM $\mathcal{F}_{\hat{\theta}}$, Teacher Vision Head $\mathcal{V}_{\hat{\psi}}$ & Vision Encoder $\mathcal{G}_{\upxi}$ \\
Global batch size & 128 & 128 & 128 \\
Batch size per GPU & 4 & 2 & 4 \\
Accumulation steps & 2 & 4 & 2 \\
Max sequence length & 2048 & 2048 & 2048 \\
DeepSpeed Zero Stage & 2 & 2 & 2 \\
Learning rate & $2.0\times 10^{-3}$ & $2.0\times 10^{-5}$ & $1.0\times 10^{-5}$ \\
Learning rate schedule & Cosine & Cosine & Cosine \\
Warmup ratio & 0.05 & 0.05 & 0.05 \\
Weight decay & 0 & 0 & 0 \\
Training steps & 4360 & 6250 & 6250 \\
Data scale & 558K & 800K & 800K \\
Optimizer & AdamW & AdamW & AdamW \\
$\beta_1, \beta_2$ & 0.9, 0.999 & 0.9, 0.999 & 0.9, 0.999 \\
Precision & bf16 & bf16 & bf16 \\
\bottomrule[1.5pt]
\end{tabular}
}
\end{table*}

\begin{table}[t]
\centering
\caption{\textbf{Hyperparameters of LaVer.}}
\label{tab:hyper_laver}
\resizebox{0.6\linewidth}{!}{
\begin{tabular}{l|ccc}
\toprule[1.5pt]
Configuration & LaVer \\
\midrule[1pt]
Visual hidden dimension & 8192 \\
Loss coefficient $\omega_{\text{MIM}}$ & 1.0 \\
Loss coefficient $\omega_{\text{CGA}}$ & 1.0 \\
Teacher temperature $\tau_{\text{tea.}}$ & 0.04 \\
Student temperature $\tau_{\text{stu.}}$ & 0.1 \\
Masking ratio & 0.05 \\
Masking schedule & Cosine \\
EMA decay rate & 0.95 \\
EMA update steps & 100 \\
EMA schedule & Cosine \\
\bottomrule[1.5pt]
\end{tabular}
}
\end{table}

\begin{table}[t]
\centering
\caption{\textbf{Hyperparameters of ReasonSeg.}}
\label{tab:reasonseg_config}
\resizebox{0.8\linewidth}{!}{
\begin{tabular}{l|ccc}
\toprule[1.5pt]
Configuration & ReasonSeg \\
\midrule[1pt]
Trainable parameters & Connector $\mathcal{H}_{\phi}$, LLM $\mathcal{F}_{\theta}$\\
Frozen parameters & Vision Encoder $\mathcal{G}_{\upxi}$ \\
Global batch size & 128 \\
Batch size per GPU & 4 \\
Accumulation steps & 2 \\
Max sequence length & 2048 \\
DeepSpeed Zero Stage & 2 \\
Learning rate & $2.0\times 10^{-4}$ \\
Learning rate schedule & Cosine \\
Warmup ratio & 0.01 \\
Weight decay & 0.01 \\
Training steps & 30000 \\
Data scale & 3.8M \\
Optimizer & AdamW \\
$\beta_1, \beta_2$ & 0.9, 0.999 \\
Precision & bf16 \\
\bottomrule[1.5pt]
\end{tabular}
}
\end{table}

\begin{figure}
\centering
% \fbox{\rule{0pt}{2.5in} \rule{0.9\linewidth}{0pt}}
\begin{subfigure}{0.6\linewidth}
% \fbox{\rule{0pt}{1in} \rule{.9\linewidth}{0pt}}
\includegraphics[width=0.99\linewidth]{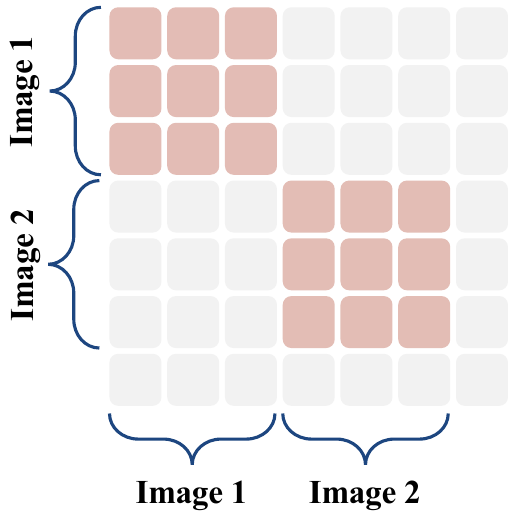}
\caption{Diagonally blocked full attention.}
\label{fig:packed_attn}
\end{subfigure}
\hfill
\begin{subfigure}{0.6\linewidth}
% \fbox{\rule{0pt}{1in} \rule{.95\linewidth}{0pt}}
\includegraphics[width=0.99\linewidth]{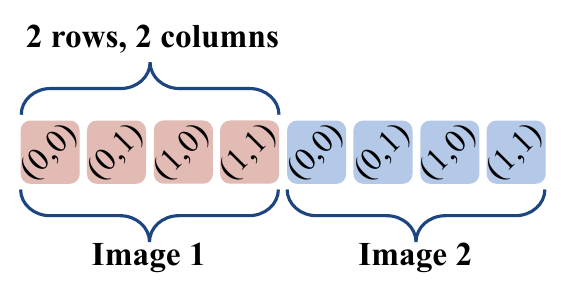}
\caption{Blocked 2D-RoPE.}
\label{fig:packed_rope}
\end{subfigure}
\vspace{-8pt}
\caption{\textbf{Illustration of packed visual sequences.} \textbf{(a)} illustrates diagonally blocked full attention for packed visual sequence. \textbf{(b)} illustrates blocked 2D-RoPE for packed visual sequence.}
\label{fig:packed_vision}
\vspace{-10pt}
\end{figure}

\begin{table*}[t]
\centering
\caption{\textbf{Computational cost comparison of stage 2.} LaVer introduces modest computational overhead compared to the baseline, with training time increases of 13-16\% and memory consumption increases of 14-26\% across different vision encoders. Despite these additional costs, the substantial performance improvements demonstrated in our experiments justify this overhead, making LaVer a practical and effective approach for enhancing MLLM capabilities.}
\label{tab:comput_cost}
\resizebox{0.8\linewidth}{!}{
\begin{tabular}{c|c|ccccc}
\toprule[1.5pt]
Vision encoder & Method & Trainable Parameters & Training Speed & Training Time & Total GFLOPs & Memory Consumption \\
\midrule[1pt]
\multirow{2}{*}{SigLIP 2} & Baseline & 7.63B & 4.87 s/iter~ & 8 h 27 min & $2.46\times 10^{10}$ & 55.24 GB \\
& LaVer & 7.65B & 5.51 s/iter & 9 h 34 min & $3.04\times 10^{10}$ & 69.82 GB \\
\midrule[1pt]
\multirow{2}{*}{CLIP} & Baseline & 7.63B & 4.23 s/iter~ & 7 h 42 min & $2.24\times 10^{10}$ & 53.24 GB  \\
& LaVer & 7.65B & 4.67 s/iter & 8 h 28 min & $2.80\times 10^{10}$ & 66.79 GB  \\
\midrule[1pt]
\multirow{2}{*}{DINOv2} & Baseline & 7.63B & 3.98 s/iter~ & 6 h 55 min & $1.85\times 10^{10}$ & 51.24 GB  \\
& LaVer & 7.65B & 4.45 s/iter~ & 7 h 44 min & $2.19\times 10^{10}$ & 58.24 GB  \\
\bottomrule[1.5pt]
\end{tabular}
}
\end{table*}

\section{Additional Discussion on Visual Feature Inconsistency}

As demonstrated in Sec. 4.2, applying the Masked Image Modeling (MIM) objective in isolation paradoxically leads to increased visual cosine similarity, signaling severe visual information degradation rather than enhancement.
Our analysis of training dynamics reveals a critical pattern: while visual cosine similarity initially decreases during early training iterations, it rapidly escalates in subsequent phases, ultimately surpassing baseline levels.
This counterintuitive behavior arises because the MIM objective, while encouraging the model to reconstruct its own visual embeddings, fails to explicitly constrain the model from generating homogeneous visual features, i.e., a degenerate solution that minimizes MIM loss at the expense of preserving meaningful visual distinctions.

This phenomenon bears striking resemblance to observations reported in prior work~\cite{9709990,oquab2024dinov,siméoni2025dinov3}, where models progressively generate visual features exhibiting high cosine similarity with global semantic tokens while discarding fine-grained local structural information.
Given the conceptual alignment between our observations under isolated MIM training and these established findings, we adopt the terminology \textit{visual feature inconsistency} to characterize this pathology, wherein visual patches exhibit spuriously high cosine similarity despite encoding fundamentally distinct local visual information.
While this phenomenon was attributed to the global semantic contrastive loss in the DINO series~\cite{siméoni2025dinov3}, our empirical findings reveal that local MIM objectives can independently induce the same degenerative behavior, highlighting a new failure mode.

To address this challenge, the \textit{Gram-Anchoring} mechanism was proposed by~\cite{siméoni2025dinov3} to explicitly enforce preservation of spatial structural information while learning discriminative local embeddings for each patch.
However, \textit{Gram-Anchoring} exhibits a fundamental limitation: it symmetrically penalizes deviations in both directions, thereby inadvertently discouraging the emergence of discriminative visual features.
Specifically, when the model attempts to produce more distinctive representations characterized by lower feature-wise cosine similarity, it incurs penalties equivalent to those for generating overly homogeneous features.
While this symmetric regularization proves benign for vision-only models~\cite{siméoni2025dinov3}, it becomes problematic in the context of MLLMs, which inherently suffer from modality imbalance.
The pre-existing bias toward textual modality creates a perverse incentive: the model can exploit this imbalance by generating nearly identical visual features to trivially minimize the MIM objective, effectively circumventing genuine visual understanding.

To overcome this limitation, we propose \textit{Clipped Gram-Anchoring}, an asymmetric regularization strategy that selectively penalizes the model only when it tends to produce visual features with excessively high cosine similarity.
By imposing penalties exclusively on the homogenization direction while permitting the model to freely explore more discriminative feature spaces, this regularizer effectively prevents visual feature inconsistency.
This design aligns with the fundamental requirement for MLLMs, i.e., maintaining rich, discriminative visual representations that can meaningfully contribute to multimodal reasoning, rather than collapsing into degenerate solutions that superficially satisfy training objectives while sacrificing genuine visual understanding capabilities.

\begin{table*}
\centering
\caption{\textbf{Scalability of LaVer on model parameters.} LaVer demonstrates strong model scaling properties, consistently improving performance across different parameter sizes (1.5B, 3B, and 7B) with both SigLIP 2 and CLIP vision encoders.}
\label{tab:model_scale}
\resizebox{\linewidth}{!}{
\begin{tabular}{l|ccc ccc ccc|ccc ccc ccc}
\toprule[1.5pt]
\multirow{3}{*}{\rotatebox[origin=c]{0}{\textbf{Benchmark}}} & \multicolumn{9}{c|}{SigLIP2} & \multicolumn{9}{c}{CLIP} \\
\cmidrule(lr){2-10} \cmidrule(lr){11-19}
& \multicolumn{3}{c}{1.5B} & \multicolumn{3}{c}{3B} & \multicolumn{3}{c|}{7B} & \multicolumn{3}{c}{1.5B} & \multicolumn{3}{c}{3B} & \multicolumn{3}{c}{7B} \\
\cmidrule(lr){2-4} \cmidrule(lr){5-7} \cmidrule(lr){8-10} \cmidrule(lr){11-13} \cmidrule(lr){14-16} \cmidrule(lr){17-19}
& \textbf{Baseline} & \textbf{LaVer} & $\Delta_{\text{Baseline}}$ & \textbf{Baseline} & \textbf{LaVer} & $\Delta_{\text{Baseline}}$ & \textbf{Baseline} & \textbf{LaVer} & $\Delta_{\text{Baseline}}$ & \textbf{Baseline} & \textbf{LaVer} & $\Delta_{\text{Baseline}}$ & \textbf{Baseline} & \textbf{LaVer} & $\Delta_{\text{Baseline}}$ & \textbf{Baseline} & \textbf{LaVer} & $\Delta_{\text{Baseline}}$  \\
\midrule[1.0pt]
GQA & 45.98 & \textbf{48.51} & \gain{2.53} & 48.99 & \textbf{53.02} & \gain{4.03} & 55.03 & \textbf{56.78} & \gain{1.75} & 46.98 & \textbf{48.25} & \gain{1.27} & \textbf{50.00} & 48.99 & \loss{1.01} & 51.51 & \textbf{54.77} & \gain{3.26} \\
MMB$^\text{EN}$ & 61.89 & \textbf{65.51} & \gain{3.62} & 70.79 & \textbf{74.23} & \gain{3.44} & 73.97 & \textbf{75.60} & \gain{1.63} & 54.33 & \textbf{60.22} & \gain{5.89} & 62.80 & \textbf{63.23} & \gain{0.43} & 68.64 & \textbf{69.93} & \gain{1.29} \\
SEED$^{\text{I}}$ & 58.65 & \textbf{63.04} & \gain{4.39} & 64.45 & \textbf{66.10} & \gain{1.65} & 67.57 & \textbf{68.62} & \gain{1.05} & 56.18 & \textbf{62.22} & \gain{6.04} & 62.63 & \textbf{62.67} & \gain{0.04} & 64.36 & \textbf{65.20} & \gain{0.84} \\
MME & 1261.35 & \textbf{1285.18} & \gain{23.83} & 1384.78 & \textbf{1445.33} & \gain{60.55} & 1510.73 & \textbf{1512.50} & \gain{1.77} & 1175.37 & \textbf{1213.01} & \gain{37.64} & 1260.48 & \textbf{1379.26} & \gain{118.78} & 1289.62 & \textbf{1474.65} & \gain{185.03} \\
RWQA & \textbf{52.68} & 52.29 & \loss{0.39} & 56.08 & \textbf{58.56} & \gain{2.48} & 53.86 & \textbf{59.35} & \gain{5.49} & 50.85 & \textbf{52.16} & \gain{1.31} & \textbf{56.73} & 55.29 & \loss{1.44} & 54.25 & \textbf{56.47} & \gain{2.22} \\
MMMU & 40.11 & \textbf{41.11} & \gain{1.00} & 41.11 & \textbf{42.00} & \gain{0.89} & 44.78 & \textbf{46.33} & \gain{1.55} & 38.33 & \textbf{39.44} & \gain{1.11} & \textbf{42.44} & 40.67 & \loss{1.77} & \textbf{44.56} & \textbf{44.56} & \gain{0.00} \\
MM$^*$ & 38.02 & \textbf{40.90} & \gain{2.88} & 46.80 & \textbf{50.07} & \gain{3.27} & 49.06 & \textbf{52.01} & \gain{2.95} & 37.42 & \textbf{40.83} & \gain{3.41} & 41.97 & \textbf{44.05} & \gain{2.08} & 43.17 & \textbf{45.45} & \gain{2.28} \\
OCRB & 237 & \textbf{258} & \gain{21} & 353 & \textbf{397} & \gain{44} & 536 & \textbf{639} & \gain{103} & 136 & \textbf{162} & \gain{26} & 226 & \textbf{265} & \gain{39} & 306 & \textbf{365} & \gain{59} \\
TVQA & 42.42 & \textbf{44.96} & \gain{2.54} & 51.33 & \textbf{55.78} & \gain{4.45} & 62.06 & \textbf{63.93} & \gain{1.87} & 31.70 & \textbf{34.65} & \gain{2.95} & 40.85 & \textbf{48.49} & \gain{7.64} & 43.86 & \textbf{49.93} & \gain{6.07} \\
CQA & 27.84 & \textbf{31.52} & \gain{3.68} & 40.00 & \textbf{42.32} & \gain{2.32} & 43.52 & \textbf{50.24} & \gain{6.72} & 18.40 & \textbf{19.04} & \gain{0.64} & 26.00 & \textbf{30.52} & \gain{4.52} & 27.36 & \textbf{39.36} & \gain{12.00} \\
AI2D & 59.94 & \textbf{62.56} & \gain{2.62} & 72.05 & \textbf{75.39} & \gain{3.34} & 73.74 & \textbf{75.55} & \gain{1.81} & 59.39 & \textbf{60.46} & \gain{1.07} & 64.73 & \textbf{65.54} & \gain{0.81} & 66.00 & \textbf{70.40} & \gain{4.40} \\
MMVP & 60.67 & \textbf{63.67} & \gain{3.00} & \textbf{65.00} & 64.00 & \loss{1.00} & 69.00 & \textbf{70.33} & \gain{1.33} & 52.00 & \textbf{54.67} & \gain{2.67} & 55.00 & \textbf{59.00} & \gain{4.00} & 60.00 & \textbf{64.00} & \gain{4.00} \\
CV-B$^{\text{2D}}$ & 52.69 & \textbf{53.13} & \gain{0.44} & 58.69 & \textbf{62.80} & \gain{4.11} & 67.87 & \textbf{69.82} & \gain{1.95} & 51.95 & \textbf{60.50} & \gain{8.55} & 57.93 & \textbf{61.17} & \gain{3.24} & 64.39 & \textbf{65.58} & \gain{1.19} \\
SQA & 72.78 & \textbf{76.45} & \gain{3.67} & 83.75 & \textbf{85.09} & \gain{1.34} & 86.51 & \textbf{89.09} & \gain{2.58} & 69.36 & \textbf{76.00} & \gain{6.64} & 72.73 & \textbf{73.18} & \gain{0.45} & 81.61 & \textbf{83.30} & \gain{1.69} \\
MathV & 35.90 & \textbf{43.10} & \gain{7.20} & 46.20 & \textbf{50.80} & \gain{4.60} & 52.20 & \textbf{55.60} & \gain{3.40} & 34.50 & \textbf{41.30} & \gain{6.80} & 41.60 & \textbf{43.30} & \gain{1.70} & 45.40 & \textbf{48.90} & \gain{3.50} \\
Hallu & 50.26 & \textbf{52.37} & \gain{2.11} & 53.99 & \textbf{54.68} & \gain{0.69} & 56.05 & \textbf{58.04} & \gain{1.99} & 46.37 & \textbf{48.37} & \gain{2.00} & \textbf{53.89} & 51.84 & \loss{2.05} & 53.42 & \textbf{55.95} & \gain{2.53} \\
POPE & 86.87 & \textbf{88.67} & \gain{1.80} & \textbf{86.03} & 85.53 & \loss{0.50} & 90.90 & \textbf{91.23} & \gain{0.33} & 86.03 & \textbf{88.60} & \gain{2.57} & 87.17 & \textbf{90.70} & \gain{3.53} & \textbf{90.50} & 90.30 & \loss{0.20} \\
\midrule[1.0pt]
Average & 46.32 & \textbf{48.74} & \gain{2.42} & 52.13 & \textbf{54.20} & \gain{2.07} & 55.72 & \textbf{57.87} & \gain{2.15} & 43.20 & \textbf{46.32} & \gain{3.12} & 48.07 & \textbf{49.38} & \gain{1.31} & 50.58 & \textbf{53.24} & \gain{2.66} \\
\bottomrule[1.5pt]
\end{tabular}
}
\end{table*}

\begin{table*}
\centering
\caption{\textbf{Scalability of LaVer on datasets.} LaVer demonstrates strong data scaling properties, consistently improving performance across different training dataset sizes.}
\label{tab:data_scale}
\resizebox{\linewidth}{!}{
\begin{tabular}{l|ccc ccc ccc|ccc ccc ccc}
\toprule[1.5pt]
\multirow{3}{*}{\rotatebox[origin=c]{0}{\textbf{Benchmark}}} & \multicolumn{9}{c|}{SigLIP 2} & \multicolumn{9}{c}{CLIP} \\
\cmidrule(lr){2-10} \cmidrule(lr){11-19}
& \multicolumn{3}{c}{800K} & \multicolumn{3}{c}{2M} & \multicolumn{3}{c|}{4M} & \multicolumn{3}{c}{800K} & \multicolumn{3}{c}{2M} & \multicolumn{3}{c}{4M} \\
\cmidrule(lr){2-4} \cmidrule(lr){5-7} \cmidrule(lr){8-10} \cmidrule(lr){11-13} \cmidrule(lr){14-16} \cmidrule(lr){17-19}
& \textbf{Baseline} & \textbf{LaVer} & $\Delta_{\text{Baseline}}$ & \textbf{Baseline} & \textbf{LaVer} & $\Delta_{\text{Baseline}}$ & \textbf{Baseline} & \textbf{LaVer} & $\Delta_{\text{Baseline}}$ & \textbf{Baseline} & \textbf{LaVer} & $\Delta_{\text{Baseline}}$ & \textbf{Baseline} & \textbf{LaVer} & $\Delta_{\text{Baseline}}$ & \textbf{Baseline} & \textbf{LaVer} & $\Delta_{\text{Baseline}}$  \\
\midrule[1.0pt]
GQA & 55.03 & \textbf{56.78} & \gain{1.75} & 53.52 & \textbf{58.29} & \gain{4.77} & 57.54 & \textbf{58.79} & \gain{1.25} & 51.51 & \textbf{54.77} & \gain{3.26} & 53.27 & \textbf{55.53} & \gain{2.26} & 51.01 & \textbf{53.52} & \gain{2.51} \\
MMB$^\text{EN}$ & 73.97 & \textbf{75.60} & \gain{1.63} & 74.83 & \textbf{75.69} & \gain{0.86} & 73.37 & \textbf{76.20} & \gain{2.83} & 68.64 & \textbf{69.93} & \gain{1.29} & 68.81 & \textbf{73.02} & \gain{4.21} & 69.50 & \textbf{76.12} & \gain{6.62} \\
SEED$^{\text{I}}$ & 67.57 & \textbf{68.62} & \gain{1.05} & 67.11 & \textbf{68.15} & \gain{1.04} & 67.61 & \textbf{68.02} & \gain{0.41} & 64.36 & \textbf{65.20} & \gain{0.84} & 63.66 & \textbf{65.70} & \gain{2.04} & 64.58 & \textbf{66.97} & \gain{2.39} \\
MME & 1510.73 & \textbf{1512.50} & \gain{1.77} & 1549.66 & \textbf{1589.66} & \gain{40.00} & 1574.33 & \textbf{1635.16} & \gain{60.83} & 1289.62 & \textbf{1474.65} & \gain{185.03} & 1312.79 & \textbf{1390.06} & \gain{77.27} & 1392.78 & \textbf{1461.37} & \gain{68.59} \\
RWQA & 53.86 & \textbf{59.35} & \gain{5.49} & 60.65 & \textbf{63.53} & \gain{2.88} & 59.61 & \textbf{63.92} & \gain{4.31} & 54.25 & \textbf{56.47} & \gain{2.22} & 54.63 & \textbf{58.04} & \gain{3.41} & 53.86 & \textbf{59.22} & \gain{5.36} \\
MMMU & 44.78 & \textbf{46.33} & \gain{1.55} & 43.56 & \textbf{45.56} & \gain{2.00} & 42.89 & \textbf{45.79} & \gain{2.90} & \textbf{44.56} & \textbf{44.56} & \gain{0.00} & 43.11 & \textbf{44.44} & \gain{1.33} & 41.00 & \textbf{42.44} & \gain{1.44} \\
MM$^*$ & 49.06 & \textbf{52.01} & \gain{2.95} & 50.67 & \textbf{50.74} & \gain{0.07} & 51.54 & \textbf{52.54} & \gain{1.00} & 43.17 & \textbf{45.45} & \gain{2.28} & 46.79 & \textbf{48.85} & \gain{2.06} & 44.85 & \textbf{50.61} & \gain{5.76} \\
OCRB & 536 & \textbf{639} & \gain{103} & 637 & \textbf{678} & \gain{41} & 665 & \textbf{684} & \gain{19} & 306 & \textbf{365} & \gain{59} & 355 & \textbf{387} & \gain{32} & 351 & \textbf{400} & \gain{49} \\
TVQA & 62.06 & \textbf{63.93} & \gain{1.87} & 64.93 & \textbf{68.65} & \gain{3.72} & 67.71 & \textbf{69.11} & \gain{1.40} & 43.86 & \textbf{49.93} & \gain{6.07} & 45.65 & \textbf{54.26} & \gain{8.61} & 52.39 & \textbf{57.73} & \gain{5.34} \\
CQA & 43.52 & \textbf{50.24} & \gain{6.72} & 58.88 & \textbf{64.48} & \gain{5.60} & 61.44 & \textbf{64.88} & \gain{3.44} & 27.36 & \textbf{39.36} & \gain{12.00} & 38.24 & \textbf{45.20} & \gain{6.96} & 42.84 & \textbf{51.04} & \gain{8.20} \\
AI2D & 73.74 & \textbf{75.55} & \gain{1.81} & 73.44 & \textbf{75.22} & \gain{1.78} & 73.15 & \textbf{75.06} & \gain{1.91} & 66.00 & \textbf{70.40} & \gain{4.40} & 66.65 & \textbf{70.98} & \gain{4.33} & 67.97 & \textbf{73.39} & \gain{5.42} \\
MMVP & 69.00 & \textbf{70.33} & \gain{1.33} & 67.00 & \textbf{71.33} & \gain{4.33} & 70.67 & \textbf{72.33} & \gain{1.66} & 60.00 & \textbf{64.00} & \gain{4.00} & 62.67 & \textbf{66.33} & \gain{3.66} & 64.00 & \textbf{69.00} & \gain{5.00} \\
CV-B$^{\text{2D}}$ & 67.87 & \textbf{69.82} & \gain{1.95} & 71.70 & \textbf{71.77} & \gain{0.07} & 72.11 & \textbf{73.35} & \gain{1.24} & 64.39 & \textbf{65.58} & \gain{1.19} & 61.27 & \textbf{64.67} & \gain{3.40} & 62.94 & \textbf{67.66} & \gain{4.72} \\
SQA & 86.51 & \textbf{89.09} & \gain{2.58} & 81.95 & \textbf{86.47} & \gain{4.52} & 84.28 & \textbf{87.01} & \gain{2.73} & 81.61 & \textbf{83.30} & \gain{1.69} & 83.89 & \textbf{84.83} & \gain{0.94} & 84.79 & \textbf{90.88} & \gain{6.09} \\
MathV & 52.20 & \textbf{55.60} & \gain{3.40} & 53.10 & \textbf{56.70} & \gain{3.60} & 53.50 & \textbf{56.70} & \gain{3.20} & 45.40 & \textbf{48.90} & \gain{3.50} & 47.00 & \textbf{51.70} & \gain{4.70} & 49.90 & \textbf{56.70} & \gain{6.80} \\
Hallu & 56.05 & \textbf{58.04} & \gain{1.99} & 60.15 & \textbf{61.41} & \gain{1.26} & 58.36 & \textbf{60.99} & \gain{2.63} & 53.42 & \textbf{55.95} & \gain{2.53} & 56.20 & \textbf{57.41} & \gain{1.21} & \textbf{57.52} & 56.89 & \loss{0.63} \\
POPE & 90.90 & \textbf{91.23} & \gain{0.33} & 91.33 & \textbf{91.47} & \gain{0.14} & \textbf{92.23} & 91.93 & \loss{0.30} & \textbf{90.50} & 90.30 & \loss{0.20} & 88.47 & \textbf{90.97} & \gain{2.50} & \textbf{90.13} & 89.07 & \loss{1.06} \\
\midrule[1.0pt]
Average & 55.72 & \textbf{57.87} & \gain{2.15} & 57.30 & \textbf{59.46} & \gain{2.16} & 58.08 & \textbf{59.88} & \gain{1.80} & 50.58 & \textbf{53.24} & \gain{2.66} & 51.84 & \textbf{54.88} & \gain{3.04} & 52.84 & \textbf{56.60} & \gain{3.77} \\

\bottomrule[1.5pt]
\end{tabular}
}
\end{table*}

\textbf{Visual Feature Homogenization v.s. Visual Feature Inconsistency.} 
To elucidate the fundamental distinctions between these two phenomena, we provide a justification of their divergent characteristics.
While both \textit{visual feature homogenization} and \textit{visual feature inconsistency} manifest the same statistical signature, namely, the generation of visual features exhibiting elevated feature-wise cosine similarity, their underlying causal mechanisms and downstream implications differ substantially.

\textit{Visual feature homogenization} emerges as a direct consequence of the modality imbalance inherent in MLLMs' training paradigm.
Under the text-centric next-token-prediction objective, models systematically exploit textual tokens to generate responses while marginalizing the majority of available visual information.
This preferential reliance on textual modality precipitates a progressive degradation of visual representations, manifesting as increased homogeneity among visual features across layers.
Critically, this phenomenon reflects a fundamental loss of discriminative visual information rather than a mere representational artifact.

In contrast, \textit{visual feature inconsistency} originates from the explicit application of the MIM objective in isolation.
While MIM provides direct supervisory signals that guide the evolution of visual embeddings, it simultaneously introduces an exploitable optimization shortcut: the model can trivially minimize MIM loss by generating nearly identical visual features, thereby achieving low reconstruction error without preserving meaningful visual distinctions.
Consequently, the isolated MIM objective exacerbates rather than ameliorates the situation, failing to provide balanced supervisory signals across modalities and inadvertently reinforcing the collapse toward homogeneous representations.

Our proposed LaVer framework addresses both pathologies synergistically.
By integrating the Clipped Gram-Anchoring mechanism with the MIM objective, LaVer effectively prevents visual feature inconsistency through asymmetric regularization that selectively penalizes excessive homogenization while permitting discriminative feature learning.
Simultaneously, by introducing explicit vision-centric supervisory signals, LaVer mitigates the underlying modality imbalance issue, enabling models to maintain rich, discriminative visual representations throughout their layers.
This dual-pronged approach culminates in substantially enhanced performance across diverse multimodal benchmarks, particularly on tasks demanding dense visual understanding.

\begin{table*}
\centering
\caption{\textbf{Ablation study on masking strategies.} LaVer demonstrates strong robustness across different masking strategies.}
\label{tab:masking_ablation}
\resizebox{\linewidth}{!}{
\begin{tabular}{l|c|cccc|cccc|c|cccc|cccc}
\toprule[1.5pt]
\multirow{3}{*}{\rotatebox[origin=c]{0}{\textbf{Benchmark}}} & \multicolumn{9}{c|}{SigLIP 2} & \multicolumn{9}{c}{CLIP} \\
\cmidrule(lr){2-19}
& \multirow{2}{*}{\rotatebox[origin=c]{0}{\textbf{Baseline}}} & \multicolumn{4}{c|}{Cosine} & \multicolumn{4}{c|}{Constant} & \multirow{2}{*}{\rotatebox[origin=c]{0}{\textbf{Baseline}}} & \multicolumn{4}{c|}{Cosine} & \multicolumn{4}{c}{Constant} \\
\cmidrule(lr){3-6} \cmidrule(lr){7-10} \cmidrule(lr){12-15} \cmidrule(lr){16-19}
& & 0.05 & 0.1 & 0.2 & 0.3 & 0.0002 & 0.01 & 0.05 & 0.1 &  & 0.05 & 0.1 & 0.2 & 0.3 & 0.0002 & 0.01 & 0.05 & 0.1 \\
\midrule[1.0pt]
GQA & 55.03 & \textbf{56.78} & 56.53 & 56.26 & 54.87 & 54.83 & 55.96 & 55.22 & 56.11 & 51.51 & \textbf{54.77} & 54.71 & 54.52 & 52.76 & 52.26 & 53.02 & 52.26 & 51.01 \\
MMB$^\text{EN}$ & 73.97 & 75.60 & \textbf{75.65} & 75.04 & 73.99 & 74.89 & 73.74 & 74.46 & 73.80 & 68.64 & 69.93 & 69.24 & 70.62 & 71.39 & 65.38 & 70.10 & \textbf{72.77} & 70.27 \\
SEED$^{\text{I}}$ & 67.57 & \textbf{68.62} & 68.50 & 68.49 & 67.48 & 67.82 & 68.42 & 68.34 & 67.96 & 64.36 & \textbf{65.20} & 65.02 & 64.27 & 64.99 & 64.67 & 64.21 & 64.81 & 65.03 \\
MME & 1510.73 & 1512.50 & \textbf{1546.12} & 1515.43 & 1534.31 & 1504.69 & 1496.10 & 1532.54 & 1495.53 & 1289.62 & \textbf{1474.65} & 1470.28 & 1407.86 & 1404.61 & 1421.08 & 1381.88 & 1399.67 & 1441.57 \\
RWQA & 53.86 & \textbf{59.35} & 58.77 & 58.62 & 55.51 & 58.33 & 55.18 & 54.55 & 56.17 & 54.25 & \textbf{56.47} & 55.59 & 55.56 & 54.12 & 52.42 & 54.51 & 55.56 & 52.94 \\
MMMU & 44.78 & \textbf{46.33} & 46.31 & 45.96 & 44.63 & 44.85 & 45.77 & 44.57 & 45.95 & 44.56 & 44.56 & \textbf{45.44} & 44.11 & 43.22 & 41.33 & 44.89 & 43.89 & 43.89 \\
MM$^*$ & 49.06 & 52.01 & \textbf{52.04} & 51.44 & 50.12 & 51.31 & 48.97 & 50.71 & 50.95 & 43.17 & 45.45 & 45.31 & 45.25 & \textbf{47.39} & 42.30 & 45.11 & 45.72 & 46.12 \\
OCRB & 536 & \textbf{639} & 638 & 625 & 582 & 619 & 606 & 607 & 597 & 306 & \textbf{365} & 354 & 345 & 301 & 292 & 302 & 300 & 297 \\
TVQA & 62.06 & 63.93 & \textbf{64.11} & 63.98 & 62.07 & 61.79 & 63.59 & 62.37 & 63.92 & 43.86 & \textbf{49.93} & 49.28 & 48.55 & 48.78 & 49.03 & 48.07 & 49.05 & 47.47 \\
CQA & 43.52 & \textbf{50.24} & 46.20 & 48.49 & 48.94 & 46.66 & 46.52 & 46.21 & 48.48 & 27.36 & \textbf{39.36} & 36.32 & 34.00 & 34.24 & 33.52 & 33.20 & 33.36 & 33.04 \\
AI2D & 73.74 & 75.55 & \textbf{75.86} & 74.68 & 73.61 & 74.37 & 73.65 & 74.35 & 73.70 & 66.00 & \textbf{70.40} & 69.46 & 68.62 & 69.59 & 65.38 & 70.14 & 69.95 & 69.24 \\
MMVP & 69.00 & \textbf{70.33} & \textbf{70.33} & 69.33 & 69.00 & 70.00 & 69.67 & 69.33 & 68.33 & 60.00 & \textbf{64.00} & 63.67 & 63.67 & 62.33 & 59.33 & 63.67 & 61.33 & 61.33 \\
CV-B$^{\text{2D}}$ & 67.87 & 69.82 & 68.96 & \textbf{70.06} & 67.67 & 68.87 & 68.68 & 67.97 & 68.40 & 64.39 & \textbf{65.58} & 65.42 & 64.26 & 62.17 & 60.36 & 62.38 & 61.20 & 62.10 \\
SQA & 86.51 & \textbf{89.09} & 86.11 & 88.78 & 88.17 & 87.09 & 86.62 & 86.65 & 87.45 & 81.61 & \textbf{83.30} & 83.29 & 82.15 & 82.90 & 80.47 & 82.40 & 82.60 & 82.30 \\
MathV & 52.20 & 55.60 & 52.16 & \textbf{55.82} & 54.32 & 52.38 & 51.98 & 54.26 & 54.16 & 45.40 & \textbf{48.90} & 46.20 & 45.90 & 47.70 & 46.10 & 46.90 & 47.10 & 46.80 \\
Hallu & 56.05 & 58.04 & 58.29 & \textbf{58.33} & 56.27 & 57.72 & 56.89 & 57.45 & 56.61 & 53.42 & \textbf{55.95} & \textbf{55.95} & 55.10 & 51.00 & 54.26 & 54.00 & 54.89 & 55.63 \\
POPE & 90.90 & 91.23 & \textbf{91.24} & 90.73 & 90.51 & 90.55 & 90.58 & 90.49 & 90.57 & 90.50 & 90.30 & \textbf{90.93} & 90.10 & 90.43 & 89.43 & 89.26 & 90.00 & 89.27 \\
\midrule[1.0pt]
Average & 55.72 & \textbf{57.87} & 57.20 & 57.49 & 56.38 & 56.63 & 56.32 & 56.36 & 56.69 & 50.58 & \textbf{53.24} & 52.75 & 52.21 & 51.99 & 50.42 & 51.93 & 52.08 & 51.61 \\
\bottomrule[1.5pt]
\end{tabular}
}
\end{table*}

\begin{table*}
\centering
\caption{\textbf{Ablation study on EMA teacher strategies.}}
\label{tab:teacher_ablation}
\resizebox{\linewidth}{!}{
\begin{tabular}{l|c|cc|cccc|cccc|c|cc|cccc|cccc}
\toprule[1.5pt]
\multirow{2}{*}{\rotatebox[origin=c]{0}{\textbf{Method}}} & \multicolumn{11}{c|}{SigLIP 2} & \multicolumn{11}{c}{CLIP} \\
\cmidrule(lr){2-12} \cmidrule(lr){13-23}
& Baseline & \multicolumn{10}{c|}{LaVer} & Baseline & \multicolumn{10}{c}{LaVer} \\
\midrule[1.0pt]
EMA Stra. & - & \cellcolor{superlighgray}Con. & \cellcolor{superlighgray}Cos. & Cos. & Cos. & Cos. & Cos. & Cos. & Cos. & Cos. & Cos. & - & \cellcolor{superlighgray}Con. & \cellcolor{superlighgray}Cos. & Cos. & Cos. & Cos. & Cos. & Cos. & Cos. & Cos. & Cos. \\
EMA Frq. & - & 100 & 100 & \cellcolor{superlighgray}1 & \cellcolor{superlighgray}50 & \cellcolor{superlighgray}100 & \cellcolor{superlighgray}200 & 100 & 100 & 100 & 100 & - & 100 & 100 & \cellcolor{superlighgray}1 & \cellcolor{superlighgray}50 & \cellcolor{superlighgray}100 & \cellcolor{superlighgray}200 & 100 & 100 & 100 & 100  \\
EMA Ratio & - & 0.95 & 0.95 & 0.95 & 0.95 & 0.95 & 0.95 & \cellcolor{superlighgray}0.9 & \cellcolor{superlighgray}0.95 & \cellcolor{superlighgray}0.99 & \cellcolor{superlighgray}0.999 & - & 0.95 & 0.95 & 0.95 & 0.95 & 0.95 & 0.95 & \cellcolor{superlighgray}0.9 & \cellcolor{superlighgray}0.95 & \cellcolor{superlighgray}0.99 & \cellcolor{superlighgray}0.999 \\
\midrule[1.0pt]
GQA & 55.03 & 55.63 & 56.78 & 53.65 & 55.22 & 56.78 & \textbf{57.19} & 56.44 & 56.78 & 55.90 & 56.07 & 51.51 & 51.72 & 54.77 & 50.91 & 51.39 & 54.77 & \textbf{55.03} & 54.09 & 54.77 & 54.00 & 53.97 \\
MMB$^\text{EN}$ & 73.97 & \textbf{77.39} & 75.60 & 70.44 & 73.70 & 75.60 & 73.43 & 75.21 & 75.60 & 75.12 & 75.70 & 68.64 & 70.99 & 69.93 & 65.54 & 70.56 & 69.93 & 70.58 & 71.19 & 69.93 & \textbf{71.22} & 69.57 \\
SEED$^{\text{I}}$ & 67.57 & 67.82 & 68.62 & 64.39 & 69.43 & 68.62 & 69.49 & \textbf{69.57} & 68.62 & 69.16 & 69.36 & 64.36 & 66.71 & 65.20 & 61.54 & 64.43 & 65.20 & 65.96 & 63.81 & 65.20 & 66.75 & \textbf{66.92} \\
MME & 1510.73 & 1546.52 & 1512.50 & 1436.02 & \textbf{1574.55} & 1512.50 & 1521.53 & 1544.95 & 1512.50 & 1501.03 & 1507.67 & 1289.62 & 1393.95 & 1474.65 & 1361.29 & 1315.01 & 1474.65 & 1366.78 & 1328.63 & 1474.65 & 1462.85 & \textbf{1481.98} \\
RWQA & 53.86 & 55.56 & \textbf{59.35} & 55.31 & 53.34 & \textbf{59.35} & 56.42 & 55.90 & \textbf{59.35} & 56.46 & 55.91 & 54.25 & 55.10 & 56.47 & 53.34 & 54.79 & 56.47 & 54.43 & 56.01 & 56.47 & \textbf{56.70} & 55.35 \\
MMMU & 44.78 & 45.48 & \textbf{46.33} & 43.21 & 45.85 & \textbf{46.33} & 45.18 & 44.81 & \textbf{46.33} & 44.69 & 44.57 & 44.56 & 44.96 & 44.56 & 42.33 & 44.24 & 44.56 & 45.43 & \textbf{45.86} & 44.56 & 44.89 & 45.02 \\
MM$^*$ & 49.06 & 50.49 & \textbf{52.01} & 49.32 & 48.84 & \textbf{52.01} & 50.35 & 49.36 & \textbf{52.01} & 51.74 & 51.93 & 43.17 & 44.80 & 45.45 & 42.18 & 45.17 & 45.45 & 43.57 & 45.79 & 45.45 & 46.23 & \textbf{46.50} \\
OCRB & 536 & \textbf{651} & 639 & 537 & 633 & 639 & 637 & 577 & 639 & 649 & 565 & 306 & 351 & \textbf{365} & 298 & 355 & \textbf{365} & 357 & \textbf{365} & \textbf{365} & 337 & 362 \\
TVQA & 62.06 & \textbf{65.79} & 63.93 & 59.38 & 62.33 & 63.93 & 65.27 & 62.29 & 63.93 & 63.97 & 65.08 & 43.86 & 45.39 & \textbf{49.93} & 46.66 & 47.42 & \textbf{49.93} & 45.40 & 46.76 & \textbf{49.93} & 46.04 & 48.64 \\
CQA & 43.52 & 47.00 & \textbf{50.24} & 44.98 & 45.54 & \textbf{50.24} & 49.01 & 47.87 & \textbf{50.24} & 49.66 & 47.48 & 27.36 & \textbf{40.53} & 39.36 & 30.08 & 36.81 & 39.36 & 33.66 & 37.27 & 39.36 & 37.22 & 40.24 \\
AI2D & 73.74 & 75.68 & 75.55 & 70.87 & 75.36 & 75.55 & \textbf{75.76} & 74.53 & 75.55 & 75.01 & 73.38 & 66.00 & 65.59 & \textbf{70.40} & 63.35 & 65.96 & \textbf{70.40} & 67.99 & 68.59 & \textbf{70.40} & 69.86 & 69.72 \\
MMVP & 69.00 & \textbf{72.67} & 70.33 & 66.00 & 71.00 & 70.33 & 72.00 & 71.33 & 70.33 & 72.33 & 70.67 & 60.00 & 62.00 & 64.00 & 59.67 & \textbf{64.67} & 64.00 & 63.00 & 62.33 & 64.00 & 62.00 & 63.33 \\
CV-B$^{\text{2D}}$ & 67.87 & 69.08 & 69.82 & 66.10 & \textbf{70.44} & 69.82 & 70.05 & 67.29 & 69.82 & 69.30 & 69.58 & 64.39 & 67.42 & 65.58 & 61.70 & 65.54 & 65.58 & 67.09 & 65.45 & 65.58 & \textbf{67.52} & 66.21 \\
SQA & 86.51 & 89.59 & 89.09 & 82.90 & 88.33 & 89.09 & 87.24 & \textbf{90.41} & 89.09 & 86.34 & 86.86 & 81.61 & 85.17 & 83.30 & 77.91 & 84.23 & 83.30 & \textbf{85.23} & 82.95 & 83.30 & 82.39 & 82.25 \\
MathV & 52.20 & \textbf{55.73} & 55.60 & 52.46 & 55.06 & 55.60 & 54.23 & 53.39 & 55.60 & 55.68 & 52.74 & 45.40 & \textbf{50.01} & 48.90 & 45.75 & 49.20 & 48.90 & 47.51 & 46.63 & 48.90 & 47.86 & 48.47 \\
Hallu & 56.05 & 56.72 & 58.04 & 54.29 & 57.15 & 58.04 & 56.55 & \textbf{59.46} & 58.04 & 59.13 & 56.13 & 53.42 & \textbf{57.28} & 55.95 & 52.25 & 55.32 & 55.95 & 56.37 & 53.10 & 55.95 & 54.98 & 56.32 \\
POPE & 90.90 & 90.44 & 91.23 & 86.37 & 90.21 & 91.23 & 91.64 & 91.61 & 91.23 & 90.07 & \textbf{91.97} & 90.50 & 90.96 & 90.30 & 85.97 & 91.77 & 90.30 & \textbf{92.56} & 90.03 & 90.30 & 91.09 & 91.68 \\
\midrule[1.0pt]
Average & 55.72 & 57.43 & \textbf{57.87} & 54.17 & 56.65 & \textbf{57.87} & 57.36 & 57.10 & \textbf{57.87} & 57.40 & 56.98 & 50.58 & 52.92 & 53.24 & 49.41 & 52.49 & 53.24 & 52.63 & 52.40 & 53.24 & 52.92 & \textbf{53.25} \\
\bottomrule[1.5pt]
\end{tabular}
}
\end{table*}

\begin{table*}
\centering
\caption{\textbf{Ablation study on spatial awareness.}}
\label{tab:spatial_ablation}
\resizebox{\linewidth}{!}{
\begin{tabular}{c|ccc|ccccccccccccccccc|c}
\toprule[1.5pt]
$\mathcal{G}_{\upxi}$ & LaVer & Mixed Attn. & 2D-RoPE & GQA & MMB$^{\text{EN}}$ & SEED$^{\text{I}}$ & MME & RWQA & MMMU & MM$^*$ & OCRB & TVQA & CQA & AI2D & MMVP & CV-B$^{\text{2D}}$ & SQA & MathV & Hallu & POPE & Avg. \\
\midrule[1.0pt]
\multirow{5}{*}{SigLIP 2} & \textcolor{Grey}{\ding{55}} &  \textcolor{Grey}{\ding{55}} & \textcolor{Grey}{\ding{55}} & 55.03 & 73.97 & 67.57 & 1510.73 & 53.86 & 44.78 & 49.06 & 536 & 62.06 & 43.52 & 73.74 & 69.00 & 67.87 & 86.51 & 52.20 & 56.05 & 90.90 & 55.72 \\
& \textcolor{Grey}{\ding{55}} & \ding{51} & \textcolor{Grey}{\ding{55}} & 54.33 & 74.98 & 68.31 & 1505.55 & 55.17 & 45.32 & 49.54 & \textbf{641} & 61.40 & 45.38 & 75.94 & \textbf{71.00} & \textbf{70.62} & 88.62 & \textbf{56.05} & 55.83 & \textbf{91.57} & 56.78 \\
& \textcolor{Grey}{\ding{55}} & \textcolor{Grey}{\ding{55}} & \ding{51} & 54.87 & 73.69 & 68.45 & 1509.39 & 52.90 & 43.99 & 48.89 & 541 & 61.72 & 43.98 & 73.97 & 68.33 & 69.14 & 86.41 & 51.46 & 56.17 & 89.56 & 55.57 \\
& \textcolor{Grey}{\ding{55}} & \ding{51} & \ding{51} & 54.07 & 74.02 & 66.82 & 1481.78 & 58.01 & 44.73 & 51.11 & 605 & 62.13 & 46.71 & \textbf{76.18} & 69.00 & 66.81 & 87.81 & 53.20 & 57.30 & 90.20 & 56.43 \\
& \ding{51} & \ding{51} & \ding{51} & \textbf{56.78} & \textbf{75.60} & \textbf{68.62} & \textbf{1512.50} & \textbf{59.35} & \textbf{46.33} & \textbf{52.01} & 639 & \textbf{63.93} & \textbf{50.24} & 75.55 & 70.33 & 69.82 & \textbf{89.09} & 55.60 & \textbf{58.04} & 91.23 & \textbf{57.87} \\
\midrule[1.0pt]
\multirow{5}{*}{CLIP} & \textcolor{Grey}{\ding{55}} &  \textcolor{Grey}{\ding{55}} & \textcolor{Grey}{\ding{55}} & 51.51 & 68.64 & 64.36 & 1289.62 & 54.25 & \textbf{44.56} & 43.17 & 306 & 43.86 & 27.36 & 66.00 & 60.00 & 64.39 & 81.61 & 45.40 & 53.42 & 90.50 & 50.58 \\
& \textcolor{Grey}{\ding{55}} & \ding{51} & \textcolor{Grey}{\ding{55}} & 51.71 & 69.00 & 63.84 & 1335.01 & 53.71 & 41.72 & 44.14 & 310 & 48.69 & 34.99 & 69.93 & 62.00 & 63.38 & 80.77 & 45.82 & 53.36 & 89.82 & 51.40 \\
& \textcolor{Grey}{\ding{55}} & \textcolor{Grey}{\ding{55}} & \ding{51} & 51.54 & 68.98 & 64.98 & 1302.20 & 54.04 & 42.02 & 42.96 & 310 & 44.33 & 27.44 & 65.37 & 61.67 & 64.69 & 82.33 & 45.37 & 53.46 & 90.05 & 50.59 \\
& \textcolor{Grey}{\ding{55}} & \ding{51} & \ding{51} & 52.76 & \textbf{69.99} & 65.15 & 1327.62 & 53.59 & 42.89 & 44.78 & 313 & 48.77 & 35.76 & 69.75 & 63.00 & 64.46 & 82.20 & 45.50 & 53.73 & \textbf{90.66} & 51.99 \\
& \ding{51} & \ding{51} & \ding{51} & \textbf{54.77} & 69.93 & \textbf{65.20} & \textbf{1474.65} & \textbf{56.47} & \textbf{44.56} & \textbf{45.45} & \textbf{365} & \textbf{49.93} & \textbf{39.36} & \textbf{70.40} & \textbf{64.00} & \textbf{65.58} & \textbf{83.30} & \textbf{48.90} & \textbf{55.95} & 90.30 & \textbf{53.24} \\
\bottomrule[1.5pt]
\end{tabular}
}
\end{table*}

\begin{table*}
\centering
\caption{\textbf{Ablation study on loss functions.}}
\label{tab:loss_ablation}
\resizebox{\linewidth}{!}{
\begin{tabular}{c|ccc|ccccccccccccccccc|c}
\toprule[1.5pt]
$\mathcal{G}_{\upxi}$ & +$\mathcal{L}_{\text{MIM}}$ & +$\mathcal{L}_{\text{GA}}$ & +$\mathcal{L}_{\text{CGA}}$ & GQA & MMB$^{\text{EN}}$ & SEED$^{\text{I}}$ & MME & RWQA & MMMU & MM$^*$ & OCRB & TVQA & CQA & AI2D & MMVP & CV-B$^{\text{2D}}$ & SQA & MathV & Hallu & POPE & Avg. \\
\midrule[1.0pt]
\multirow{4}{*}{SigLIP 2} & \textcolor{Grey}{\ding{55}}  & \textcolor{Grey}{\ding{55}}  & \textcolor{Grey}{\ding{55}} & 55.03 & 73.97 & 67.57 & 1510.73 & 53.86 & 44.78 & 49.06 & 536 & 62.06 & 43.52 & 73.74 & 69.00 & 67.87 & 86.51 & 52.20 & 56.05 & 90.90 & 55.72 \\
& \ding{51} & \textcolor{Grey}{\ding{55}}  & \textcolor{Grey}{\ding{55}}  & 55.02 & 71.97 & 64.59 & 1500.80 & 51.38 & 44.35 & 47.41 & 512 & 60.40 & 42.20 & 70.35 & 66.00 & 64.22 & 82.57 & 52.72 & 54.11 & 85.53 & 53.76 \\
& \ding{51} & \ding{51} & \textcolor{Grey}{\ding{55}} & 54.65 & 73.56 & 65.84 & 1456.21 & 58.97 & 45.21 & \textbf{52.09} & 617 & 63.07 & 49.07 & 72.26 & 67.67 & 68.94 & 85.93 & \textbf{55.89} & 55.91 & 89.49 & 56.46 \\
& \ding{51} & \textcolor{Grey}{\ding{55}}  & \ding{51} & \textbf{56.78} & \textbf{75.60} & \textbf{68.62} & \textbf{1512.50} & \textbf{59.35} & \textbf{46.33} & 52.01 & \textbf{639} & \textbf{63.93} & \textbf{50.24} & \textbf{75.55} & \textbf{70.33} & \textbf{69.82} & \textbf{89.09} & 55.60 & \textbf{58.04} & \textbf{91.23} & \textbf{57.87} \\
\midrule[1.0pt]
\multirow{4}{*}{CLIP} & \textcolor{Grey}{\ding{55}}  & \textcolor{Grey}{\ding{55}}  & \textcolor{Grey}{\ding{55}} & 51.51 & 68.64 & 64.36 & 1289.62 & 54.25 & \textbf{44.56} & 43.17 & 306 & 43.86 & 27.36 & 66.00 & 60.00 & 64.39 & 81.61 & 45.40 & 53.42 & \textbf{90.50} & 50.58 \\
& \ding{51} & \textcolor{Grey}{\ding{55}}  & \textcolor{Grey}{\ding{55}} & 48.48 & 66.08 & 63.93 & 1222.30 & 52.14 & 42.42 & 43.24 & 290 & 43.89 & 26.52 & 64.28 & 60.67 & 61.27 & 83.16 & 45.04 & 54.44 & 88.77 & 49.71 \\
& \ding{51} & \ding{51} & \textcolor{Grey}{\ding{55}} & 54.72 & 68.16 & 63.26 & 1431.67 & 53.80 & 43.78 & 44.78 & 359 & 49.14 & 37.83 & \textbf{70.67} & \textbf{64.67} & \textbf{65.68} & 79.36 & 46.46 & 53.91 & 86.98 & 52.01 \\
& \ding{51} & \textcolor{Grey}{\ding{55}}  & \ding{51} & \textbf{54.77} & \textbf{69.93} & \textbf{65.20} & \textbf{1474.65} & \textbf{56.47} & \textbf{44.56} & \textbf{45.45} & \textbf{365} & \textbf{49.93} & \textbf{39.36} & 70.40 & 64.00 & 65.58 & \textbf{83.30} & \textbf{48.90} & \textbf{55.95} & 90.30 & \textbf{53.24} \\
\bottomrule[1.5pt]
\end{tabular}
}
\end{table*}

\section{Implementation Details}

\textbf{Hyperparameters.} 
We summarize the hyperparameters for each training stage in Table~\ref{tab:hyper}. Following common practice in LLaVA-OneVision 1.5~\cite{an2025llavaonevision15fullyopenframework}, we adopt standard configurations for our three-stage training pipeline. Note that we do not use the exact datasets from LLaVA-OneVision 1.5, as they were not fully available at the time of our experiments. LaVer is applied exclusively to Stage 2, where visual knowledge is injected into the model.
The specific hyperparameters of LaVer are detailed in Table~\ref{tab:hyper_laver}. We observe that LaVer's performance is robust to most hyperparameter choices, requiring minimal tuning. Through comprehensive ablation studies on masking strategies and EMA updating strategies, we demonstrate LaVer's robustness to hyperparameter variations. Our findings indicate that a small masking ratio with cosine scheduling is sufficient for effective learning. For EMA updates, we find that the strategy is insensitive to the update schedule as long as the update frequency is not excessively high; overly frequent updates can cause the model to exploit the MIM loss by inadvertently propagating visual feature inconsistencies into the teacher model.
For the vision head architecture, we employ a lightweight 3-layer MLP with 8192 hidden dimensions, parallel to the language head. This design introduces only a negligible number of additional trainable parameters. Following the established practice in~\cite{zhou2021ibot}, we set the teacher temperature $\tau_{\text{tea.}} = 0.04$ and student temperature $\tau_{\text{stu.}} = 0.1$, which ensures stable convergence throughout training.
For fair comparison with ROSS~\cite{wang2025reconstructive}, we adopt their 2-stage training protocol with identical configurations as specified in~\cite{wang2025reconstructive}. All experiments are conducted on 16 NVIDIA A100 GPUs with 80GB memory each.

\textbf{Datasets.}
Our training pipeline employs off-the-shelf datasets across three stages.
For Stage 1, we adopt the LLaVA-558K dataset~\cite{liu2023llava} for vision-language alignment.
For Stage 2, we randomly sample 800K image-text pairs from FineVision 23M~\cite{wiedmann2025finevisionopendataneed}, maintaining the original dataset's proportions to preserve its diverse visual knowledge sources. Due to computational constraints, we do not utilize the complete dataset; however, we validate LaVer's data scaling properties by sampling up to 4M pairs from FineVision.
For Stage 3, we randomly sample 800K instruction-tuning pairs from LLaVA-OneVision 4M~\cite{li2025llavaonevision}, again preserving the original proportions. While potential overlap may exist between Stage 2 and Stage 3 data, we retain all samples as this configuration has proven empirically effective.
For fair comparison with ROSS~\cite{wang2025reconstructive}, we adopt their 2-stage protocol using LLaVA-558K~\cite{liu2023llava} for Stage 1 and Cambrian-737K~\cite{10.5555/3737916.3740687} for Stage 2.

\textbf{Vision encoders.}
We evaluate LaVer across diverse vision encoders to demonstrate its broad applicability.
\textbf{SigLIP 2}~\cite{tschannen2025siglip2multilingualvisionlanguage} employs pairwise sigmoid loss instead of softmax for enhanced image-text alignment and multilingual capabilities. For our main experiments, we adopt SigLIP 2-ViT-SO400M/14@384, which encodes $384\times 384$ images into 729 vision tokens. For comparison with ROSS~\cite{wang2025reconstructive}, we use SigLIP-ViT-SO400M/14@384~\cite{Zhai_2023_ICCV} to ensure fair evaluation.
\textbf{CLIP}~\cite{pmlr-v139-radford21a} is trained with contrastive loss for vision-language alignment. We adopt CLIP-ViT-L/14@336, which processes $336\times 336$ images into 576 vision tokens.
\textbf{DINOv2}~\cite{siméoni2025dinov3} leverages self-contrastive and self-distillation learning for visual feature extraction. We use DINOv2-Large/14@224, encoding $224\times 224$ images into 384 vision tokens.
\textbf{AIMv2}~\cite{fini2025multimodal} is a native-resolution encoder trained via autoregressive pixel-wise prediction with an auxiliary LLM backbone. We adopt AIMv2-Large/14, which patchifies images into $14\times 14$ patches, with images resized to a maximum of $224\times 224$ pixels.
\textbf{Qwen-ViT}~\cite{bai2025qwen25vltechnicalreport} serves as the native-resolution vision encoder in Qwen2.5-VL with patch size 14. We utilize the encoder from Qwen2.5-VL-7B-Instruct and set the maximum resolution to $512\times 512$ pixels.
\textbf{Encoder-free architecture.} To validate LaVer's generalizability beyond traditional vision encoders, we adopt an encoder-free architecture~\cite{lei2025sail,chen2024a,10.5555/3737916.3739581}. We employ a 3-layer MLP with 3584 intermediate hidden dimensions to project images into visual tokens with patch size 16, supporting up to $512\times 512$ pixels. During Stage 1, only the MLP is trained; in Stages 2 and 3, both the MLP projector and LLM backbone are jointly optimized.

\textbf{Packed visual sequence.}
To improve training efficiency while maintaining independent visual reconstruction without interfering with multimodal sequence modeling, we pack vision tokens from multiple images into a single sequence, excluding their corresponding text tokens.
Specifically, we construct diagonally blocked bidirectional attention and concatenated 2D-RoPE for the packed visual sequences.
Across all vision encoders, we pack vision tokens from 2 images into a single sequence of length 2048 with padding tokens, yielding two separate visual sequences per local batch on each GPU with batch size 4.
The attention and positional embedding mechanisms for packed visual sequences are illustrated in Fig.~\ref{fig:packed_vision}.

\textbf{Evaluation.}
We conduct comprehensive evaluation using the VLMEvalKit~\cite{10.1145/3664647.3685520} toolbox.
Our evaluation suite encompasses a diverse set of benchmarks: GQA~\cite{hudson2019gqanewdatasetrealworld} for compositional visual reasoning, MMBench (MMB$^{\text{EN}}$)~\cite{10.1007/978-3-031-72658-3_13} for comprehensive multimodal understanding, SEED-Image (SEED$^{\text{I}}$)~\cite{li2023seedbenchbenchmarkingmultimodalllms} for generative comprehension, MME~\cite{fu2025mmecomprehensiveevaluationbenchmark} for perception and cognition evaluation, RealWorldQA (RWQA)~\cite{xai2024grok15visionpreview} for real-world visual question answering, MMMU~\cite{yue2024mmmumassivemultidisciplinemultimodal} for massive multi-discipline understanding, MMStar (MM$^*$)~\cite{chen2024rightwayevaluatinglarge} for challenging multi-modal reasoning, OCR-Bench (OCRB)~\cite{Liu_2024} for text recognition capabilities, TextVQA (TVQA)~\cite{singh2019vqamodelsread} for reading text in images, ChartQA (CQA)~\cite{masry-etal-2022-chartqa} for chart understanding, AI2D~\cite{Kembhavi2016ADI} for diagram comprehension, CV-Bench-2D (CV-B$^{\text{2D}}$)~\cite{10.5555/3737916.3740687} for vision-centric capabilities, MMVP~\cite{10655378} for visual perception, ScienceQA (SQA)~\cite{lu2022learnexplainmultimodalreasoning} for science question answering with explanations, MathVista (MathV)~\cite{lu2024mathvistaevaluatingmathematicalreasoning} for mathematical reasoning in visual contexts, HallucinationBench (Hallu)~\cite{guan2024hallusionbenchadvanceddiagnosticsuite} for hallucination detection, and POPE~\cite{li-etal-2023-evaluating} for object hallucination evaluation.
We employ the default system prompt and instruction template for each benchmark to ensure fair comparison.
For averaged results, we compute the mean value across all benchmarks, with MME and OCR-Bench normalized to the range $[0.0, 1.0]$ to maintain consistent scaling.

For ReasonSeg~\cite{lai2024lisareasoningsegmentationlarge} evaluation, we adopt the fine-tuning protocol established in~\cite{tang2025ufounifiedapproachfinegrained}.
Specifically, we initialize models using checkpoints from stage 2 for both the baseline and LaVer configurations.
Following~\cite{tang2025ufounifiedapproachfinegrained}, we fine-tune the models to acquire sophisticated segmentation reasoning capabilities using a comprehensive dataset mixture comprising semantic segmentation datasets (COCOStuff~\cite{caesar2018cocostuffthingstuffclasses}, Mapillary~\cite{8237796}, ADE20K~\cite{8100027}, nuScenes~\cite{caesar2020nuscenesmultimodaldatasetautonomous}), referring segmentation datasets (RefCOCO~\cite{caesar2020nuscenesmultimodaldatasetautonomous}, RefCOCO+~\cite{caesar2020nuscenesmultimodaldatasetautonomous}, RefCOCOg~\cite{caesar2020nuscenesmultimodaldatasetautonomous}), and the general VQA dataset LLaVA-665K~\cite{liu2023improvedllava}, totaling approximately 4M samples.
The detailed training configuration is presented in Table~\ref{tab:reasonseg_config}, with hyperparameters primarily adopted from~\cite{tang2025ufounifiedapproachfinegrained}.

\textbf{Computational overhead analysis.}
The implementation of LaVer introduces additional computational overhead, primarily stemming from the forward passes required for both student and teacher models to process image tokens, as well as the EMA updates for the teacher model parameters.
As detailed in Table~\ref{tab:comput_cost}, we conduct a comprehensive computational cost analysis focusing on stage 2, where LaVer is applied (stages 1 and 3 remain identical to the baseline).
Across different vision encoders, LaVer incurs training time increases of 13-16\% and memory consumption increases of 14-26\% compared to the baseline.
Specifically, with SigLIP 2~\cite{tschannen2025siglip2multilingualvisionlanguage}, training time increases from 8h 27min to 9h 34min, while memory consumption rises from 55.24 GB to 69.82 GB per GPU (averaged across 16 GPUs).
Similar patterns are observed with CLIP~\cite{pmlr-v139-radford21a} and DINOv2~\cite{oquab2024dinov} vision encoders.
Despite these additional costs, the substantial performance improvements demonstrated across diverse benchmarks and architectural configurations justify this computational overhead, establishing LaVer as a practical and effective approach for enhancing MLLMs.

\begin{table}[t]
\centering
\caption{\textbf{Language performance comparison.}}
\label{tab:lang_performance}
\resizebox{0.8\linewidth}{!}{
\begin{tabular}{l|ccc|ccc}
\toprule[1.5pt]
\multirow{2}{*}{Benchmark} & \multicolumn{3}{c|}{SigLIP 2} & \multicolumn{3}{c}{CLIP} \\
\cmidrule(lr){2-4} \cmidrule(lr){5-7}
& Baseline & LaVer & $\Delta_{\text{Baseline}}$ & Baseline & LaVer & $\Delta_{\text{Baseline}}$ \\
\midrule[1pt]
IFEval & 70.34 & \textbf{70.77} & \gain{0.43} & \textbf{68.96} & 68.24 & \loss{0.72}  \\
MMLU & 56.52 & \textbf{57.48} & \gain{0.96} & 58.34 & \textbf{58.67} & \gain{0.33} \\
BBH & \textbf{33.25} & 32.96 & \loss{0.29} & 34.78 & \textbf{34.93} & \gain{0.15} \\
\midrule[1pt]
Average & 53.37 & \textbf{53.74} & \gain{0.37} & \textbf{54.03} & 53.95 & \loss{0.08} \\
\bottomrule[1.5pt]
\end{tabular}
}
\end{table}

\begin{table*}[t]
\centering
\caption{\textbf{Performance comparison with different LLM backbone.} LaVer achieves superior performance on Vicuna-7B-v1.5~\cite{zheng2023judging}, demonstrating its effectiveness and generalizability across diverse architectural configurations.}
\label{tab:vicuna}
\resizebox{0.6\linewidth}{!}{
\begin{tabular}{l|ccc|ccc|ccc}
\toprule[1.5pt]
\multirow{2}{*}{Benchmark} & \multicolumn{3}{c|}{SigLIP 2} & \multicolumn{3}{c|}{CLIP} & \multicolumn{3}{c}{DINOv2} \\
\cmidrule(lr){2-4} \cmidrule(lr){5-7} \cmidrule(lr){8-10}
& Baseline & LaVer & $\Delta_{\text{Baseline}}$ & Baseline & LaVer & $\Delta_{\text{Baseline}}$ & Baseline & LaVer & $\Delta_{\text{Baseline}}$\\
\midrule[1pt]
GQA & 50.00 & \textbf{53.52} & \gain{3.52} & 51.01 & \textbf{54.02} & \gain{3.01} & 49.75 & \textbf{51.76} & \gain{2.01} \\
MMB$^\text{EN}$ & 67.83 & \textbf{68.55} & \gain{0.72} & \textbf{66.50} & 65.13 & \loss{1.37} & 53.69 & \textbf{55.29} & \gain{1.60} \\
SEED$^{\text{I}}$ & 66.33 & \textbf{67.29} & \gain{0.96} & 64.58 & \textbf{65.36} & \gain{0.78} & 60.09 & \textbf{63.77} & \gain{3.68} \\
MME & 1426.52 & \textbf{1441.46} & \gain{14.94} & 1292.78 & \textbf{1317.35} & \gain{24.57} & 1232.67 & \textbf{1248.07} & \gain{15.40} \\
RWQA & 55.42 & \textbf{58.43} & \gain{3.01} & 53.86 & \textbf{55.56} & \gain{1.70} & 45.75 & \textbf{48.55} & \gain{2.80} \\
MMMU & 40.89 & \textbf{42.11} & \gain{1.22} & 41.00 & \textbf{44.33} & \gain{3.33} & 40.33 & \textbf{41.33} & \gain{1.00} \\
MM$^*$ & 51.74 & \textbf{53.28} & \gain{1.54} & 44.85 & \textbf{45.38} & \gain{0.53} & 40.82 & \textbf{44.24} & \gain{3.42} \\
OCRB & 389 & \textbf{401} & \gain{12} & 311 & \textbf{352} & \gain{41} & 306 & \textbf{349} & \gain{43} \\
TVQA & 65.07 & \textbf{67.36} & \gain{2.29} & 52.39 & \textbf{52.60} & \gain{0.21} & 40.66 & \textbf{42.25} & \gain{1.59} \\
CQA & 62.48 & \textbf{64.80} & \gain{2.32} & 37.84 & \textbf{44.56} & \gain{6.72} & 28.44 & \textbf{29.04} & \gain{0.60} \\
AI2D & 70.84 & \textbf{72.68} & \gain{1.84} & 67.97 & \textbf{69.09} & \gain{1.12} & 61.14 & \textbf{63.09} & \gain{1.95} \\
MMVP & 47.67 & \textbf{49.00} & \gain{1.33} & 41.00 & \textbf{44.67} & \gain{3.67} & 44.33 & \textbf{46.67} & \gain{2.34} \\
CV-B$^{\text{2D}}$ & 54.19 & \textbf{56.55} & \gain{2.36} & 42.94 & \textbf{44.45} & \gain{1.51} & 41.47 & \textbf{45.49} & \gain{4.02} \\
SQA & 73.89 & \textbf{75.08} & \gain{1.19} & \textbf{70.79} & 70.37 & \loss{0.42} & 62.93 & \textbf{65.33} & \gain{2.40} \\
MathV & 54.20 & \textbf{56.80} & \gain{2.60} & 43.90 & \textbf{46.40} & \gain{2.50} & 40.80 & \textbf{43.30} & \gain{2.50} \\
Hallu & 66.89 & \textbf{67.83} & \gain{0.94} & 67.52 & \textbf{70.25} & \gain{2.73} & 60.26 & \textbf{69.74} & \gain{9.48} \\
POPE & 87.07 & \textbf{88.07} & \gain{1.00} & 90.13 & \textbf{90.73} & \gain{0.60} & 86.90 & \textbf{87.67} & \gain{0.77} \\
\midrule[1.0pt]
Average & 53.85 & \textbf{55.43} & \gain{1.58} & 49.24 & \textbf{50.81} & \gain{1.57} & 44.60 & \textbf{46.96} & \gain{2.37} \\
\bottomrule[1.5pt]
\end{tabular}
}
\end{table*}

\begin{table*}[t]
\centering
\caption{\textbf{Performance comparison of stage 2.} LaVer achieves superior performance on stage 2, indicating a direct effect of LaVer on visual representation enhancement.}
\label{tab:stage2}
\resizebox{0.6\linewidth}{!}{
\begin{tabular}{l|ccc|ccc|ccc}
\toprule[1.5pt]
\multirow{2}{*}{Benchmark} & \multicolumn{3}{c|}{SigLIP 2} & \multicolumn{3}{c|}{CLIP} & \multicolumn{3}{c}{DINOv2} \\
\cmidrule(lr){2-4} \cmidrule(lr){5-7} \cmidrule(lr){8-10}
& Baseline & LaVer & $\Delta_{\text{Baseline}}$ & Baseline & LaVer & $\Delta_{\text{Baseline}}$ & Baseline & LaVer & $\Delta_{\text{Baseline}}$\\
\midrule[1pt]
GQA & 52.03 & \textbf{53.77} & \gain{1.74} & 50.25 & \textbf{53.28} & \gain{3.03} & 48.01 & \textbf{51.26} & \gain{3.25} \\
MMB$^\text{EN}$ & 70.51 & \textbf{73.52} & \gain{3.01} & 67.04 & \textbf{67.53} & \gain{0.49} & 57.82 & \textbf{60.77} & \gain{2.95} \\
SEED$^{\text{I}}$ & 65.63 & \textbf{66.92} & \gain{1.29} & 58.94 & \textbf{64.55} & \gain{5.61} & 60.74 & \textbf{61.91} & \gain{1.17} \\
MME & 1387.81 & \textbf{1434.21} & \gain{46.40} & 1215.99 & \textbf{1313.32} & \gain{97.33} & 1169.80 & \textbf{1224.12} & \gain{54.32} \\
RWQA & 52.43 & \textbf{58.74} & \gain{6.31} & 52.38 & \textbf{54.76} & \gain{2.38} & 48.42 & \textbf{52.29} & \gain{3.87} \\
MMMU & 42.67 & \textbf{44.67} & \gain{2.00} & 42.44 & \textbf{42.89} & \gain{0.45} & 40.22 & \textbf{41.67} & \gain{1.45} \\
MM$^*$ & 45.11 & \textbf{51.41} & \gain{6.30} & 41.63 & \textbf{44.25} & \gain{2.62} & 39.43 & \textbf{41.10} & \gain{1.67} \\
OCRB & 304 & \textbf{407} & \gain{103} & 288 & \textbf{306} & \gain{18} & 247 & \textbf{258} & \gain{11} \\
TVQA & 54.94 & \textbf{59.99} & \gain{5.05} & 60.92 & \textbf{62.35} & \gain{1.43} & 52.44 & \textbf{54.97} & \gain{2.53} \\
CQA & 51.60 & \textbf{54.16} & \gain{2.56} & 40.40 & \textbf{46.80} & \gain{6.40} & 40.40 & \textbf{41.20} & \gain{0.80} \\
AI2D & 70.37 & \textbf{74.94} & \gain{4.57} & 69.56 & \textbf{70.65} & \gain{1.09} & 71.14 & \textbf{72.47} & \gain{1.33} \\
MMVP & 38.33 & \textbf{43.67} & \gain{5.34} & 26.00 & \textbf{37.00} & \gain{11.00} & 26.67 & \textbf{28.00} & \gain{1.33} \\
CV-B$^{\text{2D}}$ & 44.46 & \textbf{49.35} & \gain{4.89} & 40.19 & \textbf{45.79} & \gain{5.60} & 40.68 & \textbf{42.61} & \gain{1.93} \\
SQA & 70.09 & \textbf{73.73} & \gain{3.64} & 65.49 & \textbf{69.12} & \gain{3.63} & 64.57 & \textbf{64.60} & \gain{0.03} \\
MathV & 49.10 & \textbf{55.60} & \gain{6.50} & 50.90 & \textbf{53.90} & \gain{3.00} & 42.10 & \textbf{43.20} & \gain{1.10} \\
Hallu & 56.15 & \textbf{59.62} & \gain{3.47} & 56.38 & \textbf{58.31} & \gain{1.93} & 55.10 & \textbf{57.05} & \gain{1.95} \\
POPE & \textbf{85.90} & 85.67 & \loss{0.23} & 89.77 & \textbf{90.83} & \gain{1.06} & 88.13 & \textbf{88.87} & \gain{0.74} \\
\midrule[1.0pt]
Average & 50.01 & \textbf{53.34} & \gain{3.33} & 47.83 & \textbf{50.76} & \gain{2.93} & 45.68 & \textbf{47.22} & \gain{1.54} \\
\bottomrule[1.5pt]
\end{tabular}
}
\end{table*}

\begin{table*}
\centering
\caption{\textbf{Compatibility of LaVer with enriched visual inputs.} LaVer consistently improves performance when combined with methods that enrich visual inputs, demonstrating its broad compatibility and effectiveness across diverse visual enhancement strategies.}
\label{tab:amof}
\resizebox{\linewidth}{!}{
\begin{tabular}{c|ccccccccccccccccc|c}
\toprule[1.5pt]
$\mathcal{G}_{\upxi}$ & GQA & MMB$^{\text{EN}}$ & SEED$^{\text{I}}$ & MME & RWQA & MMMU & MM$^*$ & OCRB & TVQA & CQA & AI2D & MMVP & CV-B$^{\text{2D}}$ & SQA & MathV & Hallu & POPE & Avg. \\
\midrule[1.0pt]
A-MoF & 51.52 & 70.83 & 62.53 & 1312.14 & 54.77 & 41.11 & 41.53 & 338 & 59.46 & 41.04 & 73.45 & 41.67 & 61.34 & 67.31 & 51.20 & 62.47 & 89.13 & 51.19 \\
A-MoF + LaVer & \textbf{52.26} & \textbf{72.02} & \textbf{65.66} & \textbf{1321.51} & \textbf{56.34} & \textbf{43.22} & \textbf{43.87} & \textbf{360} & \textbf{61.68} & \textbf{42.52} & \textbf{78.38} & \textbf{44.67} & \textbf{62.66} & \textbf{68.51} & \textbf{54.40} & \textbf{62.89} & \textbf{89.57} & \textbf{52.92} \\
\bottomrule[1.5pt]
\end{tabular}
}
\end{table*}

\section{Visualization details}
we provide the details about how the figures are generated in the paper.

\textbf{Fig. 1.} We present a comprehensive performance comparison between LaVer and the baseline across 17 benchmarks, utilizing SigLIP 2~\cite{tschannen2025siglip2multilingualvisionlanguage} as the vision encoder and Qwen2.5-7B-Instruct~\cite{qwen2025qwen25technicalreport} as the LLM backbone.

\textbf{Fig. 2.} \textbf{(a)} We randomly select an image from MMVP~\cite{10655378} and perform forward passes through both the baseline and LaVer models, extracting hidden states at different layers. Using SigLIP 2~\cite{tschannen2025siglip2multilingualvisionlanguage} as the vision encoder and Qwen2.5-7B-Instruct~\cite{qwen2025qwen25technicalreport} as the LLM backbone, we normalize the hidden states and compute feature-wise cosine similarity. The visualization reveals that vision tokens exhibit substantially higher inter-feature cosine similarity in the last layer compared to middle layers, demonstrating progressive visual representation homogenization. \textbf{(b-c)} We extract the last-layer hidden states from both baseline and LaVer models and apply t-SNE~\cite{vanDerMaaten2008} to project the high-dimensional features into 2D space. The scattered features are color-coded to distinguish vision and text tokens. Compared to the baseline, LaVer's visual features exhibit stronger interaction with textual features, indicating more effective learning of joint multimodal embeddings. \textbf{(d)} We process images from MMVP~\cite{10655378} through both models and extract hidden states of vision tokens across all layers, computing the averaged cosine similarity between normalized features. For the baseline, the averaged visual cosine similarity decreases mildly in early and middle layers but increases drastically in deeper layers, indicating rapid visual information loss in the final stages. In contrast, LaVer maintains a consistent decreasing trend throughout all layers, preserving visual discriminability. \textbf{(e)} We analyze the proportion of attention allocated to vision tokens across layers, following~\cite{lei2025sail,chen2024imageworth12tokens}. Specifically, using images and corresponding queries from MMVP~\cite{10655378}, we compute the proportion of attention scores allocated to previous vision tokens for each predicted token. These layer-wise proportions are averaged across all predictions to reveal the overall trend. The comparison demonstrates that LaVer enables the model to allocate significantly more attention to vision tokens, indicating more effective utilization of visual representations.

\textbf{Fig. 4.} \textbf{(a)} We extract the hidden states of vision tokens from the last layer and reshape them to recover their spatial structure. Applying PCA~\cite{Abdi2010} for dimensionality reduction to 3 components, we normalize the features to the range $[0, 255]$ and visualize them as a 3-channel RGB image. The visualization demonstrates that LaVer generates highly discriminative visual features while preserving clear spatial structural information. All experiments use images from MMVP~\cite{10655378}, SigLIP 2~\cite{tschannen2025siglip2multilingualvisionlanguage} as the vision encoder, and Qwen2.5-7B-Instruct~\cite{qwen2025qwen25technicalreport} as the LLM backbone. \textbf{(b)} We analyze the training dynamics of the baseline and models trained with different loss configurations. Specifically, using images from MMVP~\cite{10655378}, we compute the averaged normalized cosine similarity of last-layer visual features at 1K iteration intervals. While naive application of MIM leads to visual feature inconsistency, LaVer effectively prevents the MIM shortcut, guiding the model toward more discriminative visual representations throughout training.

\textbf{Fig. 6.} We visualize the attention scores of the last predicted token on images from MMVP~\cite{10655378}, alongside the corresponding PCA visualization of last-layer visual features. The results demonstrate that LaVer enhances the model's ability to focus on spatially relevant regions that correspond to the generated text tokens, indicating improved visual-textual alignment.

\section{Additional Experiment Results}

\textbf{Full results of the parameter scaling property of LaVer.}
Table~\ref{tab:model_scale} presents a comprehensive evaluation of LaVer's scalability across different model parameter sizes, ranging from Qwen2.5-1.5B-Instruct to Qwen2.5-7B-Instruct~\cite{qwen2025qwen25technicalreport}, using both SigLIP 2~\cite{tschannen2025siglip2multilingualvisionlanguage} and CLIP~\cite{pmlr-v139-radford21a} vision encoders. The results demonstrate that LaVer consistently delivers performance improvements over the baseline across all parameter scales. For SigLIP 2-based models, LaVer achieves average improvements of \textbf{+2.42}, \textbf{+2.07}, and \textbf{+2.15} points for 1.5B, 3B, and 7B parameter configurations, respectively. Similarly, for CLIP-based models, LaVer yields average gains of \textbf{+3.12}, \textbf{+1.31}, and \textbf{+2.66} points across the same parameter scales. Notably, LaVer exhibits particularly strong improvements on challenging benchmarks such as OCR-Bench~\cite{Liu_2024} (up to \textbf{+103} points for SigLIP 2-7B) and MME~\cite{fu2025mmecomprehensiveevaluationbenchmark} (up to \textbf{+185.03} points for CLIP-7B), ChartQA~\cite{masry-etal-2022-chartqa} (up to \textbf{+12.00} points for CLIP-7B), and MathVista~\cite{lu2024mathvistaevaluatingmathematicalreasoning} (up to \textbf{+7.20} points for SigLIP 2-1.5B). The consistent positive gains across 17 diverse benchmarks and multiple parameter scales substantiate that LaVer possesses robust scaling properties with respect to model parameters, maintaining its effectiveness in mitigating visual representation homogenization regardless of model capacity.

\textbf{Full results of the data scaling property of LaVer.}
Table~\ref{tab:data_scale} presents a comprehensive analysis of LaVer's scalability across different training dataset sizes, ranging from 800K to 4M samples, using both SigLIP 2~\cite{tschannen2025siglip2multilingualvisionlanguage} and CLIP~\cite{pmlr-v139-radford21a} vision encoders with Qwen2.5-7B-Instruct~\cite{qwen2025qwen25technicalreport} as the language model. The results demonstrate that LaVer consistently delivers substantial performance improvements over the baseline across all data scales. For SigLIP 2-based models, LaVer achieves average improvements of \textbf{+2.15}, \textbf{+2.16}, and \textbf{+1.80} points for 800K, 2M, and 4M training samples, respectively. Similarly, for CLIP-based models, LaVer yields progressively increasing average gains of \textbf{+2.66}, \textbf{+3.04}, and \textbf{+3.77} points across the same data scales, indicating enhanced effectiveness with larger training datasets. Notably, LaVer exhibits particularly strong improvements on challenging benchmarks such as TextVQA~\cite{singh2019vqamodelsread} (up to \textbf{+8.61} points for CLIP-2M). The consistent positive gains across 17 diverse benchmarks and multiple data scales substantiate that LaVer possesses robust scaling properties with respect to training data size, maintaining its effectiveness in mitigating visual representation homogenization regardless of dataset scale. Furthermore, the observation that CLIP-based models show increasing improvements with larger datasets (from \textbf{+2.66} to \textbf{+3.77}) suggests that LaVer's benefits become more pronounced when trained on more extensive data, highlighting its potential for further performance gains with additional training data.

\textbf{Full results of the ablation study on masking strategies.}
Table~\ref{tab:masking_ablation} presents a comprehensive analysis of LaVer's performance under different masking strategies, examining both cosine and constant scheduling approaches across various masking ratios. The results are evaluated on 17 diverse benchmarks using SigLIP 2~\cite{tschannen2025siglip2multilingualvisionlanguage} and CLIP~\cite{pmlr-v139-radford21a} vision encoders with Qwen2.5-7B-Instruct~\cite{qwen2025qwen25technicalreport} as the language model. For SigLIP 2-based models, the cosine scheduling strategy with a masking ratio of 0.05 achieves the best average performance of \textbf{57.87}, representing a substantial improvement of \textbf{+2.15} points over the baseline (55.72). This configuration demonstrates consistent gains across most benchmarks. Comparing scheduling strategies, cosine scheduling generally outperforms constant scheduling across different masking ratios. For instance, at a ratio of 0.05, cosine scheduling achieves \textbf{57.87} compared to constant scheduling's \textbf{56.32}, demonstrating the effectiveness of gradually varying masking intensity during training. Regarding masking ratio selection, lower ratios (0.05-0.1) consistently yield superior performance compared to higher ratios (0.2-0.3) under cosine scheduling, suggesting that moderate masking preserves sufficient visual information while effectively mitigating representation homogenization. For CLIP-based models, similar trends emerge with cosine scheduling at 0.05 ratio achieving the best average performance of \textbf{53.24}, representing a \textbf{+2.66} point improvement over the baseline. The results demonstrate LaVer's robustness across different masking configurations, with the cosine scheduling strategy at lower masking ratios consistently delivering optimal performance across both vision encoders and diverse evaluation benchmarks.
We hypothesize that the inferior performance observed with high masking ratios stems from insufficient convergence, as conventional MIM-based methods typically require large-scale training to fully develop the model's reconstruction capabilities~\cite{siméoni2025dinov3,oquab2024dinov,9709990}.
Scaling to larger datasets remains an avenue for future investigation.

\textbf{Full results of the ablation study on EMA updating strategies.}
Table~\ref{tab:teacher_ablation} presents a comprehensive analysis of LaVer's performance under different EMA teacher updating strategies, evaluated across diverse benchmarks using SigLIP 2~\cite{tschannen2025siglip2multilingualvisionlanguage} and CLIP~\cite{pmlr-v139-radford21a} vision encoders with Qwen2.5-7B-Instruct~\cite{qwen2025qwen25technicalreport} as the LLM backbone. The results demonstrate LaVer's robustness across various EMA configurations, with cosine scheduling slightly outperforming constant scheduling. Regarding updating frequency, moderate frequencies (50-200 steps) generally yield optimal performance, with 100 steps emerging as the most balanced configuration across both vision encoders. For decay rate selection, a rate of 0.95 demonstrates superior stability and performance compared to both lower (0.9) and higher (0.99, 0.999) rates. The comprehensive evaluation across diverse benchmarks confirms that LaVer's performance remains consistently strong regardless of the specific EMA strategy employed, highlighting the method's robustness and effectiveness in learning intrinsic visual representation capabilities.

\textbf{Full results of the ablation study on spatial awareness.}
Table~\ref{tab:spatial_ablation} presents a comprehensive analysis of LaVer's spatial awareness mechanisms, evaluated across diverse benchmarks. The results reveal several key insights. First, employing mixed attention alone (full attention for vision tokens) yields moderate improvements over the baseline, demonstrating that enhanced token interactions can benefit visual understanding. Second, applying 2D-RoPE in isolation shows minimal impact, with performance remaining largely comparable to the baseline. Third, combining mixed attention with 2D-RoPE without LaVer produces mixed results, with performance gains that are inconsistent across benchmarks. Most importantly, the full LaVer framework, which integrates all three components achieves the best overall performance. This configuration demonstrates consistent improvements across challenging benchmarks. These results confirm that LaVer's holistic approach to spatial awareness, combining learned visual representations with architectural enhancements, is essential for achieving superior multimodal understanding capabilities.

\begin{figure*}
\centering
\resizebox{0.8\linewidth}{!}{
\begin{subfigure}{0.4\linewidth}
    % \fbox{\rule{0pt}{2.25in} \rule{.9\linewidth}{0pt}}
    \includegraphics[width=0.99\linewidth]{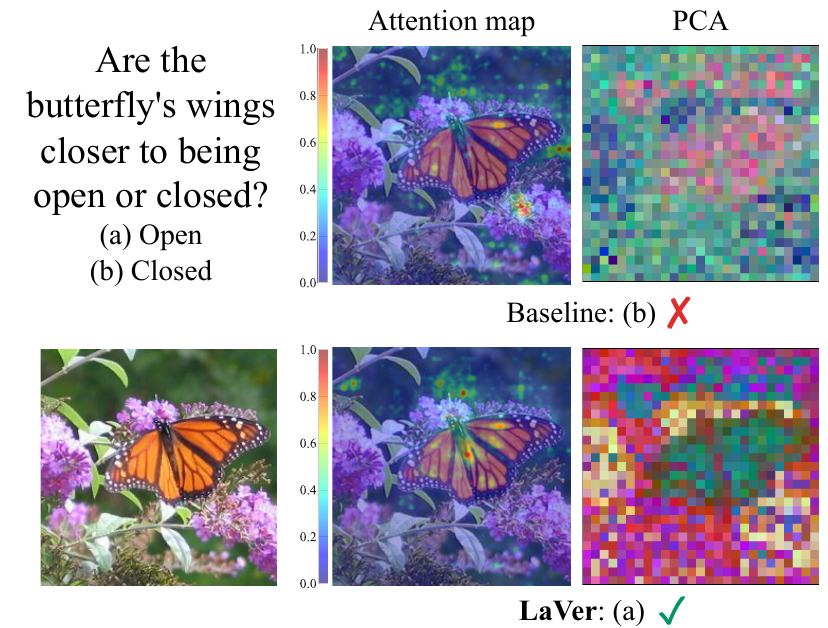}
    % \caption{Parameter Scaling of LaVer.}
    \label{fig:supp_qual_0}
\end{subfigure}
\hfill
\begin{subfigure}{0.4\linewidth}
    % \fbox{\rule{0pt}{2.25in} \rule{.9\linewidth}{0pt}}
    \includegraphics[width=0.99\linewidth]{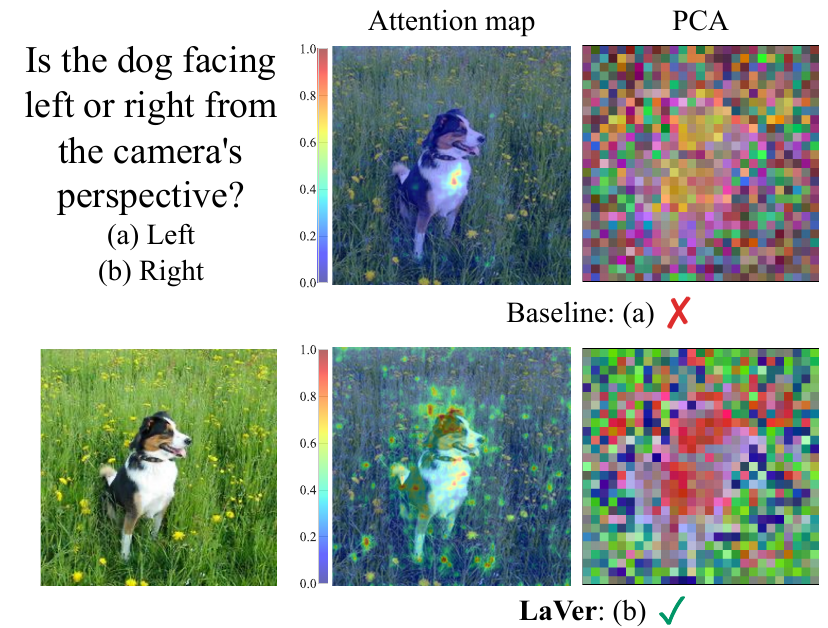}
    % \caption{Data Scaling of LaVer.}
    \label{fig:supp_qual_1}
\end{subfigure}
}
\resizebox{0.8\linewidth}{!}{
\begin{subfigure}{0.4\linewidth}
    % \fbox{\rule{0pt}{2.25in} \rule{.9\linewidth}{0pt}}
    \includegraphics[width=0.99\linewidth]{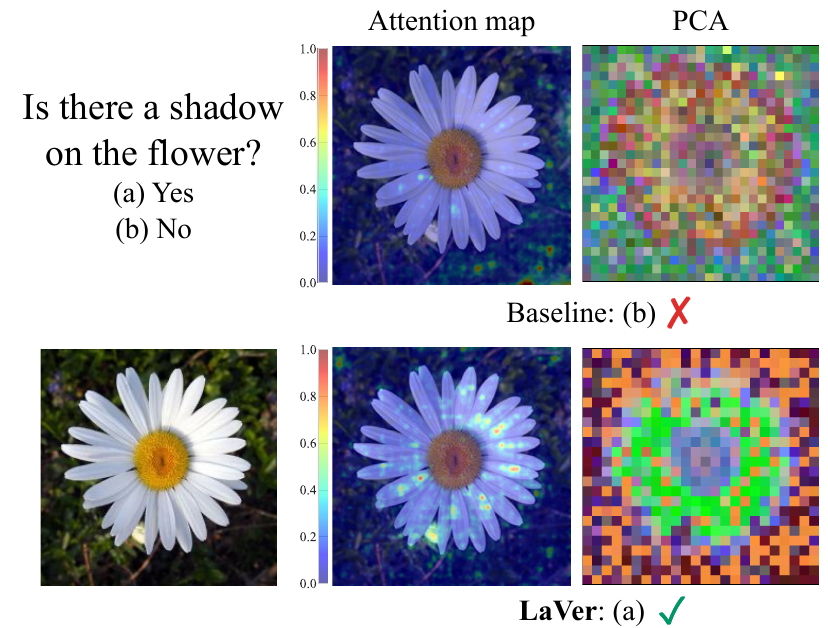}
    % \caption{Parameter Scaling of LaVer.}
    \label{fig:supp_qual_2}
\end{subfigure}
\hfill
\begin{subfigure}{0.4\linewidth}
    % \fbox{\rule{0pt}{2.25in} \rule{.9\linewidth}{0pt}}
    \includegraphics[width=0.99\linewidth]{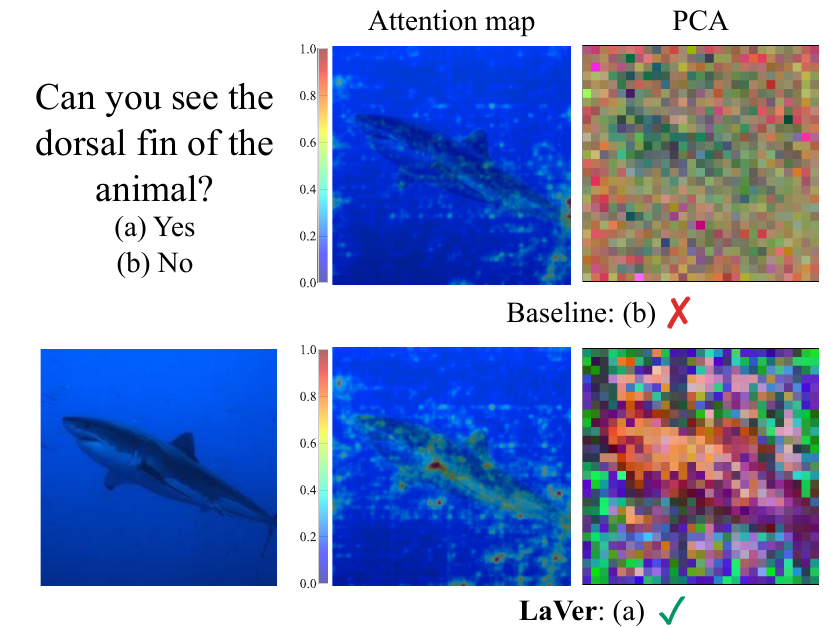}
    % \caption{Data Scaling of LaVer.}
    \label{fig:supp_qual_3}
\end{subfigure}
}
\resizebox{0.8\linewidth}{!}{
\begin{subfigure}{0.4\linewidth}
    % \fbox{\rule{0pt}{2.25in} \rule{.9\linewidth}{0pt}}
    \includegraphics[width=0.99\linewidth]{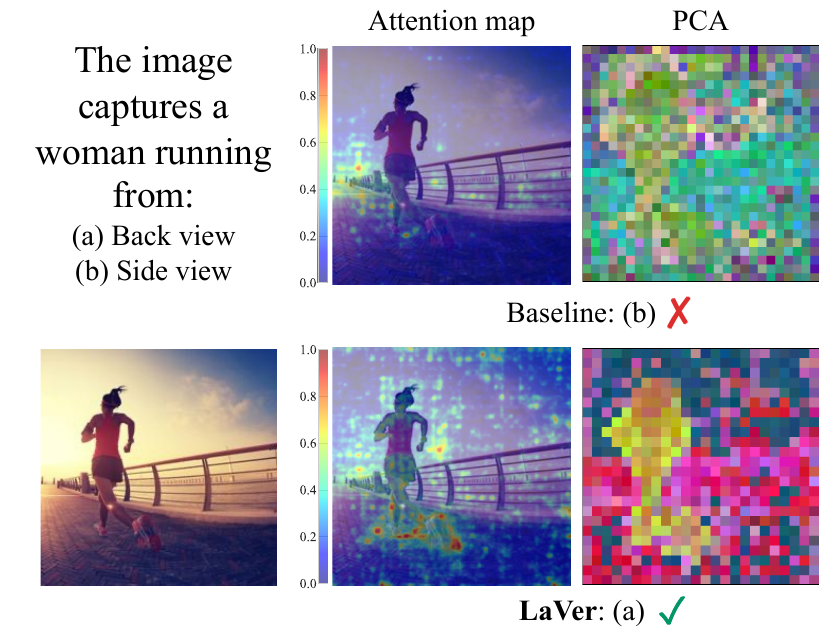}
    % \caption{Parameter Scaling of LaVer.}
    \label{fig:supp_qual_4}
\end{subfigure}
\hfill
\begin{subfigure}{0.4\linewidth}
    % \fbox{\rule{0pt}{2.25in} \rule{.9\linewidth}{0pt}}
    \includegraphics[width=0.99\linewidth]{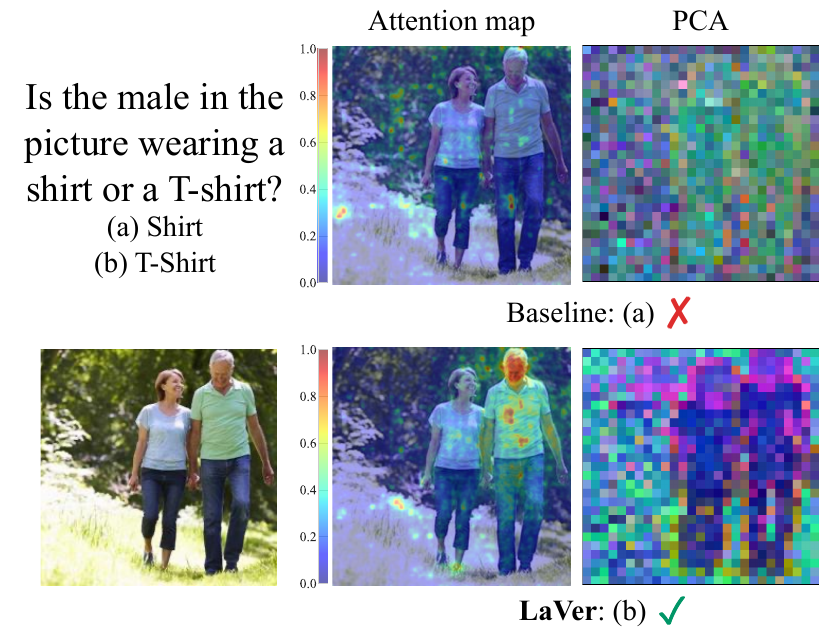}
    % \caption{Data Scaling of LaVer.}
    \label{fig:supp_qual_5}
\end{subfigure}
}
\caption{\textbf{Qualitative comparisons on MMVP~\cite{10655378}.}}
\label{fig:supp_qual}
\end{figure*}

\textbf{Full results of the ablation study on loss components.}
Table~\ref{tab:loss_ablation} provides a comprehensive analysis of different loss function configurations in LaVer, evaluated across diverse benchmarks. The results reveal three critical insights into the design of effective visual representation learning objectives. First, applying masked image modeling ($\mathcal{L}_{\text{MIM}}$) alone significantly degrades performance, with average scores dropping from \textbf{55.72} to \textbf{53.76} for SigLIP 2 and from \textbf{50.58} to \textbf{49.71} for CLIP. This deterioration stems from the \textit{visual feature inconsistency} problem, where the model exploits the MIM objective by generating identical visual features, thereby undermining the discriminative capacity essential for downstream tasks. Second, incorporating the global alignment loss ($\mathcal{L}_{\text{GA}}$) alongside $\mathcal{L}_{\text{MIM}}$ partially mitigates this issue, improving average performance to \textbf{56.46} for SigLIP 2 and \textbf{52.01} for CLIP. However, this configuration still underperforms the full LaVer framework, as the symmetric nature of $\mathcal{L}_{\text{GA}}$ constrains the model's ability to learn sufficiently discriminative representations. Third, our proposed contrastive global alignment loss ($\mathcal{L}_{\text{CGA}}$) effectively addresses both shortcomings, achieving the best overall performance with average scores of \textbf{57.87} for SigLIP 2 and \textbf{53.24} for CLIP. These results confirm that $\mathcal{L}_{\text{CGA}}$ simultaneously prevents visual feature inconsistency while encouraging discriminative feature learning, thereby enhancing both visual information preservation and utilization across diverse multimodal understanding tasks.

\textbf{Language capabilities.}
A potential concern when introducing vision-centric supervisory signals is whether they may compromise the model's language capabilities. To address this, we evaluate the language performance of both the baseline and our LaVer model using SigLIP 2~\cite{tschannen2025siglip2multilingualvisionlanguage} and CLIP~\cite{pmlr-v139-radford21a} as vision encoders, respectively, as shown in Table~\ref{tab:lang_performance}. Specifically, we adopt three representative benchmarks: IFEval~\cite{zhou2023instructionfollowingevaluationlargelanguage}, which evaluates the model's capability to strictly follow instructions; MMLU~\cite{hendrycks2021measuringmassivemultitasklanguage}, a comprehensive general language benchmark containing diverse tasks across various domains; and BBH~\cite{suzgun2022challengingbigbenchtaskschainofthought}, which assesses language reasoning and world knowledge capabilities. The results demonstrate that LaVer maintains competitive language performance across both vision encoder configurations. These findings confirm that LaVer successfully preserves language capabilities while simultaneously achieving substantial improvements in visual understanding tasks, thereby validating the effectiveness of our approach in balancing multimodal learning objectives.

\textbf{Extended comparison with different LLM backbone.}
To validate the generalizability of LaVer across different architectural configurations, we conduct comprehensive experiments using Vicuna-7B-v1.5~\cite{zheng2023judging} as the LLM backbone, paired with three vision encoders: SigLIP 2~\cite{tschannen2025siglip2multilingualvisionlanguage}, CLIP~\cite{pmlr-v139-radford21a}, and DINOv2~\cite{oquab2024dinov}.
As shown in Table~\ref{tab:vicuna}, LaVer consistently outperforms the baseline across all three vision encoder configurations, achieving average improvements of \textbf{+1.58}, \textbf{+1.57}, and \textbf{+2.37} points for SigLIP 2, CLIP, and DINOv2, respectively.
Notably, LaVer demonstrates substantial gains on benchmarks requiring fine-grained visual understanding, including ChartQA~\cite{masry-etal-2022-chartqa} (\textbf{+2.32}, \textbf{+6.72}, \textbf{+0.60}), MMVP~\cite{10655378} (\textbf{+1.33}, \textbf{+3.67}, \textbf{+2.34}), and CV-Bench-2D~\cite{10.5555/3737916.3740687} (\textbf{+2.36}, \textbf{+1.51}, \textbf{+4.02}).
These results confirm that LaVer effectively preserves and enhances multimodal capabilities across different LLM architectures, demonstrating its robustness and broad applicability in diverse scenarios.

\textbf{Performance comparison of stage 2.}
Following the state-of-the-art open-source framework LLaVA-OneVision~\cite{an2025llavaonevision15fullyopenframework}, we adopt a 3-stage training recipe and exclusively apply LaVer to stage 2, which is responsible for visual knowledge injection.
To thoroughly investigate how LaVer facilitates visual knowledge learning, we compare models with and without LaVer during stage 2, using three vision encoders: SigLIP 2~\cite{tschannen2025siglip2multilingualvisionlanguage}, CLIP~\cite{pmlr-v139-radford21a}, and DINOv2~\cite{oquab2024dinov}.
As presented in Table~\ref{tab:stage2}, LaVer consistently demonstrates substantial improvements across all configurations, achieving average gains of \textbf{+3.33}, \textbf{+2.93}, and \textbf{+1.54} points for SigLIP 2, CLIP, and DINOv2, respectively.
Notably, LaVer exhibits particularly strong performance on benchmarks requiring fine-grained visual understanding: MMVP~\cite{10655378} shows remarkable improvements of \textbf{+5.34}, \textbf{+11.00} across the three encoders.
These results reveal that the performance gains observed in stage 2 are even more pronounced than in the final stage, indicating that LaVer fundamentally enhances the model's capacity to learn representative visual features during the critical visual knowledge injection phase, thereby establishing a strong foundation for superior multimodal performance across diverse downstream tasks.

\textbf{Compatibility with other methods.}
LaVer introduces vision-centric supervisory signals by predicting masked vision tokens, a design that naturally ensures compatibility with methods that enrich visual inputs.
To validate this compatibility, we integrate LaVer with A-MoF~\cite{10655378}, which aggregates visual features from multiple vision encoders to form more informative visual embeddings.
Specifically, following~\cite{10655378}, we combine visual features from CLIP-ViT-L/14@224~\cite{pmlr-v139-radford21a} and DINOv2-Large/14@224~\cite{oquab2024dinov} with balance coefficients of ${0.25, 0.75}$, which are empirically validated as optimal hyperparameters in~\cite{10655378}.
The identical patch size and resolution ensure that visual features share the same shape and can be directly aggregated.
As presented in Table~\ref{tab:amof}, LaVer consistently delivers substantial improvements across all benchmarks when integrated with A-MoF.
Overall, LaVer achieves an average improvement of \textbf{+1.73} points across all benchmarks, demonstrating that it provides additional benefits when integrated with vision-enhanced methods and confirming its broad compatibility with diverse visual enhancement strategies.

\section{Qualitative Analysis}

We provide additional qualitative comparisons in Fig.~\ref{fig:supp_qual} on MMVP~\cite{10655378}.
As illustrated in Fig.~\ref{fig:supp_qual}, we visualize both the attention scores on vision tokens from the last predicted token and the PCA visualization of visual features extracted from the final layer.
The baseline model frequently fails to attend to spatial regions that are semantically relevant to the text query, exhibiting limited attention distributions across the visual input.
In contrast, our LaVer consistently allocates substantially higher attention weights to the corresponding regions of interest, thereby enabling more accurate and contextually appropriate responses.
Furthermore, LaVer demonstrates the ability to capture diverse visual patterns and generate highly discriminative visual features that encode rich structural information.
These qualitative results provide compelling evidence that LaVer effectively mitigates modality imbalance by incorporating visual supervisory signals, enabling the model to integrate visual and textual modalities seamlessly and achieve superior performance on multimodal tasks.

% \section{Limitations}

% failed to scaling up to larger models and bigger datasets to fully demonstates the potential of LaVer

% limited to only one-image scenario, furthern extension to multi-image, video, and audio modalities are left for future work.

% \section{Rationale}
% \label{sec:rationale}
% % 
% Having the supplementary compiled together with the main paper means that:
% % 
% \begin{itemize}
% \item The supplementary can back-reference sections of the main paper, for example, we can refer to \cref{sec:intro};
% \item The main paper can forward reference sub-sections within the supplementary explicitly (e.g. referring to a particular experiment); 
% \item When submitted to arXiv, the supplementary will already included at the end of the paper.
% \end{itemize}
% % 
% To split the supplementary pages from the main paper, you can use \href{https://support.apple.com/en-ca/guide/preview/prvw11793/mac#:~:text=Delete%20a%20page%20from%20a,or%20choose%20Edit%20%3E%20Delete).}{Preview (on macOS)}, \href{https://www.adobe.com/acrobat/how-to/delete-pages-from-pdf.html#:~:text=Choose%20%E2%80%9CTools%E2%80%9D%20%3E%20%E2%80%9COrganize,or%20pages%20from%20the%20file.}{Adobe Acrobat} (on all OSs), as well as \href{https://superuser.com/questions/517986/is-it-possible-to-delete-some-pages-of-a-pdf-document}{command line tools}.